\documentclass[a4]{article}

\usepackage{amsthm}
\usepackage{bm}
\usepackage{amsmath}
\usepackage{mathtools}
\usepackage{subfigure}
\usepackage{xcolor}
\usepackage{here}
\usepackage{natbib}
\usepackage{cite}
\usepackage{url}
\usepackage{authblk}
\usepackage{listings}
\usepackage[ruled,linesnumbered]{algorithm2e}

\usepackage{wrapfig}

\usepackage{graphicx}
\usepackage{amssymb}

\newcommand{\memo}[1]{} % メモを消す

\def\Figref#1{Figure~\ref{#1}}

\def\eqref#1{equation~\ref{#1}}
\SetKwRepeat{Do}{do}{while}
\SetKwBlock{Function}{function}{end function}
\SetKwComment{CommentTmp}{\blue{$\triangleright$\:}}{}
\DontPrintSemicolon
\newcommand{\Comment}[1]{\CommentTmp*[r]{\small\textcolor{blue}{#1}\!\!\!\!\!\!}}

\newcommand{\red}[1]{\textcolor{red}{#1}}
\newcommand{\blue}[1]{\textcolor{blue}{#1}}

\def\*#1{\boldsymbol{#1}}
\theoremstyle{plain}

\newtheorem{theo}{Theorem}[section]

\newtheorem{defini}{Definition}[section]
\newtheorem{lem}{Lemma}[section]
\newtheorem{coro}{Corollary}[section]
\newtheorem{rema}{Remark}[section]

\newcommand{\ol}[1]{\overline{#1}}

\newcommand{\prm}[0]{\prime}
\newcommand{\tw}[0]{\textwidth}
\newcommand{\ig}[2]{\includegraphics[clip,width=#1\tw]{#2}}

\newcommand{\mr}[1]{\mathrm{#1}}

\newcommand{\pd}[2]{{\frac{\partial #1}{\partial #2}}}

\newcommand{\argmin}{\mathop{\text{argmin}}}
\newcommand{\eq}[1]{(\ref{#1})}

\newcommand{\lw}[1]{\smash{\lower2.ex\hbox{#1}}}
\newcommand{\inner}[2]{\left\langle #1,#2 \right\rangle}
\newcommand{\rbr}[1]{\left(#1\right)}

\newcommand{\cbr}[1]{\left\{#1\right\}}
\newcommand{\nbr}[1]{\left\|#1\right\|}
\newcommand{\abr}[1]{\left|#1\right|}

\newcommand{\RR}{\mathbb{R}}

\newcommand{\II}{\mathbb{I}}

\newcommand{\cB}{{\mathcal B}}
\newcommand{\cC}{{\mathcal C}}

\newcommand{\cG}{{\mathcal G}}
\newcommand{\cH}{{\mathcal H}}

\newcommand{\cS}{{\mathcal S}}

\newcommand{\cY}{{\mathcal Y}}

\usepackage{geometry}
\ifdefined\draft
\newcommand{\note}[1]{}
\newcommand{\remove}[1]{}
\else
\newcommand{\note}[1]{\red{(Memo: #1)}} % メモ
\newcommand{\remove}[1]{
\begingroup
\color{blue}
#1
\endgroup
}
\fi

\title{Exact Subgraph Isomorphism Network \\ with Mixed $L_{0,2}$ Norm Constraint \\ for Predictive Graph Mining}

\author[1]{Taiga~Kojima}
\author[1]{Haruto~Kajita}
\author[1]{Ayato~Kohara}
\author[1]{Masayuki~Karasuyama\thanks{karasuyama@nitech.ac.jp}}
\affil[1]{Nagoya Institute of Technology}

\date{}

\newcommand{\biblstyle}{\bibliographystyle{apalike}}

\begin{document}

\maketitle

\begin{abstract}
 In the graph-level prediction task (predict a label for a given graph), the information contained in subgraphs of the input graph plays a key role.
 In this paper, we propose Exact subgraph Isomorphism Network (EIN), which combines the exact subgraph enumeration, a neural network, and a sparse regularization by the mixed $L_{0,2}$ norm constraint. 
 In general, building a graph-level prediction model achieving high discriminative ability along with interpretability is still a challenging problem.
 Our combination of the subgraph enumeration and neural network contributes to high discriminative ability about the subgraph structure of the input graph.
 Further, the sparse regularization in EIN enables us 1) to derive an effective pruning strategy that mitigates computational difficulty of the enumeration while maintaining the prediction performance, and 2) to identify important subgraphs that contributes to high interpretability.
 We empirically show that EIN has sufficiently high prediction performance compared with standard graph neural network models, and also, we show examples of post-hoc analysis based on the selected subgraphs. 
\end{abstract}

\section{Introduction}

% グラフ分類問題など，グラフを入力とする予測問題は，データサイエンス界で広く研究されている．
Graph-level prediction tasks, which take a graph as an input and predict a label for the entire graph, have been widely studied in the data-science community.
%
% グラフ表現は構造データを捉えるのに有効であり，グラフベースの機械学習アルゴリズムは，化学組成分析 \citep{ralaivola2005graph,faber2017prediction} や結晶構造分析 \citep{xie2018crystal,louis2020graph} など様々な応用問題に適用されている．
It is known that the graph representation is an effective approach to a variety of structure data such as chemical compounds \citep{ralaivola2005graph,faber2017prediction}, protein structures \citep{gligorijevic2021structure}, and inorganic crystal structures \citep{xie2018crystal,louis2020graph}. 
%
% 実世界のグラフデータにおいて，その部分構造，つまり部分グラフ，が重要であることが多い．
In a graph-level prediction task, substructures on the input graph, i.e., subgraphs, are often an important factor for the prediction and the analysis.
%
% 例えば，化合物の性質の予測の際に，どのような分子構造が重要なのかを知ることは有益な知見をもたらす可能性がある．
For example, in a prediction of a property of chemical compounds, identifying small substructures of the molecules can be essential for both of improving prediction accuracy and obtaining an insight about the underlying chemical mechanism.
%
% そのため，重要な部分構造の抽出は重要であり，また，部分構造が複雑な関係性を持って出力に依存する可能性がありモデルの自由度も担保したい．
Therefore, mining predictive subgraphs is a significant issue for graph-level prediction tasks, and further, those subgraphs can have a higher order dependency to the prediction that requests sufficient flexibility in the model.
%
% しかしながら重要な部分構造の発見とモデルの柔軟性を両立することは非常に難しい問題である．
However, building a prediction method that satisfies these requirements is a still challenging problem (see \S~\ref{sec:related-work} for existing studies).

\begin{figure*}[t]
 \begin{center}
  \ig{.8}{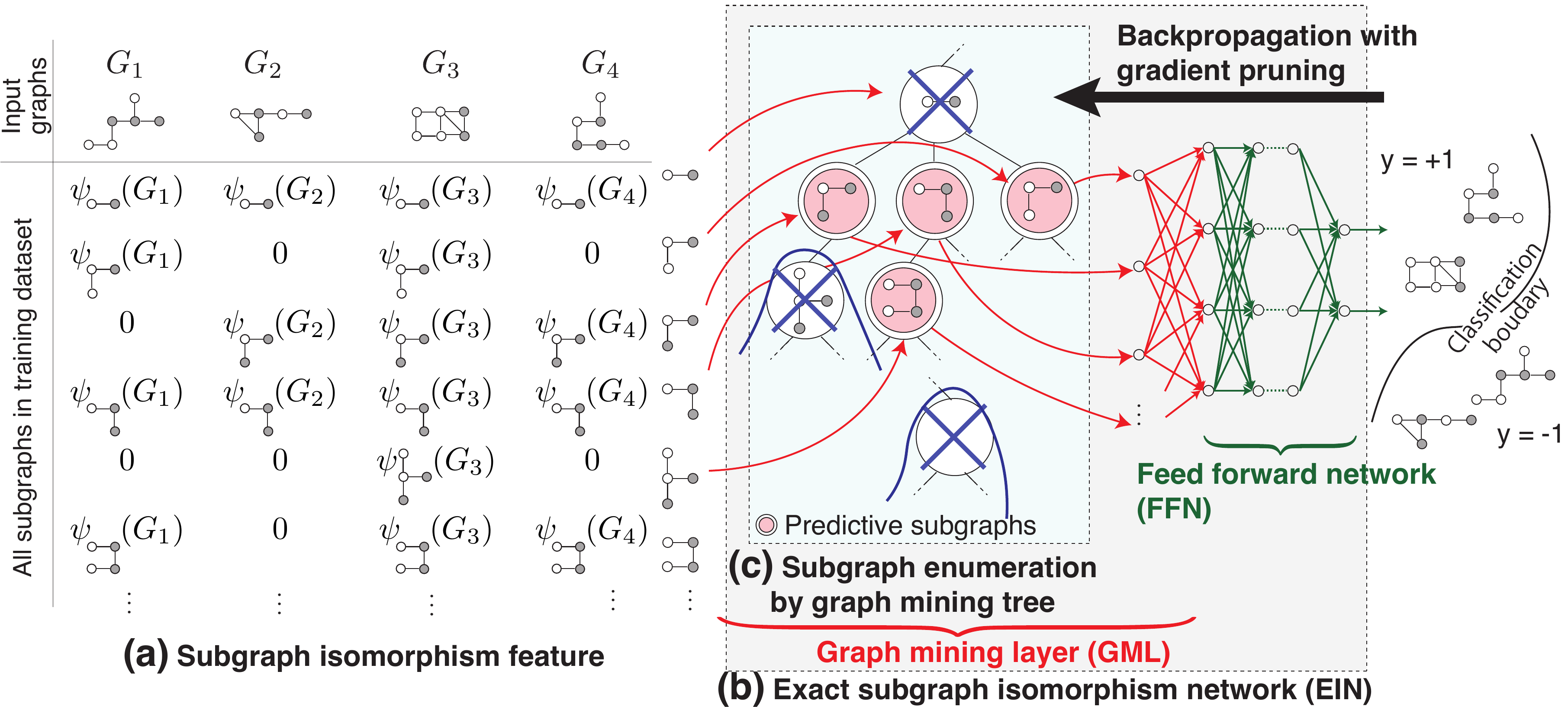}
 \end{center}
 \caption{
 Overview of proposed method. 
 }
 \label{fig:overview}
\end{figure*}

% 提案法EIN (Exact subgraph isomorphism network)は予測に重要な部分グラフを同定しつつ，柔軟な予測モデルを構築できるものである．
Our proposed method, called Exact subgraph Isomorphism Network (EIN), adaptively identifies predictive subgraphs based on which a neural network model can be simultaneously trained.
%
% EINの概念図を\figurename~\ref{fig:overview}に示す．
The overview of EIN is shown in Fig.~\ref{fig:overview}.
%
% をグラフ$G$が部分グラフ$H$を包含する場合に1，そうでなければ0を返す評価関数とする (厳密な部分グラフ同型判定). 
Fig.~\ref{fig:overview}~(a) illustrates the subgraph representation of EIN denoted as 
$\psi_H(G)$
which takes a non-zero value if the input graph $G$ contains a subgraph $H$ and takes $0$ otherwise (i.e., it is based on the exact subgraph isomorphism).
%
% \figurename~\ref{fig:overview}~(b)が示す通り，EINは訓練データに含まれる全てのsubgraphsを入力候補とする予測ネットワークである．
As shown in Fig.~\ref{fig:overview}~(b), EIN can be seen as a neural network in which the candidates of the input features are all the subgraphs contained in the training dataset.
%
% このアーキテクチャでは，部分グラフの存在を厳密に判定して特徴表現
% $\psi_H(G)$
% を定義するため，典型的なメッセージパッシングベースのグラフニューラルネットワークでは判別できないグラフ構造の差異を見分けられる．
% Because of its exact subgraph isomorphism representation, this architecture can be discriminative even for graphs that cannot be discriminated by classical message passing based graph neural networks.
Because of its exact subgraph isomorphism representation, this architecture can be highly discriminative about subgraph structures.
%
% しかしながら，$f(G)$の評価及び最適化には，大量の部分グラフに対する部分グラフ同定が発生するため愚直な実装では計算量的な困難を伴う．
However, since the number of the candidate subgraphs can be enormous, the na{\"i}ve computation of this architecture is computationally intractable.

% --------------------------------------------------
% Illustration of model
% --------------------------------------------------

% EINはGraph Mining Layer (GML)とFeed Forward Network (FFN)から構成される．
EIN consists of Graph Mining Layer (GML) and Feed Forward Network (FFN).
%
% GMLは予測に必要な少数の部分グラフのみを選別する機構をグループスパース正則化により実現する．
% 英訳してください: このlayerは予測に必要な少数の変数のみを選別する機構をグループスパース正則化により実現する．
% GML implements a mechanism to select only a small number of subgraphs necessary for the prediction, achieved through the group-sparse regularization.
% GML implements a mechanism to select only a small number of subgraphs necessary for the prediction, achieved through \red{a group sparse constraint \citep{yuan2006model}.}
GML implements a mechanism that identifies only $s$ subgraphs important for the prediction.
This is achieved through a group sparse constraint \citep{yuan2006model}, where $s$ is a pre-specified small integer (typically, less than $100$).
%
% グループスパース正則化\citep{yuan2006model}は変数をグループ単位で取捨選択できる正則化である．
% The group sparse regularization \citep{yuan2006model} is a well-known regularizer for the group-wise selection of variables. 
%
% ある部分グラフに対応する入力unitから次のlayerに向かう重みベクトルをグループとすることで，部分グラフの取捨選択が実現できる．
We regard a set of neural network weights from an input unit (corresponding to one subgraph) to the next layer as a group, by which the adaptive subgraph selection can be realized.
%
% 選択された部分グラフのみが次のFFNに向かうことで，部分グラフ間の依存関係が推定される．
Only the selected subgraphs are used in the subsequent FFN that learns the dependency between the selected subgraphs and the prediction target.
%
% しかしこのアプローチを用いても，重要部分グラフは事前には未知であるため，最適化の過程では膨大な部分グラフを全て考慮しなければならない．
Although the number of subgraphs becomes small in the trained model, this approach still suffers from the computational difficulty because all the candidate subgraphs are still required to consider during the optimization process.

% EINでは$\*\beta_{H_i}$に対してグループスパース正則化のペナルティ$\| \*\beta_{H_i} \|_2$を課す．
% %
% ただし，$\| \cdot \|_2$は２ノルムである．
% %
% これにより，$\*\beta_{H_i}$の多くが$\*0$となり，少数の$H_i$だけが$f(G)$に影響を及ぼすようにできる．
% %
% 結果として，ゼロでない$\*\beta_{H_i}$の$H_i$として重要な部分グラフを発見でき，かつ，$f(G)$の計算時に考えなければならない部分グラフ$H_i$の数を劇的に減らすことができる．
% %
% しかしこのアプローチを用いても，重要部分グラフは事前には未知であるため，最適化の過程では膨大な部分グラフを全て考慮しなければならない．
%}

%\red{
% そこで本論文ではさらに，グラフマイニングによる部分グラフ列挙を近接勾配法の最適化と組み合わせることで，寄与しない部分グラフを枝刈りする機構を提案する(図\ref{fig:overview}~(c))．
% To tackle the computational difficulty, we combine a subgraph enumeration by graph mining \citep{yan2002gspan} and the proximal gradient optimization \citep{Teboulle2017simplified,bech2009fast}, by which a pruning strategy for unnecessarily subgraphs can be derived, shown in \ref{fig:overview}~(c).
To tackle the computational difficulty, we combine a subgraph enumeration by graph mining \citep{yan2002gspan} and the iterative hard thresholding (IHT) \citep[e.g.,][]{blumensath2008iterative,herrity2006sparse,jain2014iterative,damadi2024learning}, by which a pruning strategy for unnecessarily subgraphs can be derived, as illustrated in \ref{fig:overview}~(c).
%
% 近接勾配法\citep{Teboulle2017simplified,bech2009fast}はスパース正則化の標準的な最適化アルゴリズムとして広く知られている．
% 近接勾配法は近接射影により，パラメータを閾値処理しながら更新する．
% The proximal gradient is a standard approach for the sparse modeling, in which parameters are updated through a proximal projection that typically results in a thresholding operation.
%
% この閾値処理により，現在$\*0$である$\*\beta_{H}$に対して，誤差逆伝播された勾配のノルムがある閾値を下回る場合には更新の必要がないことになる．
% この閾値処理により，勾配のノルムがある閾値を下回る場合には更新の必要がないことになる．
% This thresholding operation reveals that if the norm of the gradient corresponding to a subgraph $H$ is less than certain threshold, parameters for that subgraph is not required to update. 
In IHT, only when the norm of the gradient corresponding to a subgraph $H$ ranks within the top $s$-th largest values among all the possible subgraphs, parameters for that subgraph is required to update. 
% 英訳してください: 変数の値がx位以内に入らないなら更新する必要がない
% if the norm of the gradient corresponding to a subgraph $H$ is less than certain threshold, parameters for that subgraph is not required to update. 
%
% 我々は列挙木の部分木全体に対して勾配のノルムの上限を導出することで，枝刈り戦略を定義できることを示す．
We show that, by deriving an upper bound of the norm of the gradient, an efficient pruning strategy for the subgraph enumeration can be constructed.
%
% この枝刈り戦略には，$2$つの大きな利点がある．
Our pruning strategy has the following two important implications. 
%
% 第一が，枝刈りにより全ての候補部分グラフ$H_i$を列挙することなく最適化を実施できることであり，この性質無しには計算が実行できない場合もある重要なものである．
First, this pruning enables us to train EIN without enumerating all the candidate subgraphs, which makes EIN computationally tractable. 
%
% 第二に，この枝刈りが解の質を保つことを簡単に立証できることである．
Second, our pruning strategy maintains the quality of the prediction compared with when we do not perform the pruning.  
%
% 近接勾配法の閾値処理で削除されるものを枝刈りするだけであるため，枝刈りしない場合と厳密に同じ結果が得られる．
This is because we only omit the computations that does not have any effect on the prediction.
%}

%\red{
% 我々の貢献は次のように要約される．
Our contributions are summarized as follows.
\begin{itemize}
 \setlength{\itemsep}{0pt}  
 \item % 部分グラフの厳密な存在の有無に基づく特徴量を用いるニューラルネットワークモデルEINを構成し，重要部分グラフをグループスパース正則化で選択する定式化を示す．
       We propose EIN, which is a neural network model that uses the exact subgraph isomorphism feature.
       %
       % 提案法は厳密な部分グラフ同定により高い表現能力を持ちつつ，重要部分グラフの情報を提供できる．
       Through a group sparse constraint, we formulate EIN so that a small number of subgraphs can be identified by which an insight about the important substructure can be extracted for the given graph-level prediction task. 
 \item % グループスパース正則化と部分グラフ列挙を組み合わせて，枝刈り機構付きの誤差逆伝播法を構成し，EINの最適化アルゴリズムを提案する．
       We show that the combination of the subgraph enumeration and IHT with backpropagation can derive an efficient pruning strategy that makes the EIN training computationally tractable.
       %
       % これにより，現実的に実行可能でありかつ，解の質を保証できる手法が構築できる．
       We further reveal that our pruning strategy does not degrade the prediction quality, and the resulting algorithm has at least sublinear convergence rate to a critical point.
 \item % 人工データやベンチマークデータにより，EINが標準的なグラフニューラルネットワークと同等以上の精度を持つことを示す．
       Based on synthetic and benchmark datasets, we demonstrate that EIN has superior or comparable performance compared with standard graph prediction models while EIN actually can identify a small number of important subgraphs, simultaneously. 
       %
       % 人工データでは，典型的なグラフニューラルネットワークが判別できないグラフがEINでは判別可能なことも示す．
\end{itemize}

% ==================================================
\section{Proposed Method: Exact Subgraph Isomorphism Network}
\label{sec:proposed-method}

% 本章では，提案手法 Exact subgraph isomorphism network (EIN)について説明する．
%
% EINではグラフ$G \in \cG$を入力とする分類問題を考える, where $\cG$ is a set of labeled graphs.
Our proposed method, called Exact subgraph Isomorphism Network (EIN), considers the classification problem of a graph $G \in \cG$, where $\cG$ is a set of labeled graphs.
%
% $G$のノードとエッジの集合をそれぞれ$V_G, E_{G}$と表記する．
%
% ノード$v \in V_{G}$はそれぞれ，カテゴリラベル$L_{v}$を持つ．
$G$ consists of a set of nodes and edges between nodes and each node can have a categorical label.
%
% 訓練データは
% $\{ (G_i, y_i) \}_{i \in [n]}$
% で， 
% $y_i \in \cY$
% $y_i \in \{ -1, 1 \}$
% はラベル，$n$はデータ数であり， $[n] = \{ 1, \ldots, n\}$とする．
The training data is 
$\{ (G_i, y_i) \}_{i \in [n]}$,
where
$y_i \in \cY$
is a graph label, $n$ is the number of instances, and 
$[n] = \{ 1, \ldots, n\}$.
%
% ここでは分類問題のみに着目するが，損失関数を置き換えるだけで回帰問題にも適用可能である．
Here, although we only focus on the classification problem, the regression problem can also be handled by just replacing the loss function.

% まず，\ref{ssec:formulation}節ではモデルの定式化と最適化問題の定義を示し，\ref{ssec:optimization}節でこの効率的な最適化アルゴリズムを示す．
% First, \SS~\ref{ssec:formulation} describes the model definition and the optimization formulation, and second, the detailed optimization procedure is shown in \S~\ref{ssec:optimization}. 
First, \S~\ref{ssec:formulation} describes the formulation of our model.
Second, the optimization procedure is shown in \S~\ref{ssec:optimization}. 
Next, \S~\ref{sssec:interpret-EIN} and \ref{ssec:combine-GNN} discuss post-hoc analysis for knowledge discovery and a combination of graph neural networks, respectively.

% --------------------------------------------------
% \subsection{定式化}
% \subsection{モデルの定義}
\subsection{Model Definition}
\label{ssec:formulation}

% \subsubsection{Subgraph Inclusion Score}
% \label{sssec:SIS}

% 分類境界に寄与する重要な小さな部分グラフの抽出を考える．
%
%また，EINにおける部分グラフは頂点ラベルのみを扱い，頂点属性値は利用しない．
% 入力グラフが部分グラフ$H$を含む度合いを表す特徴量を
% とし，
% $\psi(G_i; H)$
% をSubgraph Inclusion Scoreと呼ぶ．
%
% 入力グラフが部分グラフ$H$を含むかどうかを表す特徴量を
% $\psi_H(G_i) \in \{0,1\}$
% をsubgraph isomorphism feature (SIF)と呼び，以下のように定義する: 
Let 
$\psi_H(G_i) \in \{0,1\}$
be the feature that represents whether the input graph $G_i$ contains a subgraph $H$ (we only focus on a connected subgraph), which we call subgraph isomorphism feature (SIF): 
%とする．
% EINでは，この特徴を持つモデルの特徴選択を行うことで，重要な部分グラフを特定する．
%$L(G)$を各ノードがラベル$L_{v}$を持つラベル付きグラフとする．
% このとき，SISは式(\ref{eq:SIS})で定義される：
\begin{align}
  \psi_H(G_i) = \II( H \sqsubseteq G_i ), 
%   \begin{cases}
% %   1 & \text{if } L(H) \sqsubseteq L(G_i), \\
%    1 & \text{if } H \sqsubseteq G_i, \\
%    0 & \text{otherwise}, 
%   \end{cases}
  \label{eq:SIS}
 \end{align}
% ここで，$H \sqsubseteq G_i$は$H$が$G_i$の部分グラフであることを意味する．
where $\II$ is the indicator function and $H \sqsubseteq G_i$ indicates that $H$ is a subgraph of $G_i$. 
Note that although instead of 
$\II( H \sqsubseteq G_i )$,
frequency that $H$ is included in $G_i$ can also be used for $\psi_H(G_i)$ in our framework, we employ \eq{eq:SIS} throughout the paper for simplicity (see Appendix~\ref{app:SIF-frequency} for more detail).
%
% 例えば，Fig.~\ref{fig:overview}の左側では$G_1$は
% $H = \includegraphics[width=1em]{./figs-subgraph1.pdf}$
% で
% $\phi_H(G_1) = 1$，
% $H = \raisebox{-.3em}{\includegraphics[width=1em]{./figs-subgraph3.pdf}}$
% で
% $\phi_H(G_1) = 0$
% となる．
% For example, in the left side of Fig.~\ref{fig:overview}, 
For example, in Fig.~\ref{fig:overview}~(a), 
$G_1$ is $\psi_H(G_1) = 1$ for 
$H = \includegraphics[width=1em]{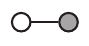}$
and
$\psi_H(G_1) = 0$ for 
$H = \raisebox{-.3em}{\includegraphics[width=1em]{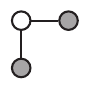}}$.
% 
%
% 本論文では解釈性のため，連結な部分グラフのみを扱う．
% For higher interpretability, we only focus on connected subgraphs.
%
% 候補となる部分グラフとして訓練データに含まれる，あるサイズ以下の全ての部分グラフを考える:
% $\cH = \{ H \mid H \sqsubseteq G_i, i \in [n], |H| \leq \text{maxpat}\}$．
We define candidate subgraphs as 
$\cH = \{ H \mid H \sqsubseteq G_i, i \in [n], |H| < \text{maxpat}\}$,
which is all the subgraphs in the training dataset whose size is at most pre-specified $\text{maxpat}$ (the size $|H|$ is the number of edges).
% $\text{maxpat}$ はユーザーが指定した部分グラフの最大サイズであり，\red{ここでは辺の数と定義する} ($\cH$の生成方法は後述する)．
%
% このとき，各部分グラフ$H \in \cH$についてSIF
% $\psi_H(G_i)$
% を並べた特徴ベクトル
% $\*\psi(G_i)$
% を定義できる．
By concatenating SIF $\psi_H(G_i)$ of each $H \in \cH$, the feature vector $\*\psi(G_i) \in \{ 0, 1 \}^{|\cH|}$ is defined. 
%
% EINでは，この特徴ベクトルを持つモデルの特徴選択を行うことで，重要な部分グラフを特定する．
EIN identifies a small number of important subgraphs from $\cH$ through the feature selection discussed later.
%
% 明らかに，SIF
% $\psi_H(G_i)$
% は部分グラフの存在について高い解釈性のある特徴量である．
SIF $\psi_H(G_i)$ is obviously highly interpretable and it can assure the existence of a subgraph $H$.
% 
% また，$G_i$と$G_j$は一つでも異なる部分グラフ$H \in \cH$を含んでいれば
% $\*\phi(G_i) \neq \*\phi(G_j)$
% となり$\cH$に含まれる部分構造に対する厳密な判別性を有する．
Further, if $G_i$ and $G_j$ contain at least one different subgraph $H \in \cH$, then, we have 
$\*\psi(G_i) \neq \*\psi(G_j)$. 
% that means SIF can exactly discriminate different subgraph structures contained in $\cH$.
%

% EINではFig.~\ref{fig:overview}に示す通り，Graph Mining Layer (GML)とFeed Forward Network (FFN)から構成される．
As shown in Fig.~\ref{fig:overview}~(b), EIN consists of Graph Mining Layer (GML) and Feed Forward Network (FFN). 
%
% EINでは最初のレイヤーとしてGraph Mining Layer (GML)を以下のように定義する
% GMLの出力ユニット数を$K$とする．
Let $K$ be the number of output units of GML.
%
% このとき，
% $\mr{GML}: \cG \rightarrow \RR^K$
% は以下のように定義される: 
We define 
$\mr{GML}: \cG \rightarrow \RR^K$
as follows.
\begin{align*}
 \mr{GML}(G_i ; \*B, \*b) &= \sigma (\*h),
 \\
 \*h &= \sum_{H \in \cH} \*\beta_H \psi_H(G_i) + \*b,
\end{align*}
% ただし，$\*B = [\*\beta_1, \dots, \*\beta_{|\cH|}] \in \RR^{K \times |\cH|}$と$\*b \in \RR^K$はパラメータ, $\sigma: \RR^K \rightarrow \RR^K$は次元ごとの活性化関数である．
where 
$\*B \in \RR^{K \times |\cH|}$ 
is a weight matrix whose columns consist of 
$\*\beta_H \in \RR^{K}$ 
for $H \in \cH$, 
$\*b \in \RR^K$
is a bias parameter, 
% $\*B = [\*\beta_1, \dots, \*\beta_{|\cH|}] \in \RR^{K \times |\cH|}$
% $\*b \in \RR^K$are parameters, 
and 
$\sigma: \RR^K \rightarrow \RR^K$
is an activation function (this time, we use the standard sigmoid function).  
%
% それぞれの$\*\beta_H$が部分グラフ$H$の存在に対応する表現になっている．
Each $\*\beta_H$ can be seen as a representation corresponding to each subgraph $H$.
%
% 予測モデル全体はは次のようにグラフ$G_i$を入力とするネットワークで定義される：
The entire prediction model is defined as 
\begin{align*}
 % f(G_i) &= \mathrm{MLP}_{\*\Theta} (\sigma (\*h)) 
 f(G_i) &= \mr{FFN} ( \mr{GML}(G_i ; \*B, \*b) ; \*\theta ),
\end{align*}
% ただし，
% $\Theta$
% は$\mr{FFN}$のパラメータである．
where
$\*\theta$
is parameters of FFN.
%
% 予測モデルは次のようにグラフ$G_i$を入力とするネットワークで定義される：
% \begin{align*}
%  % f(G_i) &= \mathrm{MLP}_{\*\Theta} (\sigma (\*h)) 
%  f(G_i) &= \mr{FFN} ( \mr{GML}(G_i ; \*B, \*b) ; \*\Theta ),
% \end{align*}
% ただし，
% \begin{align*}
%  \mr{GML}(G_i ; \*B, \*b) &= \sigma (\*h),
%  \\
%  \*h &= \sum_{H \in \cH} \*\beta_H \psi_H(G_i) + \*b,
%  % \label{eq:f}
% \end{align*}
% % ここで，
% % $\cH$は候補部分グラフの集合であり，
% \red{$\*B = [\*\beta_1, \dots, \*\beta_{|H|}], \*b, \Theta$はパラメータである．}
% $\*B$は一層目の中間ユニット数 $K$，候補部分集合の数$|\cH|$に対して$\*B \in \mathbb{R}^{K \times |\cH|}$ であり，
% $\*b$はバイアス項，$\sigma$は活性化関数，$\*\Theta$はFFNのパラメータである．
% また，
%
% さて$\*B, \*b, \*\Theta$ に対する最適化問題は，$\*\beta_{H}$にグループ正則化\red{\citep{}}を行なった以下の正則化損失最小化として定義される：
We use the cross entropy loss function by which our objective function is defined as 
$\ell(\*B, \*b, \*\theta) \coloneqq \sum_{i = 1}^n \mr{loss}(y_i, f(G_i) ; \*B, \*b, \*\theta)$. 
% 
%\ell(\*B^t, \*b^t, \*\theta^t)
%
We optimize the parameters $\*B, \*b$, and $\*\theta$ through the following optimization problem in which a group sparsity constraint is imposed on $\*B$: 
\begin{align}
 \begin{split}  
  \min_{\*B, \*b, \*\theta} \ &
  \ell(\*B, \*b, \*\theta)
  % \sum_{i = 1}^n \ell(y_i, f(G_i))
  % + \lambda \sum_{H \in \cH} \| \*\beta_H \|_2
  \\ % ---
  \text{s.t.} \ & 
  \| \*B \|_{0,2} \leq s,
 \end{split}
 \label{eq:objective}
\end{align}
where 
\begin{align*}
 \| \*B \|_{0,2} 
 = | \{ H \mid \| \*\beta_H \|_2 \neq 0 \} |  
\end{align*}
is the mixed $L_{0,2}$ norm, and $s > 0$ is the constant parameter. 
% , and $\ell$ is a differentiable loss function (we here use the cross-entropy loss).
%
% 制約条件は$\*B$がgroup-wiseでたかだか$s$個しか非ゼロ要素を持たないことを意味する．
The constraint in \eq{eq:objective} indicates that $\*B$ only has $s$ non-zero columns.
%
% これは言い換えるとたかだか$s$個のsubgraphしか選ばないことを意味する．
In other words, this means that only $s$ subgraphs are used in $f$, by which we can identify important predictive subgraphs as
$\{ H \mid \*\beta_H \neq \*0 \}$.
%
% \blue{(ToDo: re-write loss.)}
% We optimize the parameters $\*B, \*b$, and $\*\Theta$ through the following optimization problem in which the group sparse penalty \citep{yuan2006model} is imposed on $\*\beta_{H}$:
% \begin{align}
%   \min_{\*B, \*b, \*\Theta} \
%   \sum_{i = 1}^n
%   \ell(y_i, f(G_i))
%   + \lambda \sum_{H \in \cH} \| \*\beta_H \|_2
%   \label{eq:objective}
% \end{align}
% ここで$L$は微分可能な損失関数である．
% where $\ell$ is a differentiable loss function and $\lambda$ is a regularization parameter.
%
% ここで$L$は凸で微分可能な損失関数である．
%
% 本論文では，ロジスティック損失関数を採用する．
% 本論文では，cross-entropy損失関数を採用する．
% We here use the cross-entropy loss for $\ell$.
% ：
% \begin{align*}
% \ell(y_i, f(G_i)) = \log \left( 1 + \exp(-y_i f(G_i)) \right)
% \end{align*}
%
% 目的関数\eq{eq:objective}は $\*\beta_H$ に対してスパースな解を導くため，少数の重要な部分グラフは$\*\beta_H \neq \*0$となる$H$として表される．
% The group-wise penalty 
% $\| \*\beta_H \|_2$
% results in 
% $\*\beta_H = \*0$ 
% for many unnecessarily $H$ at the solution of \eq{eq:objective}, by which we can identify important predictive subgraphs as
% $\{ H \mid \*\beta_H \neq \*0 \}$.
%
% しかし，この最適化問題では，最適化変数が膨大になってしまう．
% しかし，この最適化問題では$\*B$のサイズが
% $K \times |\cH|$
% であり，
% $|\cH|$
% は部分グラフの総数であるため膨大になってしまう．
However, since the size of $\*B$, i.e., $K \times |\cH|$, is quite large, na{\"i}ve optimization of \eq{eq:objective} can be difficult.
% %
% よってこの問題を軽減する効率的な最適化アルゴリズムを提案する．

% --------------------------------------------------
\subsection{Optimization}
\label{ssec:optimization}

% 我々の最適化アルゴリズムは，（近接）勾配降下が変数のブロックを交互に更新するブロック座標更新 \citep{xu2017globally} アルゴリズムをベースとする．
% 我々は変数を交互に更新する方式を採用する．
Our optimization algorithm is based on the block coordinate descent, in which each one of $\*B, \*b$, and $\*\theta$ are updated alternately while the other two parameters are fixed.
%
% $\*B, \*b, \Theta$を交互に更新するが，残りの2つのパラメータは固定して扱う．
%
% まず，$\*B$はグループ正則化のペナルティがあるため，近接勾配更新を適用する．
Since $\*B$ has the group sparse constraint, we apply iterative hard thresholding (IHT) to $\*B$ \citep[e.g.,][]{beck2013sparsity,bao2014ell0,yuan2018gradient,damadi2022gradient}.
% we apply the well-known proximal gradient method \citep{bech2009fast,parikh2014proximal}.
%
% 次に，$\*b$，$\*\Theta$についてはスパースペナルティがないため，それぞれ通常の勾配降下更新を適用する．
On the other hand, $\*b$ and $\*\theta$ have no sparse penalty and we can simply apply the standard gradient descent.

% 以下の繰り返し
We iterate the following procedure, in which $t$ is the number of iterations:
\begin{enumerate}
 \item Update $\*B$ through the hard thresholding (HT), defined as
       \begin{align} 
	\begin{split}	 
	\*B^{t+\frac{1}{2}} &= \*B^t - \gamma^t_{\*B} \nabla_{\*B} \ell^t 
	% \*B^{t+\frac{1}{2}} &= \*B^t - \gamma^t_{\*B} \nabla_{\*\beta_H} \ell^t % \pd{\ell(\*B^t, \*b^t, \*\theta^t)}{\*B},
	% \nabla_{\*B} \ell^t,
	 \\ % ---
	\*B^{t+1} &= H_s( \*B^{t+\frac{1}{2}} ),
	% \*B^{t+1} = H_s(\*B^t - \gamma^t_{\*B} \nabla_{\*B} \ell^t ) 
	\end{split}
	\label{eq:update-B}
       \end{align}
       where 
       % $\nabla_{\*\beta_H} \ell^t = \pd{\ell(\*B^t, \*b^t, \*\theta^t)}{\*\beta_H}$ 
       $\nabla_{\*B} \ell^t = \pd{\ell(\*B^t, \*b^t, \*\theta^t)}{\*B}$ 
       is the gradient of $\*B$, 
       $\gamma^t_{\*B}$
       is a step length, and 
       \begin{align*}
	H_s(\*B) = 
	\argmin_{\| \*Z \|_{0,2} \leq s}
	\| \*Z - \*B \|_F
       \end{align*}
       is the HT operator. 
       The step length parameter $\gamma^t_{\*B}$ is determined by the backtrack search \citep{nocedal2006numerical}. 
 \item Update $\*b^t$ and $\*\theta^t$ to $\*b^{t+1}$ and $\*\theta^{t+1}$, respectively, by using the standard gradient descent. 
\end{enumerate}
%
% \red{The computation of $\nabla_{\*\beta_H} \ell^t$ for single $H$.}
%
Since $\cH$ contains a large number of subgraphs, calculating \eq{eq:update-B} for the entire $\*B \in \RR^{K \times |\cH|}$ is not directly tractable. 
In \S~\ref{sssec:pruning}, we derive a pruning rule by which we can perform \eq{eq:update-B} without enumerating all the subgraphs in $\cH$.
\S~\ref{sssec:algorithm} describes the entire procedure, and \S~\ref{sssec:convergence-analysis} provides a theoretical convergence analysis.

% --------------------------------------------------
\subsubsection{Backpropagation with Gradient Pruning}
\label{sssec:pruning}

By combining a graph mining algorithm and the HT operation, we can build an efficient pruning algorithm. 
%
%効率的なグラフマイニングアルゴリズムとして有名なgSpan\citep{yan2002gspan}を採用する．
We employ a well-known graph mining algorithm called gSpan \citep{yan2002gspan} that can efficiently enumerate all the subgraphs in a given set of graphs.
%
% \figurename~\ref{fig:mining-tree}は，グラフマイニング木の説明図である．
Figure~\ref{fig:overview}~(c) is an illustration of a graph mining tree.
%
% gSpanは木ノード内の各グラフを，拡張したグラフが，部分グラフとして与えられたグラフ集合に含まれる限り，再帰的に拡張して木を作成し重複のない一意なenumerationが可能である (detailはsee gSpan)．
At each tree node, gSpan expands the graph (add an edge and a node) as far as the expanded graph is included in the given dataset as a subgraph. 
By providing a unique code to each generated graphs (called the DFS code because it is based on depth-first search in a graph), gSpan can enumerate all the subgraphs without generating duplicated graphs (For further detail, see \citep{yan2002gspan}).
%
% \figurename~\ref{fig:mining-tree}の重要な特徴は，全てのグラフはその祖先の任意のグラフを部分グラフとして含むということである．
As shown in Fig.~\ref{fig:overview}~(c), all the graphs in the tree contains their ancestors as a subgraph.

% 以下の書き換えから$H_s$は単に列のノルムが大きい$s$列を選び他を$0$にすることに相当することは明らか
It is easy to see that the HT operator simply selects the columns with the top-$s$ largest norm because
\begin{align*}
 % \argmin_{\| \*Z \|_{2,0} \leq s} \| \*Z - \*B \|_F
 H_s(\*B) 
 % \\ % ---
 &=
 \argmin_{\| \*Z \|_{2,0} \leq s}
 \| \*Z - \*B \|_F^2
 % \\ % ---
 % &=
 =
 \argmin_{\| \*Z \|_{2,0} \leq s}
 \sum_{k = 1}^{| \cH |} \| \*Z_k - \*B_k \|_2^2,
 % \sum_{k = 1}^{| \cH |} \| \*Z_k - \*\beta_H \|_2^2.
 % \\ % ---
 % &=
 % \sum_{i=s+1}^{ |\cH| } \| \*B_{k_i} \|_2^2, 
\end{align*}
where $\*B_k$ and $\*Z_k$ is the $k$-th column of $\*B$ and $\*Z$, respectively.
Let
% $k_1, \ldots, k_s, k_{s+1}, \ldots, k_{| \cH |}$
$H_1, \ldots, H_{| \cH |}$
be the subgraphs such that 
$\| \*\beta^{t+\frac{1}{2}}_{H_1} \|_2 > \| \*\beta^{t+\frac{1}{2}}_{H_2} \|_2 > \cdots > \| \*\beta^{t+\frac{1}{2}}_{H_{|\cH|}} \|_2 $, 
where 
$\*\beta^{t+\frac{1}{2}}_{H}$
is a column of $\*B^{t+\frac{1}{2}}$.
Then, $\*\beta^{t+1}_H$, which is a column of $\*B^{t+1}$, is written as
\begin{align*}
 \*\beta^{t+1}_H = 
 \begin{cases}
  \*\beta^{t+\frac{1}{2}}_H & \text{ if } H \in \{H_1, \ldots, H_s\}, \\
  \*0 & \text{ otherwise. }
 \end{cases}
\end{align*}
%
% つまり，Miningのツリーを探索しながら，
% $\| \*\beta^{t+\frac{1}{2}}_H \|_2$
% の大きな$s$列を見つければよい．
This means that we only need to identify $H$ with the top-$s$ largest 
$\| \*\beta^{t+\frac{1}{2}}_H \|_2$
during the graph mining tree traverse because $\*\beta^{t+1}_H$ for the other $H$ is $\*0$.

% Iteration $t$で選ばれている$H$の集合を$S_t$とする．
Let $S_t$ be the set of selected $H$ at the iteration $t$.
%
% $| S_t | = s$なので，$H \in S_t$に対して$\*\beta^{t+\frac{1}{2}}_H$を計算することは容易い．
Since 
$| S_t | = s$, which is assumed to be a small value (in our later experiments, $s \leq 100$), it is not computationally difficult to calculate 
$\*\beta^{t+\frac{1}{2}}_H$
for 
$H \in S_t$. 
%
% $H \notin S_t$が多いので，$\*\beta^{t+\frac{1}{2}}_H$を全て計算するのは不可能で$s$位に入る可能性のないものを除外(Pruning)する必要がある．
On the other hand, we have an intractable number of 
$H \notin S_t$.
%
% because of which, from the mining tree, we remove $H$ that does not have a possibility of being included in $S_{t+1}$.
Therefore, we remove $H$ that does not have a possibility of being included in $S_{t+1}$.
% from the mining tree
%
% \red{Let $\nabla_{\*\beta_H} \ell^t = \pd{\ell(\*B^t, \*b^t, \*\theta^t)}{\*\beta_H}$}.
%
% $H \notin S_t$
% ならば
% $\*\beta_H^{t+\frac{1}{2}} = \gamma_{\*B}^t \nabla_{\*\beta_{H}} \ell^t$
% であるので，
% $\| \nabla_{\*\beta_{H}} \ell^t \|_2$
% が大きいものを$s$個選出する．
Since
$\*\beta^t_H = \*0$ 
for 
$H \notin S_t$,
if 
$H \notin S_t$,
we see 
% $\*\beta_H^{t+\frac{1}{2}} = \gamma_{\*B}^t \nabla_{\*\beta_{H}} \ell^t$
$\| \*\beta_H^{t+\frac{1}{2}} \|_2 = \gamma_{\*B}^t \| \nabla_{\*\beta_{H}} \ell^t \|_2$, 
where 
$\nabla_{\*\beta_{H}} \ell^t = \pd{\ell(\*B^t, \*b^t, \*\theta^t)}{\*\beta_H}$.
To define our pruning strategy, the following upper bound of 
$\| \nabla_{\*\beta_{H}} \ell^t \|_2$
plays a key role:
% --------------------------------------------------
% Theorem: Upper bound of gradient 
% --------------------------------------------------
\begin{theo} \label{th:pruning}
 Let 
 $H^\prime \sqsupseteq H$ % \red{($H \notin S_t$いる？)}
 % for 
 % $H, H^\prime \in \overline{\cW}$, 
 and 
 % $\delta_{ik} = \pd{\ell(y_i, f(G_i))}{h_k}$.
 $\delta^t_{ik} = \pd{\mr{loss}(y_i, f(G_i) ; \*B^t, \*b^t, \*\theta^t)}{h_k}$.
 Then, %we have
 \begin{align}
  % \| \*g_{H^\prm} \|_2 \leq 
  % \red{\zeta_{H^\prm}} = 
  \| \nabla_{\*\beta_{H^\prm}} \ell^t \|_2 \leq &
  \biggl\{ \sum_{k=1}^K 
  \max \bigl\{
  \bigl( \sum_{ i \in \{ i \mid \delta^t_{ik} > 0 \} }
  \delta^t_{ik} \psi_H(G_i) \bigr)^2, 
  \notag \\ % ---
  &  \ \
  \bigl( \sum_{ i \in \{ i \mid \delta^t_{ik} < 0 \} }
  \delta^t_{ik} \psi_H(G_i) \bigr)^2 \bigr\}
  \biggr\}^{1/2}
  \eqqcolon
  \mr{UB}(H).
  \label{eq:grad-UB}
 \end{align}
\end{theo}
\noindent
We here only describe the sketch of the proof (the proof is in Appendix~\ref{app:proof-pruning}).
An essential idea is that we expand
% $\*g_H$
$\nabla_{\*\beta_{H}} \ell^t$
as a linear combination of the derivatives with respect to the GML intermediate representation $h_k$, i.e., $\delta^t_{ik}$, by which the $k$-th element of the gradient can be written as 
% $(\*g_H)_k = \sum_{i = 1}^n \delta_{ik} \psi_H(G_i)$. 
$( \nabla_{\*\beta_{H}} \ell^t )_k = \sum_{i = 1}^n \delta^t_{ik} \psi_H(G_i)$. 
By using 
the monotonically non-increasing property of 
$\psi_H(G_i)$,
i.e., 
$\psi_{H^\prm}(G_i) \leq \psi_H(G_i)$ 
if 
$H \sqsubseteq H^\prm$, 
we can derive ${\rm UB}(H)$. 
%
% 通常の最適化計算でこの展開を明示的に意識する必要性はないため，これは意図的にこの形に帰着することで初めて導出できるものです
Note that, usually, there is no need to explicitly consider this specific expansion of 
% $(\*g_H)_k$; 
$( \nabla_{\*\beta_{H}} \ell^t )_k$; 
the theorem can only be derived by deliberately reducing it to the linear combination of $\delta^t_{ik}$.

Theorem~\ref{th:pruning} indicates that, for any 
$H^\prime$
that contains $H$ as a subgraph, the $L_2$ norm of the gradient 
% $\| \*g_{H^\prm} \|_2$
$\| \nabla_{\*\beta_{H^\prm}} \ell^t \|_2$
can be bounded by 
$\mr{UB}(H)$. 
%
% なお，$H^\prime$を生成せずに$\mathrm{UB}(H)$を算出でき，現在のパラメータ $\*B, \*b, \*\Theta$によるモデル予測および誤差逆伝播による勾配$\delta$，$\psi_H(G_i)$だけが必要となる．
Note that the upper bound 
$\mr{UB}(H)$ 
can be calculated without generating 
$H^\prm$.
%
% It only uses the current parameters $\*B, \*b$, and $\*\Theta$
%
% 上の規則~\eq{eq:gradient-condition}と定理~\ref{th:pruning}から以下の重要な関係が得られる．
% From the rule \eq{eq:gradient-condition} and Theorem~\ref{th:pruning}, we can immediately obtain the following important rule:
From Theorem~\ref{th:pruning}, we can immediately obtain the following pruning rule with a threshold parameter $\eta$:
\begin{coro}
 \label{coro:pruning-rule}
 % \red{$\zeta^{(1)}$}
 % \red{$\zeta$}
 % をある$H$に辿り着くまでの木探索の過程で計算した
 % $\| \nabla_{\*\beta_{H}} \ell^t \|_2$
 % % $\zeta_H$
 % for 
 % $H \notin S_t$
 % のうち$s$番目の大きさの値とする．
 Let 
 $\zeta \leq \| \*\beta^{t+\frac{1}{2}}_{H_s} \|_2 / \gamma_{\*B}^t$, 
 meaning that 
 $\zeta$
 is less than or equal to the $s$-th largest value of
 % $\| \*\beta^{t+\frac{1}{2}}_{H_s} \|_2 / \gamma_{\*B}^t$.
 $\| \*\beta^{t+\frac{1}{2}}_{H} \|_2 / \gamma_{\*B}^t$ 
 among 
 $H \in \cH$.
 % for  $H \in \cH$. 
 % among $H$ of the nodes already visited in the mining tree at that iteration.
 %
 % For $\forall H^\prime \in \{ H^\prime \mid H^\prime \sqsupseteq H, H^\prime \in \overline{\cW} \}$, 
 For 
 $\forall H^\prime \in \{ H^\prime \mid H^\prime \sqsupseteq H, \ H^\prm \notin S_t \}$, 
 \begin{align}  
  \mr{UB}(H) < \zeta
  % \lambda
  %  \text{ and } H \in \overline{\cW}	 
  \ \ \Rightarrow \ \
  % \mathrm{prox} \left( \*\beta_{H^\prm} - \eta \ \*g_{H^\prm} \right) = \*0	
  \*\beta_{H^\prm}^{t+1} = \*0
  % \\
  % & \qquad \text{ for } 
  % \forall H^\prime \in \{ H^\prime \mid H^\prime \sqsupseteq H, H^\prime \in \overline{\cW} \}.	
  \label{eq:pruning-zeta}
 \end{align} 
\end{coro}
\noindent
This is a direct consequence from the fact that if the conditions in \eq{eq:pruning-zeta} holds, we have 
% $\| \*g_{H^\prime} \|_2 \leq \mr{UB}(H) \leq \lambda$
\begin{align*}
 \| \*\beta^{t+\frac{1}{2}}_{H^\prm} \|_2 
 = \gamma_{\*B}^t \| \nabla_{\*\beta_{H^\prm}} \ell^t \|_2 
 \leq \gamma_{\*B}^t \mr{UB}(H) 
 < \gamma_{\*B}^t \zeta
 \leq \| \*\beta^{t+\frac{1}{2}}_{H_s} \|_2
\end{align*}
from \eq{eq:grad-UB}. %Theorem~\ref{th:pruning}. 
%
% As a result, by using \eq{eq:gradient-condition}, we obtain \eq{eq:pruning-zeta}. 
As a result, we see that 
$\| \*\beta^{t+\frac{1}{2}}_{H^\prm} \|_2 $
% $\| \nabla_{\*\beta_{H^\prm}} \ell^t \|_2$
cannot be ranked in the top-$s$ largest values.
%
% Corollary~\ref{coro:pruning-rule} means that if the condition in \eq{eq:pruning-zeta} holds, 
Therefore, we can omit the update for $\forall H^\prime$ that contains $H$ as a subgraph if  
$\mr{UB}(H) < \zeta$
holds. 
% \red{(ToWrite: About step length. where?)}

% \red{$\zeta$}をどう決めるか．
To use the pruning rule Corollary~\ref{coro:pruning-rule}, we need to determine $\zeta$.
% 
% Step幅$\gamma_{\*B}^t$が事前に固定されている場合は簡単．
We first consider the case that the step length $\gamma_{\*B}^t$ is fixed beforehand, in which a simple approach can be applied.
% A simple approach can be taken if the step length $\gamma_{\*B}^t$ is fixed beforehand.
%
% まず$\| \*\beta_H^{t+\frac{1}{2}}\|$を$H \in S_t$ に対し計算．
First, 
$\| \*\beta_H^{t+\frac{1}{2}}\|$
is calculated for all
$H \in S_t$.
%  \blue{(勾配の計算コストが$|\cH|$に依存してないことをどこかで言うべき？$H$を展開すればその$H$に関する勾配が計算できること)}
%
% 次に，mining treeを探索しながら各木頂点での$\| \*\beta_H^{t+\frac{1}{2}}\|$を一つずつ計算．
After that, the graph mining tree is traversed during which each one of
$\| \*\beta_H^{t+\frac{1}{2}} \|$
is calculated for the subgraph $H$ of each visited tree node. 
Suppose that $\cH_{\rm visited}$ represents the union of $S_t$ and a set of $H$ corresponding to the already visited tree nodes.
% 探索の過程で計算された$\| \*\beta_H^{t+\frac{1}{2}}\|$のうち$s$番目に大きなものを使って\red{$\zeta$}と設定すると，必ず\red{$\zeta \leq \| \*\beta^{t+\frac{1}{2}}_{H_s} \|_2 / \gamma_{\*B}^t$}を満たす．
%
% \blue{(すでに訪れたnodeのHの集合？$\cH_{\rm visited}$)}
The threshold of the pruning rule \eq{eq:pruning-zeta} can be set as 
% $\| \*\beta_H^{t+\frac{1}{2}}\| / \gamma_{\*B}^t$ 
% $\| \*\beta_H^{t+\frac{1}{2}}\| / \gamma_{\*B}^t$ 
$\zeta = s\operatorname{-max}\{ \| \*\beta_H^{t+\frac{1}{2}}\| / \gamma_{\*B}^t \}_{H \in \cH_{\rm visited}}$,
where 
$s\operatorname{-max}$
represents the $s$-th largest value,
% by using the $s$-th largest
% $\| \*\beta_H^{t+\frac{1}{2}}\|$ 
% among $H \in \cH_{\rm visited}$, 
% among subgraphs $H$ of already visited nodes, 
by which we can guarantee 
$\zeta \leq \| \*\beta^{t+\frac{1}{2}}_{H_s} \|_2 / \gamma_{\*B}^t$. 
%\blue{(途中の$s$番目は本当の$s$番目より小さい)}.
%
% このようすることで，木探索の過程でその時点の\red{$\zeta$}に対して，$\mr{UB}(H) \leq \red{\zeta}$が成立した時点で，その下の頂点全てを枝刈りできる．
As a result, if 
$\mr{UB}(H) < \zeta$
holds for the threshold 
$\zeta$
at any node during the tree traverse, we can discard the entire subtree below the current node.
%
% しかし，backtrackを行う場合は$\gamma_{\*B}^t$が更新前に決まらず$\| \*\beta_H^{t+\frac{1}{2}} \|_2$が計算できない．
However, when we use a backtrack strategy for the step length, we cannot calculate 
$\| \*\beta_H^{t+\frac{1}{2}} \|_2$
before the update of $\*B^t$ because $\gamma_{\*B}^t$ is not determined yet. 
%
% 次にこの時にも\red{$\zeta$を設定する方法を述べる．}
% Next, we show that an effective threshold \red{$\zeta$} still can be defined in this case.

% \red{$\zeta \leq \| \*\beta^{t+\frac{1}{2}}_{H_s} \|_2 / \gamma_{\*B}^t$}
% の条件を満たしつつ
% $\gamma_{\*B}^t$
% を使わない閾値として
% % which satisfies \red{$\zeta \leq \| \*\beta^{t+\frac{1}{2}}_{H_s} \|_2 / \gamma_{\*B}^t$}
Instead, we can also build a threshold $\zeta$ without using 
$\gamma_{\*B}^t$ 
as follows.
\begin{align}
 % s\operatorname{-max}_{H \in \cH_{\rm visited}}\{ \| \*\beta_H^{t+\frac{1}{2}}\| \} / \gamma_{\*B}^t  
 &
 s\operatorname{-max} \{ \ \| \*\beta_H^{t+\frac{1}{2}} \|_2 / \gamma_{\*B}^t \ \}_{H \in \cH_{\rm visited}} 
 \notag \\ % ---
 &=
 s\operatorname{-max}
 \cbr{ 
 \{ 
 \frac{1}{\gamma_{\*B}^t}
 \| \*\beta_{H}^t - \gamma_{\*B}^t \nabla_{\*\beta_{H}} \ell^t \|_2
 \}_{H \in S_t}
 \cup
 \{ \| \nabla_{\*\beta_{H}} \ell^t \|_2 \}_{H \in \cH_{\rm visited} \setminus S_t}
 }
 \notag \\ %---
 &\geq 
 s\operatorname{-max} %_{H \in \cH_{\rm visited}}
 \cbr{ \eta_H }_{H \in \cH_{\rm visited}},
 \label{eq:zeta}
 % \cbr{ 
 % \{
 % \min_{\gamma \in \Gamma}
 % \frac{1}{\gamma}
 % \| \*\beta_{H}^t - \gamma \nabla_{\*\beta_{H}} \ell^t \|_2
 % \}_{H \in S_t}
 % \cup
 % \{ \| \nabla_{\*\beta_{H}} \ell^t \|_2 \}_{H \in \cH_{\rm visited} \setminus S_t}
 % }, 
\end{align}
where 
\begin{align*}
 \eta_H 
 \coloneqq 
 \begin{cases}
  \min_{\gamma \in \Gamma}
  \frac{1}{\gamma}
  \| \*\beta_H^t - \gamma \nabla_{\*\beta_H} \ell^t \|_2
  & \text{ if }
  H \in S_t,
  \\ % ---
  \| \nabla_{\*\beta_H} \ell^t \|_2
  & \text{ if }
  H \notin S_t, 
 \end{cases}
\end{align*}
and $\Gamma$ is a candidate space of the step length $\gamma_{\*B}^t$.
We use this lower bound \eq{eq:zeta} as $\zeta$.
% これは$H \in S_t$とそれ以外に分けて考えている．
%
In the derivation of \eq{eq:zeta}, $H \in S_t$ and other $H \in \cH_{\rm visited} \setminus S_t$ are separately considered.
%
% $H \notin S_t$の場合（），
% $\gamma_{\*B}^t$
% が消えるためステップ幅に依存しない形でそのまま計算ができる．
For 
$H \in \cH_{\rm visited} \setminus S_t$ (i.e., $H \notin S_t$), 
$\gamma_{\*B}^t$ is vanished in $\eta_H$, and therefore, the calculation does not depend on the step length. 
%
% $H \in S_t$の場合はステップ幅が残るため，$\min$をとっている．
On the other hand, for 
$H \in S_t$, 
since the step length cannot be directly removed, we derive a lower bound with respect to possible step lengths.
%
% この$\min$は簡単に計算できる(Appendix)．
This lower bound 
$\min_{\gamma \in \Gamma}$ 
can be easily calculated (See Appendix~\ref{app:lb-step-length} for detail). 
%
% \red{(ToWrite?:$\zeta$ is $s\operatorname{-max} %_{H \in \cH_{\rm visited}}
%  \cbr{ \eta_H }_{H \in \cH_{\rm visited}}$)}

% --------------------------------------------------
% Alg: Regularization path
% --------------------------------------------------
\begin{algorithm}[t] % [H]
 \caption{Optimization for EIN} \label{alg:regularization-path}
 % \Function(Train-EIN{(}$\Lambda, \*B, \*b, \*\Theta, \cW${)}){
 \Function(Train-EIN{(}$\cS${)}){
  $\*B \leftarrow \*0$ \\
  Initialize $\*b$ and $\*\theta$ randomly \\
  $H_0 \leftarrow \text{a graph at root of mining tree}$ \\
  \For{$s$ {\rm in} $\cS$}{
  \Repeat{ {\rm terminate condition met} }{
  Calculate $\eta_H$ for $\forall H \in S_t$ \\
  % $\min_{\gamma \in \Gamma}
  % \frac{1}{\gamma}
  % \| \*\beta_{H}^t - \gamma \nabla_{\*\beta_{H}} \ell^t \|_2$  
  % $\cH_{\rm visited} \leftarrow S_t$ \\
  $S_{t+\frac{1}{2}} \leftarrow S_t$ \\
  $\zeta \leftarrow s\operatorname{-max} \cbr{ \eta_{H} }_{H \in S_{t+\frac{1}{2}}}$\\
  $S_{t+\frac{1}{2}} \leftarrow \mr{Traverse}(H_0, s, \zeta, S_{t+\frac{1}{2}})$\\ 
  % $S_{t+\frac{1}{2}} \leftarrow \mr{Traverse}(H_0, s, S_{t+\frac{1}{2}}, \cH_{\rm visited})$\\ % $\cW \leftarrow \cW \cup \mr{Traverse}(H_0, ~\lambda)$\\  
  $\*B^{t+1}_{ S_{t+\frac{1}{2}} } \leftarrow H_s(\*B_{ S_{t+\frac{1}{2}} }^t - \gamma_{\*B}^t \nabla_{\*B_{ S_{t+\frac{1}{2}} } } \ell^t )$, \linebreak 
  where $\gamma_{\*B}^t$ is determined by backtrack search \\ 
  $\*b^{t+1} \leftarrow \mr{GradientDescent}(\*b^t)$\\
  $\*\theta^{t+1} \leftarrow \mr{GradientDescent}(\*\theta^t)$ 
    }
   } 
  }
\end{algorithm}
% --------------------------------------------------
% Alg: Pruning
% UB(H) >= eta_H > zetaなら UB(H) <= zetaはありえない
% --------------------------------------------------
\begin{algorithm}[t] % [H]
  % \caption{Working Set Selection} \label{alg:gpruning}
  \caption{Traverse Mining Tree} \label{alg:gpruning}
     \Function(Traverse{(}$H, s, \zeta, S_{t+\frac{1}{2}}${)}){
             % $\mathcal{W} \leftarrow \emptyset$\\
             % $\zeta \leftarrow s\operatorname{-max} \cbr{ \eta_{H^\prm} }_{H^\prm \in S_{t+\frac{1}{2}}}$ \\
             Calculate $\eta_H$ and $\mr{UB}(H)$ for current $H$ \\ % $\| \nabla_{\*\beta_{H}} \ell^t \|_2$ \\
             \uIf{$\mr{UB}(H) < \zeta$}{
                 \Return{$S_{t+\frac{1}{2}}$} \Comment{Pruning}
                 % \Return{$\emptyset$} \Comment{Gradient pruning}
             }
             \ElseIf{$\eta_H > \zeta$}{ 
                 $S_{t+\frac{1}{2}} \leftarrow S_{t+\frac{1}{2}} \cup \{ H \}$  \\
                 $\zeta = s\operatorname{-max} \cbr{ \eta_{H^\prm} }_{H^\prm \in S_{t+\frac{1}{2}}}$\\
                 $S_{t+\frac{1}{2}} \leftarrow S_{t+\frac{1}{2}} \setminus \{ H^\prm \mid \eta_{H^\prm} < \zeta, \ H^\prm \notin S_t\}$ 
             }
             % \textbf{if } $\| \*g_H \|_2 > \lambda$ \textbf{ then } $\cW \leftarrow \cW \cup \{ H \}$  \\
             $\cC \leftarrow \mr{Expand}(H)$\\
             \For{$H^\prime \in \cC$}{
                 $S_{t+\frac{1}{2}} \leftarrow \mr{Traverse}(H, s, \zeta, S_{t+\frac{1}{2}})$
             }
             \Return{$S_{t+\frac{1}{2}}$}
         }
     \Function(Expand{(}$H${)}){
             \uIf{ {\rm children of $H$ have never been created by gSpan} }{
                 $\cC \leftarrow$ all graphs expanded from $H$ by gSpan\\
                 Set $\{ \psi_{H^\prime}(G_i) \}_{i=1}^n$ \textbf{ for } $H^\prime \in \cC$                 \;
%                  \For{ $H^\prime \in \cC$ }{
%                      Set $\{ \psi_{H^\prime}(G_i) \}_{i=1}^n$ based on gSpan
% %                 }
              }
              \Else{
                 $\cC \leftarrow$ Retrieve already expanded children of $H$ \;
              }
              \Return{$\cC$}
         }
\end{algorithm}

As a result, the following important consequence is obtained
\begin{rema}
% gSpanによる木の探索の際に，現在の木のノードで部分グラフ$H$に対して$\mr{UB}(H) \leq \lambda$が成立すれば，全ての子孫ノードを枝刈りできる．
By evaluating 
$\mr{UB}(H) \leq \zeta$
during the tree traverse of gSpan (e.g., depth-first search), all the descendants of $H$ in the tree can be pruned if the inequality holds.
%
% つまり， $\overline{\cW}$の全要素を調べなくとも，\eq{eq:update-W}で$\cW$を更新できるということである． 
This means that we can perform the update \eq{eq:update-B} without enumerating all the possible subgraphs
$\cH$.
%
% In other words, we do not need to enumerate all the subgraphs $\cH$ for \eq{eq:update-B}.
\end{rema}
%
% さらに，以下も強調すべき特徴となる．
% Further, this immediately indicates the following remark.   
\noindent
Further, another notable remark about the pruning is as follows.   
\begin{rema}
% また，この操作は0になる$\*\beta_H$をスキップするだけなので結果に影響を与えない．
Since our pruning strategy only omit the update of unnecessarily parameters, i.e., $\*\beta_H = \*0$, the pruning does no change any parameter values compared with the case that we do not use the pruning.
%
% つまりpruningしない場合と同じ解となる
This also means that the final prediction performance also does not change because of the pruning. 
\end{rema}

% --------------------------------------------------
\subsubsection{Algorithm}
\label{sssec:algorithm}

% ここでは，EINの最適化の全手順を説明する．
We here describe the optimization procedure of EIN.
%
% 正則化パラメータ$\lambda$が大きな値の場合，解がよりスパースになり比較計算が容易となる．
% When $s$ is set as a small value, $\*B$ becomes highly sparse, by which computations usually become faster, because more subgraphs are expected to be pruned.
%
% Therefore, after starting from a smaller value of $s$, we gradually increase $s$ while optimizing parameters. 
The basic strategy is that we first set $s$ as a small value (e.g., $1$) and gradually increase $s$ while optimizing parameters. 
Let 
$\cS = ( s_1, \ldots, s_K )$
be a sequence of $s$, where 
$s_1 < s_2 < \cdots < s_K$. 
For each $s$, we use the solution of previous $s$ as an initial solution.

The Train-EIN function of Algorithm~\ref{alg:regularization-path} performs the optimization \eq{eq:objective} for each $s$.
%
% 5行目でgraph miningによるtree探索を実行して，gradient pruningにより$\cW$を作る．
In line 10, the Traverse function performs the graph mining tree search that generates a set of candidate subgraphs denoted as 
$S_{t+\frac{1}{2}}$,
consisting of subgraphs that are not pruned (i.e., $S_{t+1} \subseteq S_{t+\frac{1}{2}}$).
%
% この過程は\ref{alg:gradient-pruning}に示す．
The detailed procedure of the candidate generation is in Algorithm~\ref{alg:gpruning}.
%
% この関数は，グラフマイニング木を再帰的に探索する．
The Traverse function recursively searches the graph mining tree. 
%
% 木の各ノードでは，まず，可能であれば部分木を枝刈りするために，$\mathrm{UB}(H)$を評価する．
At each tree node, 
$\mr{UB}(H)$ 
is evaluated and the entire subtree below the current node can be pruned if 
$\mr{UB}(H) < \zeta$ (line~3-4). 
%
% もし$\|g_H(\*B)\|_2 > \lambda_m$であれば，$H$は$\cW$に含まれることになる．
On the other hand, if $\eta_H > \zeta$, $H$ is added to $S_{t+\frac{1}{2}}$ (line~6).
Then, the threshold $\zeta$ is updated (line~7), after which we can remove 
$\{ H^\prm \mid \eta_{H^\prm} < \zeta, \ H^\prm \notin S_t\}$ (line~8).
This guarantees $|S_{t+\frac{1}{2}}| \leq 2 s$ (for simplicity, we assume there is no tie of $\eta_H$).
%
% $H$からの展開はgSpanが行い，訓練データに含まれる部分グラフのみを生成できる(このとき，$psi_H(G_i)$を計算できる)．
The children nodes of $H$ are created by gSpan in the Expand function, by which only the subgraphs $H^\prm$ included in the training dataset can be generated and as a byproduct of this process, we obtain $\psi_{H^\prm}(G_i)$. 
Note that if the children are already generated in the previous iterations, we can reuse them.
%} % \blue{(b and theta are randomly initialize)}
%
Once 
$S_{t+\frac{1}{2}}$ 
is determined, $\*B^t$ is updated in line~11 of Algorithm~\ref{alg:regularization-path}, in which the step length $\gamma_{\*B}^t$ is determined by the batcktrack search.
Here, $\*B_{S_{t+\frac{1}{2}}}$ is a submatrix of $\*B$ in which only the columns specified by $S_{t+\frac{1}{2}}$ are extracted.
In line 12-13 of Algorithm~\ref{alg:regularization-path}, $\*b$ and $\*\theta$ are updated by the standard gradient descent, in which the batcktrack search is also performed.

% --------------------------------------------------
\subsubsection{Convergence Analysis}
\label{sssec:convergence-analysis}

% ここでは提案法が最低でもcritical pointへのsublinear convergenceを保証できることを示す．
We here show that EIN can converge to a critical point with at least sublinear convergence.

% $\*b$と$\*\theta$のgradient descentはそれぞれ$\tau_{\max}$回更新するとする．
Suppose that the gradient descent of $\*b$ and $\*\theta$ iterates $\tau_{\max}$ times. 
%
% この内側の更新の添字$\tau$で以下のように表現する
The index $\tau$ represents the number of iteration of this inner loop as follows.
\begin{align*}
 \*b^{t,\tau+1} &= \*b^{t,\tau} - \gamma^{t,\tau}_{\*b} \nabla_{\*b} \ell^{t,\tau},
 \\ %---
 \*\theta^{t,\tau+1} &= \*\theta^{t,\tau} - \gamma^{t,\tau}_{\*\theta} \nabla_{\*\theta} \ell^{t,\tau},
\end{align*}
% ただし，$\gamma^{t,\tau}_{\*b}$と$\gamma^{t,\tau}_{\*\theta}$はそれぞれ$\tau$-th iterationでのステップ幅である．
where 
$\nabla_{\*b} \ell^{t,\tau} = \pd{\ell(\*B^{t+1}, \*b^{t,\tau}, \*\theta^t)}{\*b}$, 
$\nabla_{\*\theta} \ell^{t,\tau} = \pd{\ell(\*B^{t+1}, \*b^{t+1}, \*\theta^{t,\tau})}{\*\theta}$, 
and
$\gamma^{t,\tau}_{\*b}$
and
$\gamma^{t,\tau}_{\*\theta}$
are the step lengths of the $\tau$-th iteration, respectively.
%
% Backtrackは適当な定数$c_{\*B}, c_{\*b}, c_{\*\theta} \in (0,1)$に対して
Let $c_{\*B}, c_{\*b}, c_{\*\theta} \in (0,1)$ be constants used in the backtrack termination conditions.
Our backtrack, in which the step length iteratively decays with a ratio $\rho \in (0,1)$, stops when 
\begin{align}
 \ell(\*B^t, \*b^t, \*\theta^t) - \ell(\*B^{t+1}, \*b^t, \*\theta^t)
 & \geq 
 % {c_{\*B} ( 1 - \rho ) \frac{\gamma^t_{\*B}}{2} \| \nabla_{\*B} L^t_{S_t} \|_F^2}
 {c_{\*B} ( 1 - \rho ) \frac{\gamma^t_{\*B}}{2} \| \nabla_{\*B} \ell^t_S \|_F^2}
 \label{eq:BT-condition-B}
 \\ % ---
 \ell(\*B^{t+1}, \*b^{t,\tau}, \*\theta^t) - \ell(\*B^{t+1}, \*b^{t,\tau+1}, \*\theta^t)
 & \geq 
 c_{\*b}
 \frac{\gamma^{t,\tau}_{\*b}}{2} \| \nabla_{\*b} \ell^{t,\tau} \|_2^2
 \label{eq:BT-condition-b-inner}
 \\ % ---
 \ell(\*B^{t+1}, \*b^{t+1}, \*\theta^{t,\tau}) - \ell(\*B^{t+1}, \*b^{t+1}, \*\theta^{t,\tau+1})
 & \geq 
 c_{\*\theta}
 \frac{\gamma^{t,\tau}_{\*\theta}}{2} \| \nabla_{\*\theta} \ell^{t,\tau} \|_2^2	       
 \label{eq:BT-condition-theta-inner}
\end{align}
are satisfied, respectively.
% が満たされるように更新する．
%
These conditions are from the well-known Armijo condition \citep{nocedal2006numerical}.
%
% これを満たすステップ幅が存在することを証明できる．

% 以下の証明を通して$L$-smoothnessを$\ell$について仮定する．
Throughout the proof, we assume differentiability and $L$-smoothness of $\ell$: 
%
% --------------------------------------------------
% Definition: L-smoothness
% --------------------------------------------------
\begin{defini}
 Let $L$ be the Lipschitz constant and $\*z$ be the vector that concatenates all the elements of $\*B$, $\*b$, and $\*\theta$. 
 The (restricted) $L$-smoothness of $\ell$ is defined as
 \begin{align}  
  \nbr{ \pd{\ell(\*B, \*b, \*\theta)}{\*z} - \pd{\ell(\*B^\prm, \*b^\prm, \*\theta^\prm)}{\*z} }_2
  \leq L \| \*z - \*z^\prm \|_2
  \label{eq:Lipschitz-grad}
 \end{align}
 for any pair of 
 $(\*B, \*b, \*\theta)$
 and
 $(\*B^\prm, \*b^\prm, \*\theta^\prm)$, 
 where 
 $\| \*B \|_0 \leq s$,
 $\| \*B^\prm \|_0 \leq s$.
\end{defini}
%
% \noindent
% It is known that $L$-smoothness is equivalent to 
% \begin{align}
%  \ell(\*B^\prm, \*b^\prm, \*\theta^\prm)
%  \leq 
%  \ell(\*B, \*b, \*\theta) + 
%  \inner{ \pd{\ell(\*B, \*b, \*\theta)}{\*z} }{\*z^\prm - \*z} 
%  + \frac{L}{2} \| \*z^\prm - \*z \|_2^2
%  % \ \text{ for } \ \| \*B \|_0 \leq s, \| \*B^\prm \|_0 \leq s,
%  \label{eq:RSS-all}
% \end{align}
% for
% $\| \*B \|_0 \leq s$ and $\| \*B^\prm \|_0 \leq s$.
% {(Memo: not required here?)}
% which is called decent lemma.

% 以下の補題はBacktrack条件を満たすステップ幅が存在することを保証(多分十分条件): 
% We first guarantee that the backtrack conditions can be satisfied with sufficiently small non-zero step lengths.
We first guarantee the existence of the positive step lengths that satisfy the backtrack conditions.
% --------------------------------------------------
% Lemma: Step length existence
% --------------------------------------------------
\begin{lem}
 There exist positive step lengths 
 {$\gamma_{\*B}^t \geq \rho^2 / L$,}
 $\gamma_{\*b}^{t,\tau} \geq \rho /L$, 
 and 
 $\gamma_{\*\theta}^{t,\tau} \geq \rho / L$ 
 that satisfy \eq{eq:BT-condition-B}, \eq{eq:BT-condition-b-inner}, and \eq{eq:BT-condition-theta-inner}, respectively.
 \label{lem:step-length}
\end{lem}
%
% この証明はAppendix~\red{XX}. 
\noindent
The proof is in Appendix~\ref{app:proof-step-len-existence}.
%
% $\*B$についてはほぼ\citep{damadi2022gradient}で示されたものと同じ，$\*b$と$\*\theta$について既知の既知の結果から直ちにわかる. 
For $\*B$, the proof is mainly from \citep{damadi2022gradient}, and for $\*b$ and $\*\theta$, it is from a known property of the gradient descent.
%
% $L0$制約をペナルティに書き換えた上で，劣微分が$0$を含む点へ収束することを保証 (コーシー列)
The convergence of our procedure is shown by the following theorem, in which our problem \eq{eq:objective} is re-written as an equivalent unconstrained problem:
% --------------------------------------------------
% Theorem: 収束保証 (部分列ではなく全列収束)
% --------------------------------------------------
\begin{theo}
 % Let $\*z$ be a vector that concatenates all the elements of $\*B$, $\*b$, and $\*\theta$, and
 Let
 \begin{align*}
  \tilde{\ell}(\*z)
  = \ell(\*B, \*b, \*\theta) + \delta_s(\*B)
 \end{align*}
 be the objective function with the penalty on the constraint violation
 \begin{align*}
  \delta_s(\*B) = 
  \begin{cases}
   0, & \text{ if } \| \*B \|_{2,0} \leq s, 
   \\
   \infty, & \text{ otherwise. }
  \end{cases}
 \end{align*}
 The sequence $\*z^t$, defined by $\*B^t$, $\*b^t$, and $\*\theta^t$, is assumed to be bounded.
 % Then, the sequence $\*z^t$, defined by $\*B^t$, $\*b^t$, and $\*\theta^t$, 
 Then, $\*z^t$ is a Cauchy sequence that converges to a critical point $\*z^*$ satisfying 
 $\partial \tilde{\ell}(\*z^*) \ni \*0$, where 
 $\partial \tilde{\ell}(\*z)$ 
 is the limiting subdifferential of $\tilde{\ell}$. 
 \label{theo:theo-stationary-point}
\end{theo}
\noindent
See Appendix~\ref{app:proof-stationary-point} for the proof.
The limiting subdifferential \citep{mordukhovich2012variational} is a generalized notion of subdifferential applicable to an arbitrary proper and lower semi-continuous (extended-real-valued) function (See Appendix~\ref{app:subdifferential} for the definition).
The generalized Fermat's rule \citep{rockafellar2009variational} indicates 
$\partial \tilde{\ell}(\*z^*) \ni \*0$
is a necessarily condition of a local minimum.
%
% bounded assumptionはx^tが無限遠点に発散しないことを保証するためのものです．実践的にこれを厳密に保証したければ，例えば，パラメーターに$L_2$ penalty (weight decay)を導入すればよいです（正則化係数は任意に小さくて構いません）．
The boundedness assumption of $\*z^t$ is introduced to ensure that $\*z^t$ does not diverge to infinity. 
In practice, if one wishes to guarantee this rigorously, a possible approach is to introduce an $L_2$ penalty (weight decay) on the parameters (for which the regularization coefficient can be chosen to be arbitrarily small).
We further derive the following convergence rate:
% --------------------------------------------------
% Theorem: 収束rate
% --------------------------------------------------
\begin{theo}
 Under the same assumption as Theorem~\ref{theo:theo-stationary-point}, our optimization algorithm has at least the following sublinear convergence rate:
 \begin{align*}
  \| \*z^t - \*z^* \|_2 = O(t^q), 
 \end{align*}
 where $q < 0$. 
 \label{thm:convergence-rate}
\end{theo}
\noindent
Based on properties of $\*z^t$ obtained through the proof of Theorem~\ref{theo:theo-stationary-point}, the above rate can be derived by the almost same proof as Theorem~2 of \citep{attouch2009convergence}.
%
% Although the proof is almost same as \citep{attouch2009convergence}, 
We show the proof in Appendix~\ref{app:proof-rate} for completeness.
% 証明はほぼ\citep{attouch2009convergence}と同じだが完全性のためにAppendix~\red{XX}に．

% --------------------------------------------------
\subsection{Post-hoc Analysis for Knowledge Discovery}
\label{sssec:interpret-EIN}
 
% EIN学習後は少数の$\*\beta_H$だけが非零となるため，予測に寄与する部分グラフを知ることができる．
% After the optimization, only a small number of $\*\beta_H$ have non-zero values, by which we can identify predictive subgraphs. 
After the optimization, only $s$ subgraphs have non-zero values of $\*\beta_H$, by which we can identify predictive subgraphs. 
%
% Let
% $\cS = \{ H \in \cH \mid \*\beta_H \neq \*0 \}$
% be the set of the selected subgraphs by EIN.
% % 後の実験では典型的には$\cS$の要素数はa few hundreds以下だった．
% % In our later experiments, we observed that the number of selected $H$ was typically less than a few hundreds.
% In our later experiments, we observed that $|\cS|$ was typically ranged from around 50 to at most a few hundreds.
% less than a few hundreds.
In our later experiments, we observed that small $s$ (up to $100$) can achieve sufficiently high predictive performance.
% 
% これらをdomain専門家に提示することで知見に繋げることができる可能性がある
% このような場合はこれらをdomain専門家が直接観察してinsightになる可能性がある
% Then, some insight may be obtained just by directly observing all of those selected subgraphs by the domain expert of the data.
Therefore, examining the selected subgraphs—possibly by a domain expert—may lead to new insights about the data.

% EINは$|\cS|$次元のbinary vectorを受け取るニューラルネットワークとして見ることができる．
The trained EIN can be seen as an $s$-dimensional input neural network, to which post-hoc knowledge discovery methods \citep{bhati2025survey} can be applied.
%
% このニューラルネットワークに対して，post-hocなKnowledge Discovery手法を適用することが当然可能である
%
% 例えば有名なSHAP\citep{lundberg2017unified}が含まれる．
For example, well-known SHAP \citep{lundberg2017unified} and LIME \citep{ribeiro2016why}, both of which provide local feature importance, is applicable.
Another typical approach is to use the trained EIN as a teacher and fit an interpretable surrogate model (e.g., decision tree), from which a possible decision rule can be estimated \citep{molnar2025interpretable}.
% あるいはEINをteacherにして，解釈可能なsurrogateをfittingする
% \citep{molnar2025interpretable}ことも可能となる
%
% 通常，部分グラフ有無に対してこういった方法あ必ずしも容易ではない．
Note that, usually, applying interpretable machine learning methods directly to exhaustive subgraph features can be computationally intractable. 
% exhaustive subgraph isomorphism features can be computationally intractable. 
%
% 少数の部分グラフを選び出すEINによって，様々な解析が可能になっている
% Because of EIN, which selects a few important subgraphs, a variety of post-hoc analyses become much easier to apply.
Benefiting from EIN, we can only focus on a small number of important selected subgraphs, by which a variety of post-hoc analyses become much easier to apply.

% --------------------------------------------------
% Fig: EIN + GNN
% --------------------------------------------------
\begin{figure}[t]
% \begin{wrapfigure}[12]{R}{.35\tw}

% \vspace{-2em}

 \centering

 \ig{.35}{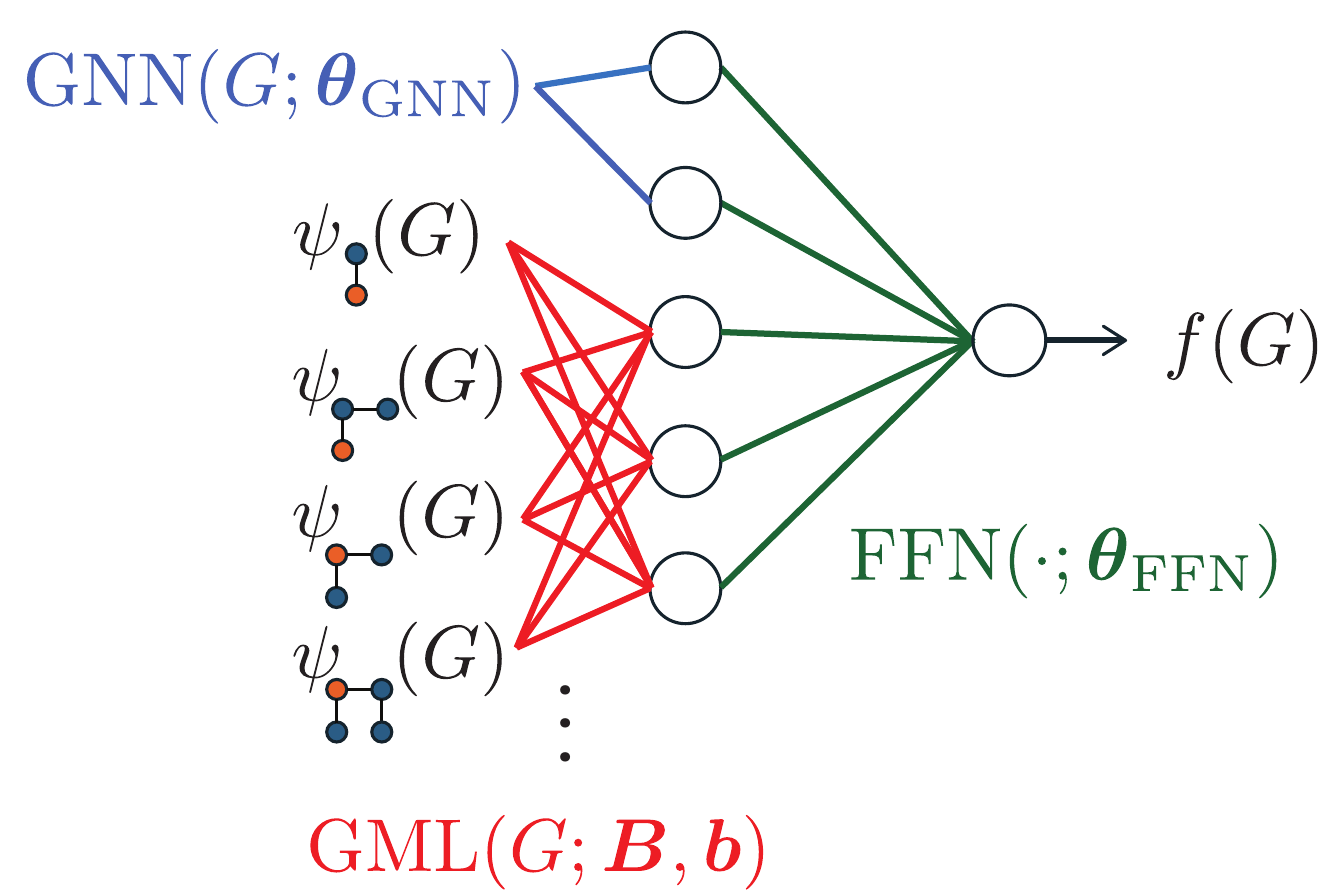}

 \caption{
 % EIN+GNNの例．
 A simple example of EIN combined with GNN.
 %
 % EINの出力とGNN畳み込みの結果を融合してFFNに通す機構を持つ．
 }
 \label{fig:EIN+GNN}
\end{figure}
% \end{wrapfigure}

% --------------------------------------------------
% \subsection{GNNとの融合拡張}
\subsection{Combining with Graph Neural Network}
\label{ssec:combine-GNN}

% EINではFFNを用いたモデルを構築し，誤差逆伝播に基づいて最適化および枝刈りを行う．
Since EIN is based on the standard backpropagation mechanism, general neural network models can be combined flexibly.
%
% この枠組みにより，一般的な深層学習モデルを柔軟に組み込むことが可能となる．
%
% そこで，本論文ではEINとグラフニューラルネットワーク（GNN）と組み合わせた拡張としてEIN+GNNを提案する：
For example, a simple approach to combining EIN with a graph neural network (GNN) is
\begin{align}
 % 	f(G_i) = \sum_{H \in \cH} \psi(G_i ; H) \beta_{H} + \beta_0 = \*\psi_i^\top \*b + \beta_0, 
  % f(G_i) &= \mathrm{MLP}_{\*\Theta} (\mathrm{GNNConv}_{\*\Theta}(G_i) \oplus \sigma (\*h)) 
 \begin{split}
  f(G_i) &= \mr{FFN} (
  \mr{GNN}(G_i ; \*\theta_{\rm GNN}) \\
  & \qquad
  \oplus 
  \mr{GML}(G_i ; \*B, \*b) ; \*\theta_{\rm FFN}),
 \end{split}
 \label{eq:ein-gnn}
\end{align}
% ここで，$\oplus$はベクトルの結合操作を表す．
where $\oplus$ is the vector concatenation, $\mr{GNN}$ is any GNN that outputs arbitrary dimension representation vector, and
$\*\theta_{\rm GNN}$ 
and 
$\*\theta_{\rm FFN}$
are parameters of $\mr{GNN}$ and $\mr{FFN}$, respectively.
This combination enables any GNNs to enhance the discriminative ability in terms of the selected subgraphs by EIN.
% In this paper, this combined model is denoted as EIN+GNN.
A schematic illustration of a simplest case is shown in Fig.~\ref{fig:EIN+GNN}. 
%
% EIN+GNNの計算も同様の交互最適化で実行できる．
Computations of the combined model can also be performed by almost the same alternating update shown in Algorithm~\ref{alg:regularization-path}. % \eq{eq:update-B}-\eq{eq:update-theta}.
%
% FFN部分のパラメータ更新\eq{eq:update-theta}において，
% % EIN+GNNのアルゴリズムはEINと同様であり，$\eq{eq:update-theta}$によりGNNの畳み込み層のパラメータ
% $\*\Theta_{\rm FFN}$
% と同時に
% $\*\Theta_{\rm GNN}$
% を最適化する．
The only difference is that the FFN parameter update (line~13 of Algorithm~\ref{alg:regularization-path})
% \eq{eq:update-theta} 
is replaced with the simultaneous update of
$\*\theta_{\rm FFN}$
and
$\*\theta_{\rm GNN}$.
%
% \red{なぜこうするのか，がない．解釈性と柔軟性のtrade-off.}
% %
% \red{また，EINでは頂点ラベルのみを利用したが，GNN部分には頂点属性値を利用する．}
% EIN+GNNではGNNの表現力を保ちながら，モデルが注目した局所的な部分グラフ構造を得ることができる．

%%%%%%%%%%%%%%%%%%%%%%%%%%%%%%%%%%%%%%%%%%%%%%%%%%%%%%%%%%%%%%%%%%%%%%
\section{Related Work}
\label{sec:related-work}

For GNNs, the message passing based approach has been widely employed \citep{zhou2020graph}.
It is known that the expressive power of classical message passing approaches is limited by the first order Weisfeiler-Lehman (1-WL) test \citep{zhao2022from}.
Many studies have tackled this limitation, some of which actually can be comparable with higher order WL tests.
For example, $k$-hop extensions of the message passing \citep[e.g.,][]{haija2019mixhop,nikolentzos2020khop,wang2021multi,chien2021adaptive,brossard2020graph} and the higher order (tensor) representations \citep[e.g.,][]{morris2019weisfeiler,maron2018invariant,maron2019provably,maron2019on,keriven2019universal,geerts2022expressiveness,pmlr-v97-murphy19a} are popular approaches.
Further, several studies \citep{bouritsas2023improving,cotta2021reconstruction,barcelo2021graph,bevilacqua2022equivariant,yan2024efficient} incorporate small subgraph information, such as pre-specified motifs and ego-networks, into neural networks.
%
% On the other hand, EIN identifies a small number of (globally) predictive subgraphs adaptively, based on the exact enumeration of all the subgraphs in the dataset, which is obviously different approach from the above popular GNN studies.
On the other hand, EIN performs the data-adaptive selection of a small number of (globally) predictive subgraphs, based on the exact enumeration of all the subgraphs in the dataset, which is obviously different approach from the above popular GNN studies.
Further, most of GNNs can be combined with EIN by \eq{eq:ein-gnn}.

% 最近ではGNNの解釈に関する研究が盛んに行われている．
Although explainable GNNs are also studied, according to \citep{yuan2022explainability}, most of them are for the `instance-level' explanation, which considers an explanation for the prediction of the specific one input graph.
%
% しかし，GNNの解釈は一般的に容易ではなく，\citep{yuan2022explainability}によると，GNNの説明可能性に関する研究の多くはインスタンスレベルである．
%
% 例外として，XGNN\citep{yuan2020xgnn}では，学習済みのGNNに対してターゲットラベルに対するGNN出力を最大化することで，重要な識別グラフを生成できる．
% An exception is XGNN \citep{yuan2020xgnn} that maximizes the output of a trained GNN with respect to 
A few exceptions are approaches based on the maximization of the trained GNN output for a target label \citep{yuan2020xgnn}, the latent space prototype learning \citep{azzolin2023global}, and the kernel-based filtering \citep{Feng2022kergnns}. 
%
% しかし，我々の手法と異なり，識別グラフは入力データの部分構造である保証はない．
However, unlike EIN, they cannot guarantee that the identified important graphs are actually subgraphs of the input graphs because none of them are based on the exact matching.

% An exception is XGNN \citep{yuan2020xgnn}, in which important discriminative graphs are generated for a given already trained GNN by maximizing the GNN output for a target label. However, unlike our method, the prediction model itself remains black-box, and thus, it is difficult to know underlying dependency between the identified graphs and the prediction.
%
%また予測モデルがブラックボックスであるため，
%特定されたグラフと予測間の依存関係を知ることは困難である．
% GNNの持つグラフ同型性処理能力に関する話

In the context of graph mining, discriminative pattern mining has also been studied \citep[e.g.,][]{thoma2010discriminative,potin2025pattern,chen2022contrast}.
However, most of them are not directly incorporated in a learning model.
Typically, the subgraph enumeration (using some pruning strategies) is performed based on a discriminative score (e.g., ratio of frequency between two classes) independent from a learning model, and the selection results can be used in a subsequent prediction model training.
Although \citep{nakagawa2016safe,yoshida2019learning,yoshida2021distance,tajima2024learning} enumerate subgraphs during a model optimization, they are limited to a linear combination of the subgraphs. 
%
% Note that \citet{tajima2024learning} also use a combination of proximal gradient and gSpan, from which our optimization procedure is extended to the neural network backpropagation and the group-sparsity. 
%
Although \citep{nakagawa2016safe,yoshida2019learning,yoshida2021distance} provide a stronger pruning rule that can satisfy the global optimality, the convex formulation of the model training is required.   
Several studies have considered learning $L_0$ based sparse neural networks \citep[e.g.,][]{louizos2018learning,damadi2024learning}.
However, to our knowledge, a discriminative subgraph mining that directly combines the exact subgraph enumeration with an $L_{0,2}$ constraint neural network has not been investigated.

% \red{(Memo: $L_0$ + NN \citep{louizos2018learning,damadi2024learning})}

% ==================================================
\section{Experiments}

% 人工データおよびベンチマークデータを用いて，EINの予測性能と解釈性を実証する．
We verify the prediction performance and interpretability of EIN through synthetic and benchmark datasets.
%
% また，RFによる部分グラフの重要度と，決定木による部分グラフにより表現されるルールを確認する．
%
For all the datasets, we partitioned them into 
${\rm train}:{\rm valid}:{\rm test} = 6:2:2$. 

% --------------------------------------------------
\paragraph{Synthetic Datasets. }

% --------------------------------------------------
% Fig: Cycle 
% --------------------------------------------------
\begin{figure}[t]

 \centering 

 \subfigure[$H_p$]{\ig{.23}{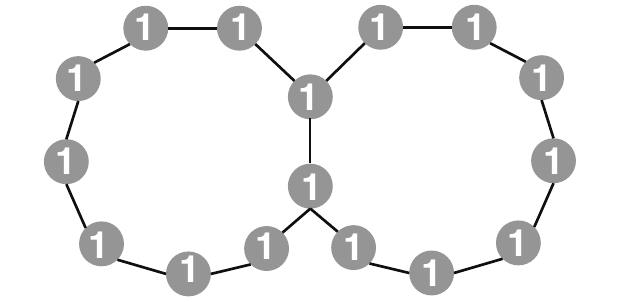}}
 \subfigure[$H_n$]{\ig{.23}{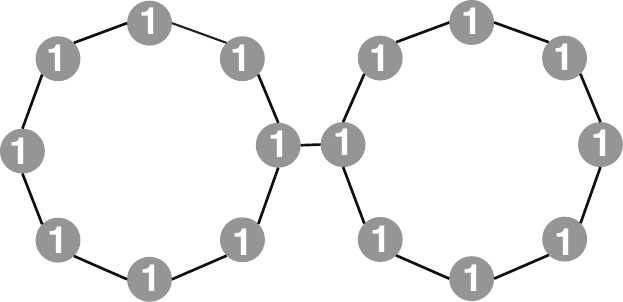}}
 % \subfigure[$H_{\rm padding}$]{\ig{.16}{figs/experiment/Cycle89_XOR/tree.pdf}}
 \subfigure[$H_{\rm padding}$]{\ig{.14}{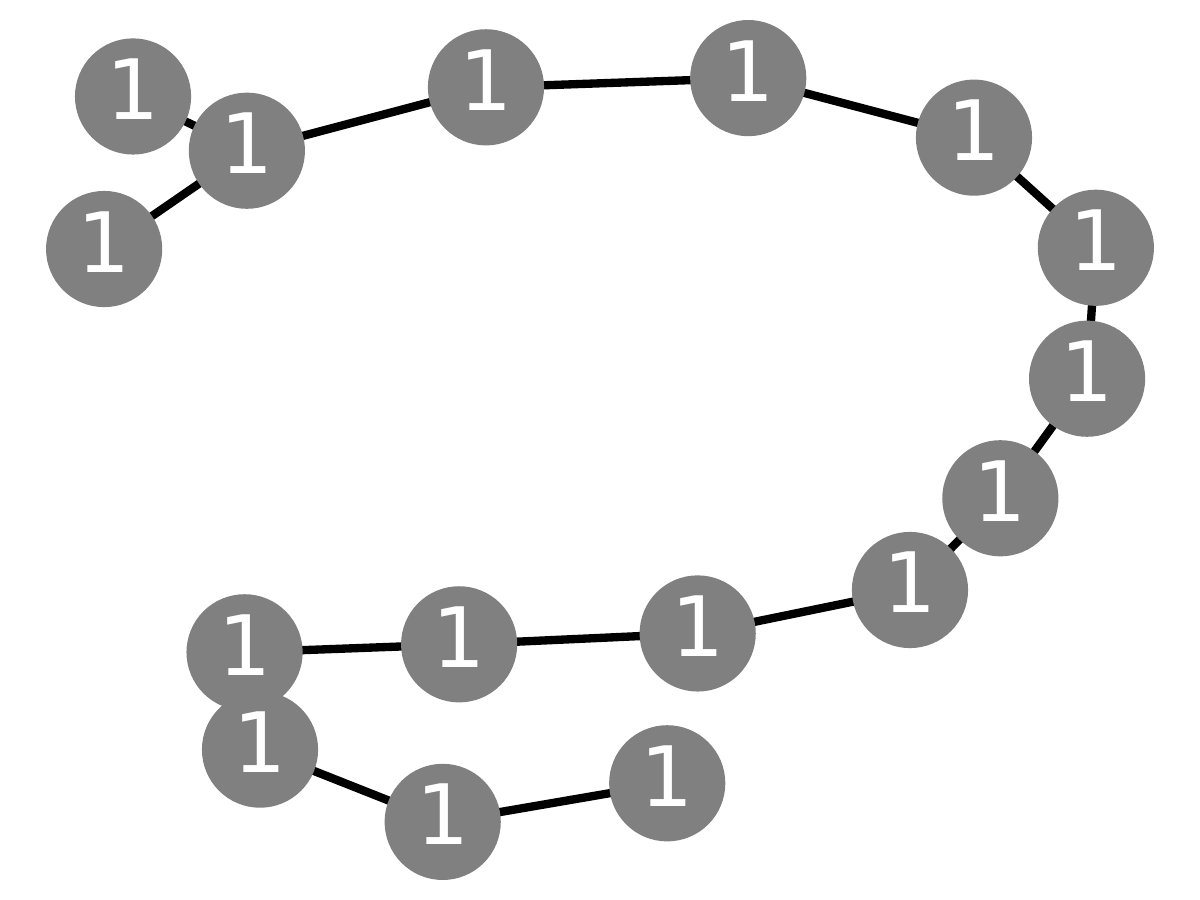}}

 \caption{
 Subgraphs (a) and (b) have closed paths whose lengths are 8 and 9, respectively, which are difficult to discriminate.
 Subgraph (c) is used for adjusting the number of nodes that have label `1'.
 Note that all subgraphs (a), (b), and (c) have 16 nodes.
 }
 \label{fig:cycle}
\end{figure}
% \begin{figure}[t]
%     \centering
% 	\includegraphics[width=0.7\linewidth]{figs/experiment/cycle89.pdf}
%     \caption{長さ8 or 9の閉路を持つ分類困難なグラフ．Left: $G_p$, Right: $G_n$}
% 	\label{fig:cycle89}
% \end{figure}
% \begin{figure}[t]
%     \centering
% 	\includegraphics[scale=0.15]{figs/experiment/Cycle89_XOR/tree.pdf}
%     \caption{Cycle89\_XORデータに用いる木のデータ}
% 	\label{figure-Cycle89-tree}
% \end{figure}

% 人工データセットCycle89とCycle89\_XORを利用する．
% 図~\ref{fig:cycle89}に示すcycle構造を持つ部分グラフ$G_p, G_n$を使って人工データを作成した．
We created two synthetic datasets by using subgraphs shown in Figure~\ref{fig:cycle}. 
%
% $G_p, G_n$は長さ8 or 9の閉路を持つ (頂点数はどちらも16)．
$H_p$ and $H_n$ are subgraphs with 16 nodes, which is known to be difficult to discriminate by standard GNNs.
%
% これは標準的なGNNでは判別が困難なものである．
%
% 簡潔に作り方を述べる細かい話はAppendix
% We here only briefly describe how the two datasets are created.
%
% See \red{Appendix~\ref{}} for details.

% 一つ目のデータは，$G_p$と$G_n$の判別性を評価するものである．
The first dataset use
$H_p$
and
$H_n$,
by which positive and negative classes are defined.
%
% ランダムなグラフに$G_p$, $G_n$を一つ埋めこみ，各クラス300サンプルずつ生成する．
We first generated a random connected graph, connected with one of $H_p$ or $H_n$ randomly. 
%
% ランダムなグラフは頂点ラベル0とし，$G_p$と$G_n$の頂点ラベルを1とする．
The initial random graph has node labels `0', and $H_p$ and $H_n$ have node labels `1'.
We generated $300$ instances for each of the positive and the negative classes.
%
% これをCycle datasetと呼ぶ．
We call this dataset `Cycle'.
See Appendix~\ref{app:syn-data} for further details of the initial random graph.

% 次に，非線形な規則性(XOR)を含むCycle\_XORデータも作成した．
In the second data, the positive and the negative classes are defined by the XOR rule, i.e., a nonlinear rule, of $H_p$ and $H_n$, which we call `Cycle\_XOR'
%
% 最初にランダムなグラフを生成し，$G_p$, $G_n$をXOR規則で埋め込む (ランダムなグラフは頂点ラベル0とし，$G_p$と$G_n$の頂点ラベルを1とする)．
By using the same random graph as the Cycle dataset, $H_p$ and/or $H_n$ are embedded into input graphs.
%
% つまり，$G_p$と$G_n$のどちらか片方を持つ場合に$y = 1$，両方存在もしくはどちらも存在しない場合に$y = 0$となる．
Therefore, if a graph have one of $H_p$ or $H_n$, the class label is $y = 1$, otherwise $y = 0$, i.e., in the case that both of $H_p$ and $H_n$ are included or neither of them are included. 
%
% 4つの状態，各状態150サンプルずつ生成する (つまり各クラス300サンプルずつ)．
We generated $150$ instances each of four states of XOR, which results in $300$ instances each of the positive and the negative classes.
%
% ただし，$G_p$と$G_n$が含まれる数に依存して，頂点ラベル「１」の数が変わってしまう(0だと0, 1つだと16, 2つだと32)．
However, in this setting, the difference in the number of the nodes labeled as `1' may make the discrimination easier (32 if both of $H_p$ and $H_n$ exist, 16 if one of $H_p$ and $H_n$ exists, and 0 if neither of $H_p$ and $H_n$ exists).
To avoid this, we added an simple subgraph $H_{\rm padding}$ so that the number of the nodes labeled as `1' is 32 for all graphs.
% 1の数によって判別されることを防ぐため，以下の方法で1の数が同じになるようにする．
% %
% 具体的には，ラベル1のノード数を揃えるために$G_p$, $G_n$を両方とも含まない場合には\Figref{figure-Cycle89-tree}の部分グラフを二つ，$G_p$のみ含む場合には$G_p$を二つ($G_n$の場合は$G_n$を二つ)，両方とも含む場合には$G_p$, $G_n$をそれぞれ一つ埋め込む．
% %
% \Figref{figure-Cycle89-tree}の部分グラフは先端に分岐構造を持つものの閉路を形成していない単純なものとした．

% 実験では，EINがこれらのグラフを分類できるか，また真に重要な部分グラフを発見することができるかどうかを検証する．

% --------------------------------------------------
\paragraph{Benchmark Datasets. }

% ベンチマークデータセットBZR，COX2，DHFR，ENZYMES\citep{tudataset}，ToxCast，SIDER\citep{wu2018moleculenetbenchmarkmolecularmachine}を使用した．
% %
% （ENZYMES，ToxCast，SIDERの場合は，複数クラスのうち2つのみを使用して2値化問題を作成）．

% As benchmark datasets, we used BZR, COX2, DHFR, and ENZYMES from \citep{morris2020tudataset}, and ToxCast and SIDER from \citep{wu2018moleculenet}.
As benchmark datasets, we used BZR, COX2, DHFR, ENZYMES, and PTC MR from \citep{morris2020tudataset}, and ToxCast from \citep{wu2018moleculenet}.

% --------------------------------------------------
% \paragraph{Baselines.}
\paragraph{Compared Methods.}

% GCN \citep{kipf2017semi}, GAT \citep{velickovic2018graph}, GATv2 \citep{brody2022how}, GIN \citep{xu2018how}, PNA \citep{corso2020principal}, GNN-AK \citep{zhao2022from}, PPGN \citep{maron2019provably}を使用した．
% For performance comparison, we used GNN methods such as GCN \citep{kipf2017semi}, GAT \citep{velickovic2018graph}, GATv2 \citep{brody2022how}, GIN \citep{xu2018how}, PNA \citep{corso2020principal}, GNN-AK \citep{zhao2022from}, and PPGN \citep{maron2019provably}.
For performance comparison, we used GNN methods such as GCN \citep{kipf2017semi}, GAT \citep{velickovic2018graph}, GATv2 \citep{brody2022how}, GIN \citep{xu2018how}, PNA \citep{corso2020principal}, GIN-AK \citep{zhao2022from}, PPGN \citep{maron2019provably}, and ESC-GNN \citep{yan2024efficient}.
%
% EINは頂点ラベルのみを利用するが，GNNは頂点属性を利用する．
For node attributes, EIN only used discrete node labels, while compared GNN methods also incorporate continuous node attributes. 
%
% GCN，GATについては，ユニット数$\{ 64, 128, 256 \}$とエポック数は検証セットによって最適化し，上記以外の設定については，\citep{you2021identity}に従った．
GCN, GAT, GATv2, and GIN optimizes the number of the units $\{ 64, 128, 256 \}$ and the number of epoch by the validation set, and other settings follow \citep{you2021identity}.
%
% またGATv2，GINは上記設定と同様に設定した．
%
% PNAはユニット数$\{ 16, 32, 64 \}$とエポック数は検証セットによって最適化し，messege passingの回数は2回とした．
PNA optimizes the number of the units $\{ 16, 32, 64 \}$ and the number of epochs by the validation set, and the number of the message passing is set as $2$. 
%
% GNN-AKにはGINをbaseline GNNとして利用する．GNN-AKとPPGNのその他の設定は著者実装に従った．
% GNN-AK uses GIN as the base GNN. 
%
Other settings of GIN-AK, PPGN, and ESC-GNN follow the author implementation.
For EIN, 
$\text{maxpat} \in \{5, 10\}$, 
$K \in \{2, 6, 10\}$, 
and
$\tau_{\rm max} \in \{1, 30\}$ for $\*\theta$ ($\tau_{\rm max}$ for $\*b$ is fixed as $1$)
% $\sigma \in \{\mathrm{sigmoid}, \mathrm{LeakyReLU}\}$
are selected by the validation performance. 
% , and the activation function is the sigmoid function.
%
We also evaluate performance of the combination of EIN and GNN described in \S~\ref{ssec:combine-GNN}, in which GIN is employed as the combined GNN.
In the results, the single EIN is denoted as EIN-$L_{0,2}$ and the EIN combined with GIN is denoted as EIN-$L_{0,2}$-GNN.
For both of EIN-$L_{0,2}$ and EIN-$L_{0,2}$-GNN, the candidate of the sparsity $s$ is $\cS = \{ 1, 5, 10, 25, 50, 75, 100 \}$.
% For GNN, we employed GIN and the combined model is denoted as EIN+GIN.
%
For other details, see Appendix~\ref{app:settings-EIN}.

% --------------------------------------------------
\subsection{Prediction Accuracy Comparison}

% --------------------------------------------------
% Tab: Accuracy 
% --------------------------------------------------
\begin{table*}[t]
% \begin{minipage}{.75\tw} 
 \caption{Accuracy comparison on synthetic and benchmark datasets (mean and std of 10 trials).
 The bold-face indicates that the result is comparable with the best average method in a sense of one-sided $t$-test (significance level 5\%).
 }
	\label{table-accuracy}
	\begin{center}
 	\scalebox{.75}{
\begin{tabular}{lcccccccc}
\hline
Method & BZR & COX2 & DHFR & ENZYMES & PTC MR & ToxCast & Cycle & Cycle\_XOR \\
\hline
GCN & 81.6 $\pm$ 2.3 & 78.9 $\pm$ 1.1 & 69.5 $\pm$ 3.3 & 75.0 $\pm$ 4.6 & {\bf 57.8 $\pm$ 6.5} & 59.3 $\pm$ 1.9 & 47.7 $\pm$ 1.7 & 49.9 $\pm$ 3.3 \\
GAT & 82.1 $\pm$ 3.3 & 77.6 $\pm$ 2.6 & 71.8 $\pm$ 2.0 & 73.2 $\pm$ 3.9 & 57.1 $\pm$ 5.4 & {\bf 60.7 $\pm$ 2.7} & 48.9 $\pm$ 2.0 & 51.8 $\pm$ 3.8 \\
GATv2 & 82.3 $\pm$ 2.6 & 78.4 $\pm$ 2.3 & 69.8 $\pm$ 3.9 & 73.8 $\pm$ 4.0 & {\bf 60.2 $\pm$ 5.1} & {\bf 60.3 $\pm$ 2.6} & 50.9 $\pm$ 3.5 & 51.3 $\pm$ 3.3 \\
GIN & 80.9 $\pm$ 2.8 & 77.8 $\pm$ 2.6 & 69.1 $\pm$ 2.0 & 75.2 $\pm$ 6.1 & 55.5 $\pm$ 4.8 & 58.9 $\pm$ 1.9 & 50.3 $\pm$ 2.8 & 74.0 $\pm$ 2.0 \\
PNA & 82.9 $\pm$ 3.0 & 78.7 $\pm$ 0.7 & 66.0 $\pm$ 4.1 & 72.5 $\pm$ 7.9 & {\bf 59.9 $\pm$ 4.7} & 57.9 $\pm$ 2.8 & 50.3 $\pm$ 4.3 & 74.3 $\pm$ 1.9 \\
PPGN & 83.0 $\pm$ 3.6 & {\bf 79.8 $\pm$ 1.6} & 76.9 $\pm$ 4.0 & 65.8 $\pm$ 8.1 & {\bf 59.2 $\pm$ 7.8} & 60.2 $\pm$ 1.9 & 50.9 $\pm$ 2.2 & 78.5 $\pm$ 11.7 \\
GIN-AK & {\bf 84.0 $\pm$ 2.2} & {\bf 80.0 $\pm$ 3.3} & 75.6 $\pm$ 3.5 & {\bf 83.0 $\pm$ 8.6} & {\bf 57.7 $\pm$ 4.0} & {\bf 60.8 $\pm$ 1.7} & 52.9 $\pm$ 3.5 & 73.2 $\pm$ 4.3 \\
ESC-GNN & {\bf 85.4 $\pm$ 2.2} & {\bf 80.9 $\pm$ 2.7} & {\bf 80.0 $\pm$ 2.7} & {\bf 79.0 $\pm$ 5.6} & 56.4 $\pm$ 4.3 & {\bf 62.1 $\pm$ 2.9} & 54.0 $\pm$ 2.9 & 72.3 $\pm$ 2.9 \\
EIN-$L_{0,2}$ & {\bf 86.0 $\pm$ 2.4} & {\bf 80.5 $\pm$ 3.0} & {\bf 79.4 $\pm$ 3.6} & 59.8 $\pm$ 6.7 & {\bf 59.1 $\pm$ 4.5} & 60.0 $\pm$ 1.5 & {\bf 100.0 $\pm$ 0.0} & {\bf 100.0 $\pm$ 0.0} \\
EIN-$L_{0,2}$-GNN & 83.4 $\pm$ 2.9 & {\bf 81.1 $\pm$ 2.3} & {\bf 80.4 $\pm$ 2.4} & 73.3 $\pm$ 6.7 & {\bf 56.5 $\pm$ 4.4} & {\bf 61.6 $\pm$ 1.9} & {\bf 100.0 $\pm$ 0.0} & {\bf 98.5 $\pm$ 2.8} \\
\hline
EIN-$L_{0,2}$ \# non-zeros & 39$\pm$30 & 80$\pm$29 & 55$\pm$19 & 47$\pm$31 & 70$\pm$29 & 32$\pm$16 & 1$\pm$0 & 5$\pm$0 \\
\hline
\end{tabular}
    }
    \end{center}
% \end{minipage}
%\begin{minipage}{.21\tw}
% \includegraphics[width=\tw]{figs-experiment-Cycle_XOR-decision_tree_CycleXOR.pdf}
% \end{minipage}
\end{table*}

% Table\ref{table-accuracy}には人工データ，ベンチマークデータにおける分類精度を示す．
Table~\ref{table-accuracy} shows classification accuracy on synthetic and benchmark datasets.
%
% Cycle, Cycle\_XORデータの精度から，標準的なGNNだけでなく，表現力を高めたモデルGIN-AK，PPGNを用いても長さ8以上の閉路の検出が行えていないことがわかる．
The results on Cycle and Cycle\_XOR indicate that GNN based methods cannot discriminates closed paths shown in Fig.~\ref{fig:cycle} (a) and (b).
%
% 一方，EINではCycleで100\%，Cycle\_XORで99\%の精度で分類できており，既存モデルと比較して高い表現力を持つことがわかる．
On the other hand, in Cycle and Cycle\_XOR datasets, EIN-$L_{0,2}$ and EIN-$L_{0,2}$-GNN achieved almost 100\% classification.
This indicates that EIN has a high discriminative ability about the subgraph structure, and flexibility that can capture a nonlinear relation.

% ベンチマークデータにおいても，EINおよびEIN+GNNはENZYMESを除くすべてのデータで最高の平均精度を記録し，EINにもつ部分グラフに対する高い表現力による効果が示されている．
For the other benchmark datasets, EIN-$L_{0,2}$ and/or EIN-$L_{0,2}$-GNN show superior to or comparable with other GNNs except for ENZYMES.
%
% また，EIN+GINは全てのデータセットでGINより高い精度を記録しており，部分グラフ情報を利用することによる精度改善が示されている．
% EIN+GIN improves GIN for all datasets, which suggests that the exact subgraph information is essential for the prediction.
We do not observe large differences between EIN-$L_{0,2}$ and EIN-$L_{0,2}$-GNN for most of datasets.
EIN-$L_{0,2}$-GNN largely improves EIN-$L_{0,2}$ only for ENZYMES.
For the ENZYMES dataset, the accuracy of EIN-$L_{0,2}$ is not high, which suggests achieving high discriminability only from SIF is difficult in this dataset.
In such a case, combining EIN with GNN can be effective by incorporating message passing based feature. % (though in EIN is advatageous in a sense of interpretability, because the prediction is purely from SIF). 
% この場合はGNNのmessage passing baseな特徴を取り入れることで改善が可能だと示唆されてる（ただし，純粋にSIFだけのモデルの方が解釈性は高い）

% さらに，EINは\tablename~\ref{table-accuracy}の下段に示す数の部分グラフしか使用しておらず，予測において重要な部分グラフに着目できていると言える．
% Further, the number of the selected subgraphs in EIN, shown in the bottom of \tablename~\ref{table-accuracy} is at most a few hundreds, from which we see that EIN can effectively identify a small number of important subgraphs for the prediction.
Further, the number of the selected subgraphs in EIN, shown in the bottom of \tablename~\ref{table-accuracy} is obviously small, from which we see that EIN can effectively identify a small number of important subgraphs for the prediction.

% --------------------------------------------------
\subsection{Examples of Post-hoc Analysis}

% --------------------------------------------------
% Fig: SHAP
% --------------------------------------------------
\begin{figure*}[t]

 \centering

 \begin{minipage}{.68\tw}  

 \begin{minipage}{.53\tw}  
 \subfigure[pat23]{\ig{.32}{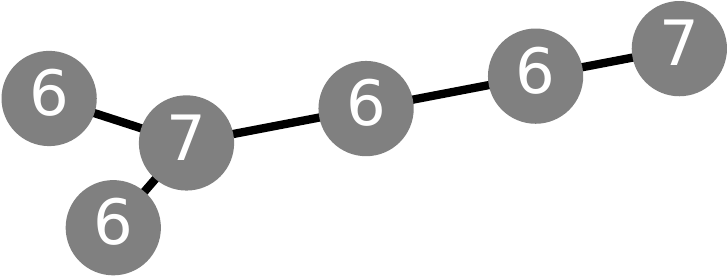}}
 \subfigure[pat0]{\ig{.25}{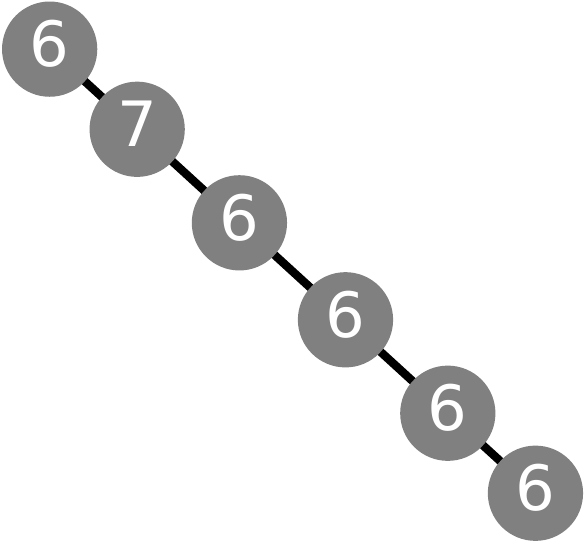}}
 \subfigure[pat5]{\ig{.27}{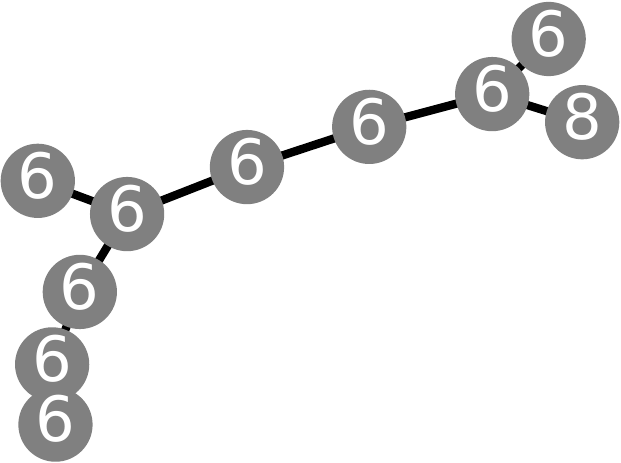}}\\
 \subfigure[pat9]{\ig{.27}{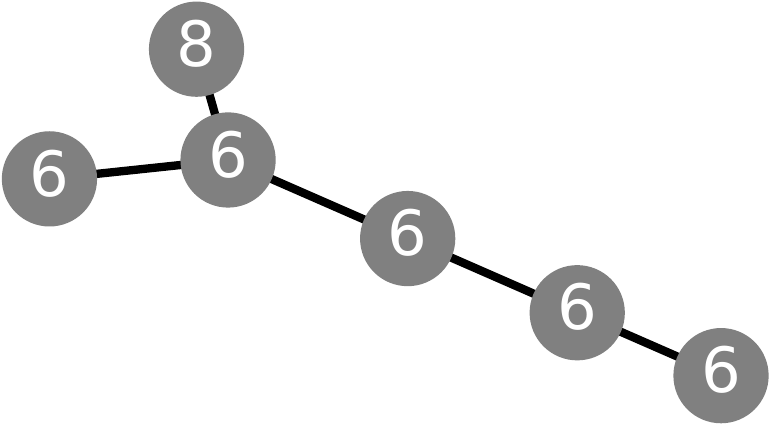}}
 % \subfigure[pat21]{\ig{.02}{./figs-experiment-ToxCast-patterns-21-crop.pdf}}
 \subfigure[pat21]{\ig{.33}{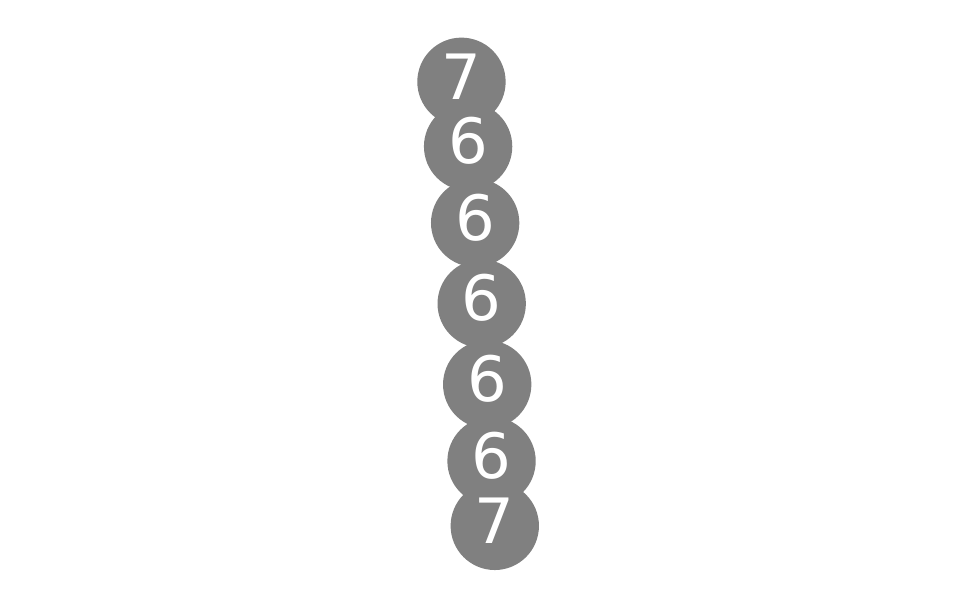}}

 \end{minipage}
 \begin{minipage}{.45\tw}  
  \subfigure[SHAP values]{\ig{1}{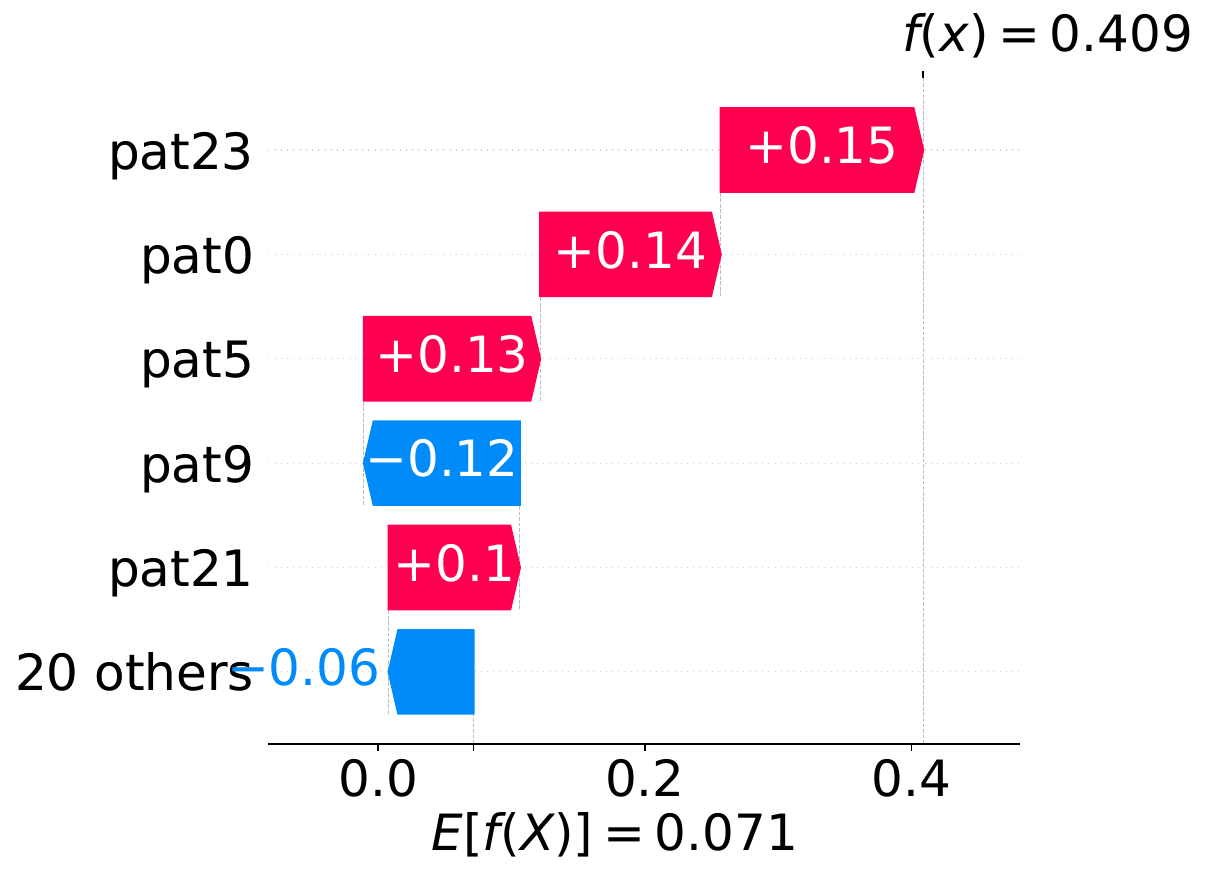}}
 \end{minipage}

 \caption{Example of SHAP applied to an EIN prediction from the ToxCast dataset.}
 \label{fig:shap-toxcast}

 \end{minipage}
 \begin{minipage}{.3\tw}

  % --------------------------------------------------
  % Fig: Decision tree
  % --------------------------------------------------

  \includegraphics[width=.9\tw]{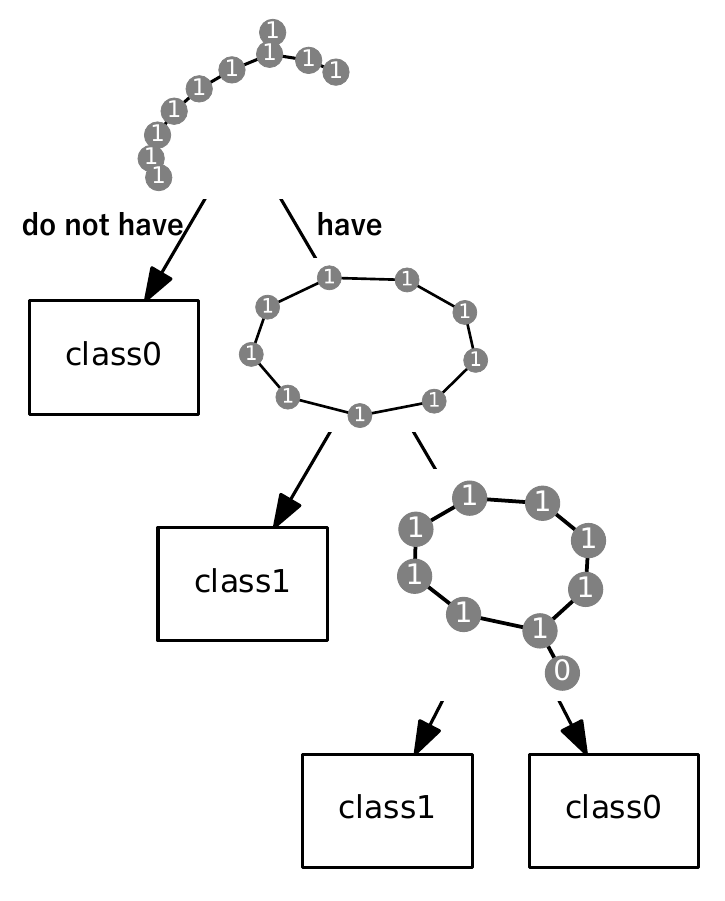}

  \caption{Decision tree on Cycle\_XOR}
  \label{decision-tree_cycle89XOR}
 \end{minipage}

\end{figure*}

% --------------------------------------------------
% Table: Pruning rate
% --------------------------------------------------
\begin{table*}[t]
 \caption{Mean pruning rates of EIN-$L_{0,2}$.}
 \label{tab:pruning-rates}
 \vspace{-.5em}
 \begin{center}
  \scalebox{.85}{
\begin{tabular}{lcccccccc}
        & BZR & COX2 & DHFR & ENZYMES & PTC\_MR & ToxCast & Cycle & Cycle\_XOR \\ \hline
 % WorkingSetSize & 13 & 11 & 21 & 21 & 15 & 10 & 38 & 40 \\
        \# Traverse & 1832 & 1727 & 978 & 9738 & 120 & 466 & 52 & 93 \\
        \# all subgraphs & 112319 & 76418 & 108486 & 1157661 & 61111 & 72734 & 31828 & 39883 \\
        Pruning rate & 0.9837 & 0.9774 & 0.9910 & 0.9916 & 0.9980 & 0.9936 & 0.9983 & 0.9977 \\
      \end{tabular}
  % \begin{tabular}{lcccccccc}
  %  & BZR & COX2 & DHFR & ENZYMES & ToxCast & SIDER & Cycle & Cycle\_XOR \\ \hline
  %  Working set size & 76 & 146 & 91 & 164 & 60 & 193 & 4 & 50 \\ 
  %  \# Traverse nodes & 12637 & 5060 & 20596 & 45829 & 3710 & 6122 & 3 & 828 \\ 
  %  \# all subgraphs $|\cH|$ & 112944 & 76120 & 108124 & 1133298 & 72734 & 101965 & 31953 & 40627 \\ 
  %  Pruning rates (\%) & 88.81 & 93.35 & 80.95 & 95.96 & 94.90 & 94.00 & 99.99 & 97.96 \\ 
  % \end{tabular}
    }
 \end{center}
\end{table*}

% --------------------------------------------------
% Table: Computational time of EIN
% --------------------------------------------------
\begin{table*}[t]
 \caption{Mean computational time for the entire Algorithm~\ref{alg:regularization-path} (sec).}
 \label{tab:computational-time-EIN}
 \vspace{-.5em}
  \begin{center}
    \scalebox{0.85}{
      \begin{tabular}{lrrrrrrrr}
        & BZR & COX2 & DHFR & ENZYMES & PTC\_MR & ToxCast & Cycle & Cycle\_XOR \\ \hline
       % UpdateTime & 1126 & 1047 & 4356 & 487 & 1723 & 4563 & 36163 & 7710 \\
        Traverse times & 50 & 59 & 338 & 1087 & 24 & 158 & 336 & 332 \\
        All times & 1311 & 1248 & 5079 & 1942 & 1829 & 5323 & 40200 & 11563 \\
      \end{tabular}
    }
  \end{center}
\end{table*}

% Post-hoc analysisの例を示す．
We here show examples of the post-hoc analysis using the trained EIN.
%
% 図~\ref{}はSHAPの適用例である (ToxCast)．
Figure~\ref{fig:shap-toxcast} is an example of SHAP for the EIN prediction on a negative instance of the ToxCast dataset (Note that the sign of $f$ is flipped in the plot (f)).
For SHAP, we used the python library \url{https://shap.readthedocs.io/en/latest/}, and see the document for detail. 
%
% これは正クラスの例であり，特に図~\ref{}~(a)が予測値を正に押し上げていると推定したことがわかる．
% The SHAP values are for the predicted class (positive class), and in this case, we can see that a subgraph in Fig.~\ref{fig:shap-toxcast}~(a) has a particularly strong contribution to the prediction.
From Fig.~\ref{fig:shap-toxcast}, we can see that subgraphs (a), (b), (c), and (e) contribute to the negative class, while (d) contributes to the positive class.
%
% \Figref{decision-tree_cycle89XOR}, \Figref{decision-tree_SIDER}には決定木による部分グラフの存在に基づくルールを示す．
% \Figref{decision-tree_cycle89XOR} shows example of fitted decision trees obtained by the same procedure as RF for the Cycle\_XOR and SIDER datasets.
\Figref{decision-tree_cycle89XOR} shows an example of a fitted decision tree to the trained EIN prediction for the Cycle\_XOR training and test datasets. 
% and \ref{decision-tree_SIDER}
%
% \Figref{decision-tree_cycle89XOR}はCycle89\_XORデータに対して作成した決定木である．
%
% まず1番目の分岐により，$G_p$, $G_n$を両方含まないサンプルを分類し，次に2, 3番目の分岐で，長さ8と9の閉路の有無に基づくXOR関係が抽出されている．
In Fig.~\ref{decision-tree_cycle89XOR}, the top node classifies a graph that do not have both of $H_p$ and $H_n$ as $y = 0$ (the top subgraph is included both in $H_p$ and $H_n$, but not in $H_{\rm padding}$), and the second and the third nodes consists of the XOR rule by the length 8 and 9 closed paths (the second node is the length 9 closed path and the third node is the length 8 closed path).
%
% これはデータの特性を正しく示すものであり，EINが真に重要な部分グラフを得られていることがわかる．
% Figure~\ref{decision-tree_SIDER} is an example of a decision tree obtained for the SIDER dataset.
Additional examples are shown in Appendix~\ref{app:post-hoc-extra}.

% --------------------------------------------------
\subsection{Pruning Rates}

% Table\ref{table-pruning-rates}には，maxpat=10における全$\lambda$での平均WorkingSetサイズと探索ノード数，全部分グラフ数を示す．
% Table~\ref{tab:pruning-rates} shows the mean working set size (mean over all $\lambda$), the mean number of the traversed nodes (mean over all $\lambda$), and the number of all the subgraphs $|\cH|$ for ${\rm maxpat} = 10$.
Table~\ref{tab:pruning-rates} shows the mean number of the traversed nodes (mean over all $s$), and the number of all the subgraphs $|\cH|$ for ${\rm maxpat} = 10$.
%
% 全部分グラフ数に対する探索ノード数の割合がPruning ratesであり，EINは最悪でも80\%程度，最高では95\%以上枝刈りによって探索空間を削減できていることがわかる．
% The pruning rates are defined by the ratio between the number of the traversed nodes and $|\cH|$. 
The pruning rates are defined by the ratio between them.
We can obviously see that a large amount of nodes are pruned, which indicates practical effectiveness of our pruning strategy. 
%
% また，WorkingSetのサイズは$10^2$程度であり，スパースな$\*\beta_H$が得られていることがわかる．
% Further, the mean size of the working set is at most a few hundreds, which means that $\*B$ is highly sparse during the optimization.

The computational time spent on Algorithm~\ref{alg:regularization-path} is shown in Table~\ref{tab:computational-time-EIN}, which is the total time for solving \eq{eq:objective} for all $\cS$.
The results with $\mr{maxpat}=10$ and $\tau_{\max} = 30$ for FFN is shown, which is a slower setting (other parameters are from the best setting in a sense of the validation accuracy). 
In the table, `Traverse Times' corresponds to the Traverse function in Algorithm~\ref{alg:regularization-path}.
We see that the times required for the subgraph enumerations depend on the datasets (depends on a variety of factors such as node sizes, edge sizes, and the pruning rate), but were not dominant for most of datasets.
Taking into account the fact that EIN handles all the possible subgraphs with the exact matching, we consider that EIN performs highly efficient computations. 
Computational time comparison with other methods are also reported in Appendix~\ref{app:time}.

\section{Conclusions}

% 本論文ではExact subgraph isomorphism network (EIN)を提案した．
We propose Exact subgraph Isomorphism Network (EIN). 
%
% EINはexact subgraph enumeration, neural network, and group sparse regularizationを融合した枠組みである．
EIN combines the exact subgraph enumeration, neural networks, and the $L_{0,2}$ group sparse constraint\footnote{Note that, based on a similar idea, $L_{1,2}$ based formulation is also possible, for which we show some detail and a reason we employ $L_{0,2}$ in Appendix~\ref{app:L12}.}. 
%
% これによりpredictive subgraphsを枝刈りして見つけられる, 精度を損なわずに．
We show that predictive subgraphs can be identified efficiently by pruning unnecessarily subgraphs during the IHT without sacrificing the quality of the model.
Further, the theoretical convergence of our alternating optimization is also established.
%
% 実験的にGNNと同精度．少数のグラフが実際に見つけられることを示した．
We demonstrated that EIN has sufficiently high prediction accuracy compared with well-known graph neural networks despite that EIN only uses a small number of selected subgraphs.
Examples of post-hoc analysis by which subgraph based decision rules can be estimated in an interpretable manner are also shown.
Further improvement on the scalability is one of our important future work.

\clearpage

% ==================================================
\bibliography{ref}
\biblstyle

\clearpage

\appendix

% ==================================================
% \section{Detail of Experimental Settings}
\section{Additional Information on Experiments}

% --------------------------------------------------
\subsection{Synthetic Dataset}
\label{app:syn-data}

% We first generate a random connected graph and randomly connected with one of $H_p$, $H_n$, and $H_{\rm padding}$. 
For both Cycle and Cycle\_XOR, we first generate a random graph by the following procedure.
%
% The node size of the initial random graph is randomly chosen from $5, 6, \ldots, 10$. 
The node size is randomly chosen from $5, 6, \ldots, 10$, and the number of edges are at most $20$, which are also randomly generated by choosing node pairs uniformly. 
From $H_p$, $H_n$, or $H_{\rm padding}$, randomly selected $n_{\rm connect} \in \{ 3, 4, 5, 6\}$ nodes are connected to a randomly selected node in the initial random graph.

% --------------------------------------------------
\subsection{Other Settings of EIN}
\label{app:settings-EIN}

We set the number of layers of FFN as $1$.
We set the terminate condition in the alternating update of Algorithm~\ref{alg:regularization-path} as that the validation loss does not improve $5$ times.
%
% In EIN+GIN, the message passing in GIN was set as $3$, the number of the middle unit was $16$, and the activation function was ReLU.
In EIN-$L_{0,2}$-GNN, the message passing in GIN was set as $3$, the number of the middle unit was $16$, and the activation function was ReLU. 
% (Memo: youkakuninn)

% ==================================================
\subsection{Other Examples of Post-hoc Analysis}
\label{app:post-hoc-extra}

% --------------------------------------------------
% Fig: Decision Tree XX
% --------------------------------------------------
\begin{figure}[t]
 \centering

 % \ig{.5}{figs/experiment/SIDER/decision-tree_maxdepth3_traintest_label-crop.pdf}
 \ig{.45}{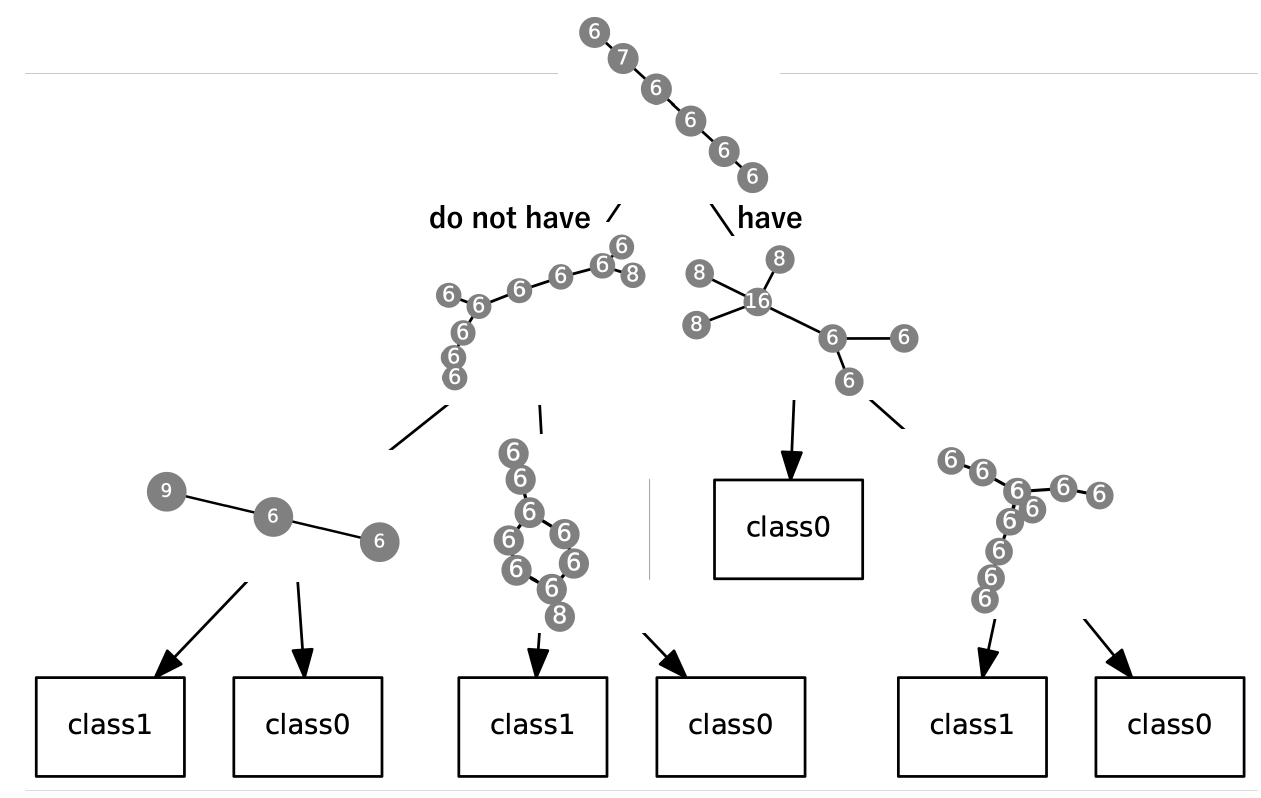}

 \caption{Decision tree on ToxCast.}
 \label{fig:decision-tree-sider}

\end{figure}

% --------------------------------------------------
% Fig: Feature importance
% --------------------------------------------------
\begin{figure}[t]
 \centering

 % \ig{.5}{figs/experiment/cycle89_XOR/rf_feature_importance_ver2.pdf}
 \ig{.5}{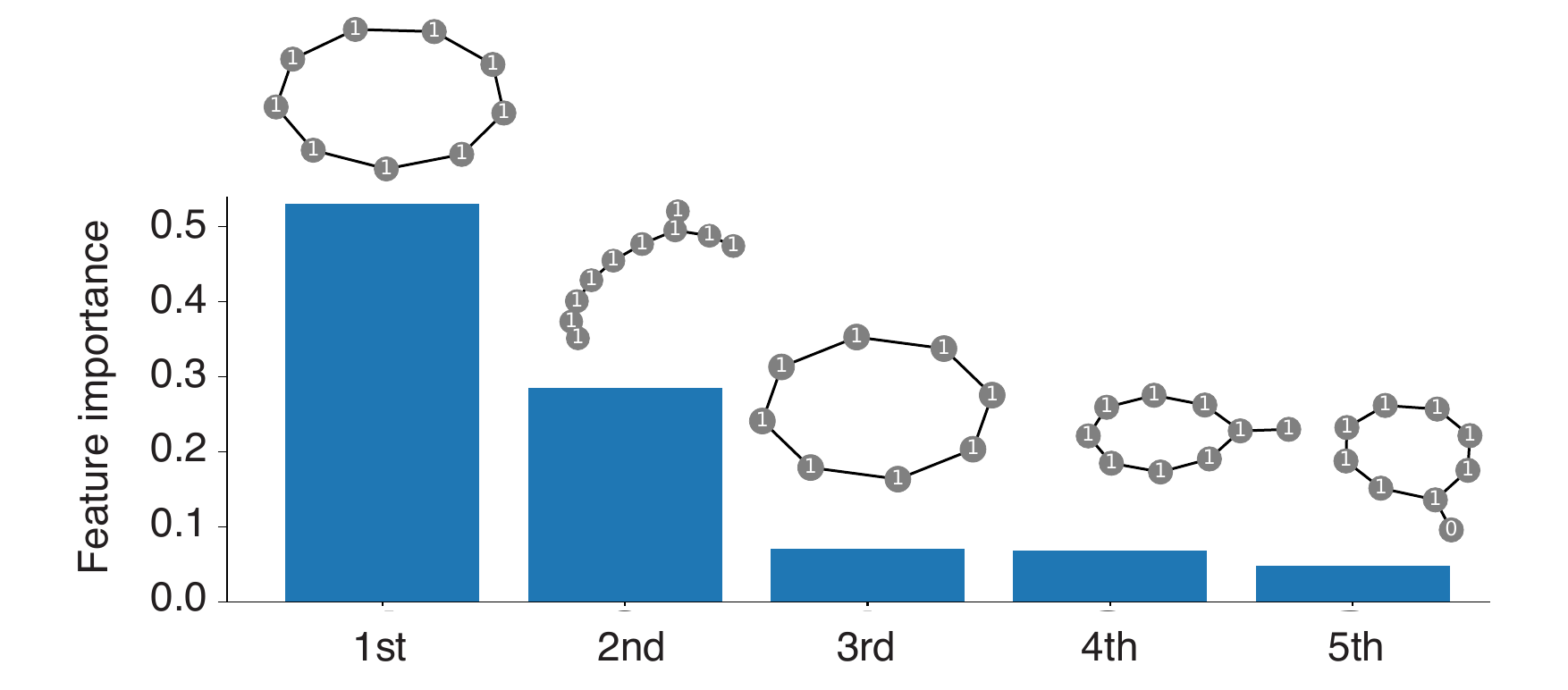}

 \caption{Subgraph importance estimated by RF on Cycle\_XOR dataset.}
 \label{fig:RF-cycle-xor}
\end{figure}

Figure~\ref{fig:decision-tree-sider} shows an example of a fitted decision tree to the trained EIN prediction for the ToxCast training and test datasets, which provides possible decision rule behind subgraphs and the target label.
%
% As another simple example, Fig.~\ref{fig:rf-cycle_xor} and \ref{fig:rf-tox} show subgraph importance estimated by Random Forest (RF) \citep{breiman2001random} for Cycle\_XOR and ToxCast datasets.
As another simple example of post-hoc analysis, Fig.~\ref{fig:RF-cycle-xor} shows subgraph importance estimated by Random Forest (RF) \citep{breiman2001random} for Cycle\_XOR. % and ToxCast dataset.
We fitted RF to a set of 
$(\*\psi_{\cS}(G_i), f(G_i))$
created by the training and test datasets 
% $\{ (\*\psi_{\cS}(G_i), f(G_i)) \}_{i=1}^n$, 
and the importance is evaluated by {\tt scikit-learn} feature importance of {\tt RandomForestRegressor}, which is based on mean decrease of impurity by that feature. 
% % \Figref{rf-Cycle89XOR}, \Figref{rf-ToxCast}にはRFによる部分グラフの重要度を示す．
% %
% % いずれも上位5つのみを示している．
Top five important subgraphs are shown in the figures. 
% %
% % \Figref{rf-Cycle89XOR}の結果から，重要な構造である$G_p$, $G_n$を含む部分グラフが上位の重要度をもつことがわかる（最も重要な部分グラフが長さ9の閉路，２番目が長さ８の閉路）．
Figure~\ref{fig:RF-cycle-xor} indicates that the closed paths of $H_p$ and $H_n$ are identified by EIN (the top subgraph is the length 9 closed path and the third subgraph is the length 8 closed path).

% ==================================================
\subsection{Computational Times}
\label{app:time}

Table~\ref{tab:computational-time-all} shows the computational times of all compared methods (For EIN-$L_{0,2}$, the values are same as Table~\ref{tab:computational-time-EIN}).
%
% For EIN, the results with $\mr{maxpat}=10$ and $\tau_{\max} = 30$ for FFN is shown, which is a slower setting (other parameters are from the best setting in a sense of the validation accuracy). 
%
Classical GNNs (GCN, GAT, GATv2, GIN, and PNA) were relatively fast, while more recent expressive GNNs (PPGN, GIN-AK, and ESC-GNN) were much slower than them.
Overall, EIN-$L_{0,2}$ was often slower than classical GNNs, while it was faster or similar to those expressive GNNs.

On the other hand, EIN-$L_{0,2}$-GNN is time-consuming in the current implementation. 
Our Algorithm~\ref{alg:regularization-path} employs the gradient descent for $\*\theta$ (not the `stochastic' gradient descent, unlike compared GNNs).
This full gradient update was time consuming for the combined GNN (backpropagation of message passing) because it was performed $\tau_{\max} (=30)$ times in each iteration of alternating update that was further performed for each $s \in \cS$.
Examining the possibility of computational time reduction of EIN-$L_{0,2}$-GNN (e.g., by the stochastic gradient with fewer epochs) is one of our future works.
% In this time, we employ the gradient descent for $\*\theta$ to be coincide with the update of other parameters (in particular, we consider that, for $\*B$, the full gradient is practically preferable because traversing the tree at every stochastic update might be computationally intensive). 
% %
% On the other hand, it is also possible to use the stochastic gradient descent for $\*\theta$, which is usually default choice for GNNs. % (though the theoretical analysis should be modified). 
% %
% Examining the possibility of improvement on computational time of EIN-$L_{0,2}$-GNN by the stochastic alternative with fewer epochs is one of our future works.

% --------------------------------------------------
% Table: Time for all methods
% --------------------------------------------------
\begin{table*}[t]
 \caption{Mean computational time of all methods (sec).} % 全モデルの平均CPU時間（10回の平均）
 \label{tab:computational-time-all}
  \begin{center}
    \scalebox{0.9}{
      \begin{tabular}{lrrrrrrrr}
        & BZR & COX2 & DHFR & ENZYMES & PTC\_MR & ToxCast & Cycle & Cycle XOR \\ \hline
        GCN & 756 & 806 & 1355 & 426 & 433 & 2006 & 770 & 717 \\
        GAT & 851 & 1108 & 1556 & 494 & 514 & 2394 & 855 & 975 \\
        GATv2 & 1420 & 1806 & 2459 & 841 & 850 & 4063 & 1470 & 1568 \\
        GIN & 967 & 1175 & 1784 & 539 & 541 & 2537 & 878 & 968 \\
        PNA & 1297 & 1366 & 2118 & 985 & 939 & 3573 & 1144 & 1503 \\ \hline
        PPGN & 41874 & 56586 & 97590 & 25621 & 12763 & 52513 & 32149 & 66286 \\
        GIN-AK & 5539 & 6186 & 7427 & 4295 & 3680 & 6542 & 5943 & 6909 \\
        ESC-GNN & 35988 & 44971 & 69881 & 22519 & 22309 & 84796 & 47337 & 59356 \\ \hline
        EIN-$L_{0,2}$ & 1311 & 1248 & 5079 & 1942 & 1829 & 5323 & 40200 & 11563 \\
%        EIN\_L1 & 443 & 1026 & 4029 & 2180 & 554 & 1990 & 9067 & 10206 \\
        EIN-$L_{0,2}$-GNN & 26850 & 109417 & 210749 & 24186 & 29841 & 238816 & 608597 & 530489 \\
%        EIN-GNN\_L1 & 11824 & 45896 & 21122 & 29212 & 21935 & 44750 & 225047 & 259099 \\
      \end{tabular}
    }
  \end{center}
\end{table*}

% ==================================================
% \section{LLM Usage}

% In this manuscript, LLM was only used to polish writing.

% ==================================================
\section{SIF by Frequency}
\label{app:SIF-frequency}

The frequency based SIF is used in \citep{yoshida2021distance} in the context of a subgraph pattern based distance metric learning.
They define $\#(H \sqsubseteq G_i)$ as the frequency of the subgraph $H$ contained in $G_i$, where nodes or edges among the counted subgraphs are not allowed.
Then, the feature value is defined as 
\begin{align*}
 \phi_H(G_i) = g( \#(H \sqsubseteq G_i) ),
\end{align*}
where $g$ is a monotonically non-decreasing and non-negative function such as identity function $g(x) = x$ or the indicator function $g(x)= \II(\#(H \sqsubseteq G_i) > 0)$ (\citet{yoshida2021distance} employed $g(x) = \log(1+x)$for their evaluation). 
They pointed out computing the frequency without overlapping is NP-complete \citep{schreiber2005frequency}, and approximate count is also provided (see \citep{yoshida2021distance} for detail). 
Our pruning theorem (Theorem~\ref{th:pruning}) holds for both the exact $\phi_H(G_i)$ and its approximation.

% ==================================================
% \section{Proximal Gradient}

% The proximal projection with respect to $\eta \lambda \| \cdot \|_2$ is defined as
% \begin{align*}
%  \mr{prox}(\*a) = 
%  \argmin_{\*b} \cbr{
%  \eta \lambda \| \*b \|_2 + \frac{1}{2} \| \*b - \*a \|_2^2
%  },
% \end{align*}
% which can be transformed into \eq{eq:prox-ope} \citep[e.g.,][]{parikh2014proximal}.

% ==================================================
\section{Proof of Theorem~\ref{th:pruning}}
\label{app:proof-pruning}

% $\*\beta_H$
% での微分をGMLの中間表現
% $\*h$
% で仲介して表現する: 
We first transform the derivatives of the loss function with respect to  
$\*\beta_H$
so that it is represented through the GML intermediate variable   
$\*h$:
\begin{align*}
 % \*g_H 
 \nabla_{\*\beta_{H}} \ell^t
 &= \pd{}{\*\beta_H} \sum_{i=1}^n \mr{loss}(y_i, f(G_i) ; \*B^t, \*b^t, \*\theta^t)
 \\ % --
 &= \sum_{i=1}^n \pd{\*h^\top}{\*\beta_H} \pd{ \mr{loss}(y_i, f(G_i) ; \*B^t, \*b^t, \*\theta^t) }{ \*h }. 
\end{align*}
%
% Since
% $\*h = \sum_{H \in \cH} \*\beta_H \psi_H(G_i) + \*b$, 
From the definition of $\*h$, 
we see
$\pd{\*h^\top}{\*\beta_H} = \psi_H(G_i) \*I_K$,
where 
$\*I_K$
is the
$K \times K$
identity matrix.
As a result, we have
\begin{align*}
 % \*g_H = \sum_{i=1}^n \pd{\ell(y_i, f(G_i))}{ \*h } \psi_H(G_i). 
 \nabla_{\*\beta_{H}} \ell^t = \sum_{i=1}^n \pd{ \mr{loss}(y_i, f(G_i) ; \*B^t, \*b^t, \*\theta^t) }{ \*h } \psi_H(G_i). 
\end{align*}
Since % $\delta_{ik} = \pd{\ell(y_i, f(G_i))}{h_k}$, 
$\delta^t_{ik} = \pd{\mr{loss}(y_i, f(G_i) ; \*B^t, \*b^t, \*\theta^t)}{h_k}$,
\begin{align}
% \| \*g_{H^\prm} \|_2 = 
 \| \nabla_{\*\beta_{H^\prm}} \ell^t \|_2 = 
 \sqrt{
 \sum_{k = 1}^K \rbr{ \sum_{i = 1}^n \delta^t_{ik} \psi_{H^\prm}(G_i) }^2 
 }.
 \label{eq:grad-norm}
\end{align}
% %
% For the inner sum, we divide each term depending on the sign of $\delta_{ik}$:
% \begin{align*}
%  \sum_{i = 1}^n \delta_{ik} \psi_{H^\prm}(G_i)
%  = 
%  \sum_{\{ i \mid \delta_{ik} < 0 \}} \delta_{ik} \psi_{H^\prm}(G_i)
%  + 
%  \sum_{\{ i \mid \delta_{ik} > 0 \}} \delta_{ik} \psi_{H^\prm}(G_i). 
% \end{align*}
%
From the definition, 
$0 \leq \psi_{H^\prm}(G_i) \leq \psi_H(G_i)$. 
Then, the upper and lower bound of the inner sum of \eq{eq:grad-norm}, i.e., 
$\sum_{i = 1}^n \delta^t_{ik} \psi_{H^\prm}(G_i)$,
can be derived as
\begin{align*}
 &
 \sum_{i = 1}^n \delta^t_{ik} \psi_{H^\prm}(G_i) 
 \leq 
 \sum_{\{ i \mid \delta^t_{ik} > 0 \}} \delta^t_{ik} \psi_{H^\prm}(G_i)
 \leq 
 \sum_{\{ i \mid \delta^t_{ik} > 0 \}} \delta^t_{ik} \psi_{H}(G_i), 
 \\
 &
 \sum_{i = 1}^n \delta^t_{ik} \psi_{H^\prm}(G_i) 
 \geq 
 \sum_{\{ i \mid \delta^t_{ik} < 0 \}} \delta^t_{ik} \psi_{H^\prm}(G_i)
 \geq 
 \sum_{\{ i \mid \delta^t_{ik} < 0 \}} \delta^t_{ik} \psi_{H}(G_i).
\end{align*}
Therefore,
\begin{align*}
 &
 \rbr{ \sum_{i = 1}^n \delta^t_{ik} \psi_{H^\prm}(G_i) }^2 
 \\
 & \leq 
 \max \cbr{
 \rbr{ \sum_{\{ i \mid \delta^t_{ik} > 0 \}} \delta^t_{ik} \psi_{H}(G_i) }^2, \ 
 \rbr{ \sum_{\{ i \mid \delta^t_{ik} < 0 \}} \delta^t_{ik} \psi_{H}(G_i) }^2
 },
\end{align*}
which results in \eq{eq:grad-UB}.

% ==================================================
% \section{\red{Lower Bound $\*\beta^{t+\frac{1}{2}}_H$ with respect to Step Length}}
\section{Lower Bound with respect to Step Length}
\label{app:lb-step-length}

\begin{align*}
 &
 \argmin_{ \gamma \in \Gamma}
 \frac{1}{\gamma}
 \| \*\beta_{H}^t - \gamma \nabla_{\*\beta_{H}} \ell^t \|_2
 \\ % ---
 &=
 \argmin_{ \gamma \in \Gamma}
 \frac{1}{\gamma^2}
 \| \*\beta_{H}^t - \gamma \nabla_{\*\beta_{H}} \ell^t \|_2^2
 \\ % ---
 &=
 \argmin_{ \gamma \in \Gamma}
 \| \*\beta_{H}^t \|_2^2 \frac{1}{\gamma^2}
 - 2 \inner{ \*\beta_{H}^t}{ \nabla_{\*\beta_{H}} \ell^t }
 \frac{1}{\gamma}
 \\ % ---
 &=
 \argmin_{ \gamma \in \Gamma}
 \| \*\beta_{H}^t \|_2^2
 \rbr{
 \frac{1}{\gamma} - 
 \frac{ \inner{ \*\beta_{H}^t}{ \nabla_{\*\beta_{H}} \ell^t } }{ \| \*\beta_{H}^t \|_2^2 }
 }^2
\end{align*}
Because of
$\gamma > 0$, 
if
$\inner{ \*\beta_{H}^t}{ \nabla_{\*\beta_{H}} \ell^t } > 0$, 
we see that
\begin{align*}
 \gamma = 
 \frac{ \| \*\beta_{H}^t \|_2^2 }{ \inner{ \*\beta_{H}^t}{ \nabla_{\*\beta_{H}} \ell^t } }
\end{align*}
is the minimizer.
On the other hand, if  
$\inner{ \*\beta_{H}^t}{ \nabla_{\*\beta_{H}} \ell^t } \leq 0$,
the minimizer is 
$\gamma = \max \Gamma$.
%
% 適当な最大Backtrack数の元で$\gamma$を選んでいる場合のように$\Gamma$が既知の有限集合なら$\gamma$に関する$\min$は全ての候補に対して計算しても構わない($|\Gamma|$も通常は大きくない)

Further, in practice, if $\Gamma$ is fixed beforehand, it is also possible to simply calculate 
$ \frac{1}{\gamma} \| \*\beta_{H}^t - \gamma \nabla_{\*\beta_{H}} \ell^t \|_2$
for all 
$\gamma \in \Gamma$.

% ==================================================
\section{Proof of Lemma~\ref{lem:step-length}}
\label{app:proof-step-len-existence}

\begin{proof}
First, we consider the update of $\*B$ based on \citep{damadi2022gradient}. 
For $\*B$, we assume that $\gamma^t_{\*B}$ is sufficiently small such that 
$L \leq \rho / \gamma^t_{\*B} < 1 / \gamma^t_{\*B}$, 
which can be satisfied by finite iterations of the backtrack search.

It is known that $L$-smoothness \eq{eq:Lipschitz-grad} is equivalent to 
\begin{align}
 \ell(\*B^\prm, \*b^\prm, \*\theta^\prm)
 \leq 
 \ell(\*B, \*b, \*\theta) + 
 \inner{ \pd{\ell(\*B, \*b, \*\theta)}{\*z} }{\*z^\prm - \*z} 
 + \frac{L}{2} \| \*z^\prm - \*z \|_2^2
 % \ \text{ for } \ \| \*B \|_0 \leq s, \| \*B^\prm \|_0 \leq s,
 \label{eq:RSS-all}
\end{align}
for
$\| \*B \|_0 \leq s$ and $\| \*B^\prm \|_0 \leq s$.
From \eq{eq:RSS-all}, 
\begin{align}
 \ell(\*B^{t+1}, \*b^t, \*\theta^t)
 & \leq 
 \ell(\*B^t, \*b^t, \*\theta^t) + 
 % \inner{ \pd{\ell(\*B^t, \*b^t, \*\theta^t)}{\*B} }{\*B^{t+1} - \*B^t} 
 \inner{ \nabla_{\*B} \ell^t }{\*B^{t+1} - \*B^t} 
 + \frac{L}{2} \| \*B^{t+1} - \*B^t \|_F^2
 % \notag \\ % --
 % & =
 % \ell(\*B^t, \*b^t, \*\Theta^t) + \inner{ \nabla_{\*B} \ell^t_S }{\*B^{t+1}_S - \*B^t_S} + \frac{L_{\*B}}{2} \| \*B^{t+1}_S - \*B^t_S \|_F^2,
 \label{eq:RSS-B}
\end{align}
where $\| \*B \|_0 \leq s$ and $\| \*B^\prm \|_0 \leq s$.
Let
\begin{align*}
S &= S_{t+1} \cup S_{t}, % \qquad \text{ \# $t$と$t+1$で選ばれてる変数の和集合}
\\
\ol{S}_{t+1} &= S \setminus S_{t+1}, % \qquad \text{ \# $t+1$では選ばれてないが$t$では選ばれてた}
\\
\ol{S}_{t} &= S \setminus S_{t}. % \qquad \text{ \# $t$では選ばれてないが$t+1$では選ばれる}
\end{align*}
When these indices are used as a subscript of $\*B$, it represents a submatrix of $\*B$ consisting of columns specified by these indices.
%  S = (S_{t+1} \cap S_t) \cup \ol{S}_t \cup \ol{S}_{t+1}
%
From the definition of the hard thresholding $H_s$, 
\begin{align*}
 \*B^{t+1}_{S_{t+1}} &= \*B^{t+1}_{S_{t+1}} - \gamma^t_{\*B} \nabla_{\*B} \ell^t_{S_{t+1}}, 
 % \*G^t_{S_{t+1}}
 \\
 \*B^{t+1}_{\ol{S}_{t+1}} &= \*0.
\end{align*}
By combining these with \eq{eq:RSS-B}, 
\begin{align}
 & 
 \ell(\*B^{t+1}, \*b^t, \*\theta^t)
 - 
 \ell(\*B^{t}, \*b^t, \*\theta^t)
 \notag \\ % ---
 &\leq
 \inner{ \nabla_{\*B} \ell^t_S }{\*B^{t+1}_S - \*B^t_S} + \frac{L}{2} \| \*B^{t+1}_S - \*B^t_S \|_F^2
 \notag \\ % ---
 &=
 \inner{\nabla_{\*B} \ell^t_{S_{t+1}}}{ \*B^{t+1}_{S_{t+1}} - \*B^t_{S_{t+1}}}_F - 
 \inner{\nabla_{\*B} \ell^t_{\ol{S}_{t+1}}}{ \*B^t_{\ol{S}_{t+1}}}_F 
 \notag \\
 & \qquad + \frac{L}{2} \| \*B^{t+1}_{S_{t+1}} - \*B^t_{S_{t+1}} \|^2_F
 + \frac{L}{2} \| \*B^t_{\ol{S}_{t+1}} \|^2_F
 \notag \\ % --- 
 &=
 - \gamma^t_{\*B} \| \nabla_{\*B} \ell^t_{S_{t+1}} \|_F^2
 - \inner{\nabla_{\*B} \ell^t_{\ol{S}_{t+1}}}{ \*B^t_{\ol{S}_{t+1}}}_F 
 \notag \\ 
 & \qquad + \frac{L}{2} (\gamma^t_{\*B} )^2 \| \nabla_{\*B} \ell^t_{S_{t+1}} \|_F^2
 + \frac{L}{2} \| \*B^t_{\ol{S}_{t+1}} \|^2_F
 \notag \\ % ---
 &=
 - \rbr{ 1 - \frac{ \gamma^t_{\*B} L }{2} } \gamma^t_{\*B} \| \nabla_{\*B} \ell^t_{S_{t+1}} \|_F^2
 + \frac{L}{2} \| \*B^t_{\ol{S}_{t+1}} \|^2_F
 - \inner{\nabla_{\*B} \ell^t_{\ol{S}_{t+1}}}{ \*B^t_{\ol{S}_{t+1}}}_F 
 \label{eq:ub-strong-smooth-B}
\end{align}
By using $L \leq 1/\gamma_{\*B}^t$, for the second and the third terms of \eq{eq:ub-strong-smooth-B}, 
\begin{align}
 &
 \frac{L}{2} \| \*B^t_{\ol{S}_{t+1}} \|^2_F
 - \inner{\nabla_{\*B} \ell^t_{\ol{S}_{t+1}}}{ \*B^t_{\ol{S}_{t+1}}}_F  
 \notag \\ % ---
 & \leq
 \frac{1}{2 \gamma^t_{\*B}} \| \*B^t_{\ol{S}_{t+1}} \|^2_F
 - \inner{ \nabla_{\*B} \ell^t_{\ol{S}_{t+1}} }{ \*B^t_{\ol{S}_{t+1}}}_F  
 \notag \\ % ---
 & =
 \frac{1}{2 \gamma^t_{\*B}}
 \nbr{
 \*B^t_{\ol{S}_{t+1}} - \gamma^t_{\*B} \nabla_{\*B} \ell^t_{\ol{S}_{t+1}}
 }_F^2
 - \frac{\gamma^t_{\*B}}{2} \nbr{ \nabla_{\*B} \ell^t_{\ol{S}_{t+1}} }_F^2
 \label{eq:ub-strong-smooth-B-sub}
\end{align}
%
% $H_s$の定義から明らかに(選ばれてる方がノルムが大きい)
From the definition of the hard thresholding $H_s$, for
$\forall p \in \ol{S}_{t+1}$
and
$\forall q \in \ol{S}_{t}$,  
we see 
\begin{align*}
 \| \*B^t_p - \gamma^t_{\*B} \nabla_{\*B} \ell^t_p \|_2 
 & \leq 
 \| \*B^t_q - \gamma^t_{\*B} \nabla_{\*B} \ell^t_q \|_2
 \\ % ---
 &=
 \gamma^t_{\*B} \| \nabla_{\*B} \ell^t_q \|_2
 % \forall q \in S_{t+1}. 
\end{align*}
% \begin{align*}
%  \| \*B^t_p - \gamma^t_{\*B} \nabla_{\*B} L^t_p \|_2 
%  \leq 
%  \| \*B^t_q - \gamma^t_{\*B} \nabla_{\*B} L^t_q \|_2
%  \text{ for }
%  \forall 
%  p \not \in S_{t+1},
%  \forall q \in S_{t+1}. 
% \end{align*}
% さらにその中でも以下について考える
% \begin{align*}
%  \nbr{ \rbr{ \*B^t_{\ol{S}_{t+1}} - \gamma^t_{\*B} (\nabla_{\*B} L^t)_{\ol{S}_{t+1}} }_i }_2
%  \leq
%  \nbr{ \rbr{ \*B^t_{\ol{S}_{t}} - \gamma^t_{\*B} (\nabla_{\*B} L^t)_{\ol{S}_{t}} }_j }_2
%  = 
%  \nbr{ \rbr{ \gamma^t_{\*B} (\nabla_{\*B} L^t)_{\ol{S}_{t}} }_j }_2
%  \ \text{ for } \forall i, j
% \end{align*}
% また，
Since 
\begin{align*}
 \abr{ \ol{S}_{t+1} } = s - \abr{ \ol{S}_{t+1} \cap \ol{S}_{t}} = \abr{ \ol{S}_{t} }, 
 % \qquad \text{ \# いなくなったものの数と新しく入ってくる数が同じだから}
\end{align*}
%
%よって（同じ数だけ２乗和とっても不等式関係はそのままなので）
we obtain an upper bound of the first term of \eq{eq:ub-strong-smooth-B-sub} as
\begin{align}
 \| \*B^t_{\ol{S}_{t+1}} - \gamma^t_{\*B} \nabla_{\*B} \ell^t_{\ol{S}_{t+1}} \|^2_F
 \leq
 (\gamma^t_{\*B})^2 \| \nabla_{\*B} \ell^t_{\ol{S}_{t}} \|^2_F. 
 \label{eq:ub-old-vars-B}
\end{align}
By applying 
\eq{eq:ub-old-vars-B}
and 
\eq{eq:ub-strong-smooth-B-sub}
to 
\eq{eq:ub-strong-smooth-B}, 
\begin{align*}
 &
 \ell(\*B^{t+1}, \*b^t, \*\theta^t) - \ell(\*B^{t}, \*b^t, \*\theta^t) 
 \\ % ---
 & \leq
 - \rbr{ 1 - \frac{ \gamma^t_{\*B} L }{2} } \gamma^t_{\*B} \| \nabla_{\*B} \ell^t_{S_{t+1}} \|_F^2
 + \frac{ \gamma^t_{\*B} }{ 2 } \| \nabla_{\*B} \ell^t_{\ol{S}_{t}} \|^2_F
 - \frac{ \gamma^t_{\*B} }{ 2 } \| \nabla_{\*B} \ell^t_{\ol{S}_{t+1}} \|_F^2
 \\ % ---
 & =
 - \rbr{ 1 - \frac{ \gamma^t_{\*B} L }{2} } \gamma^t_{\*B} \| \nabla_{\*B} \ell^t_{S_t \cap S_{t+1}} \|_F^2
 - \rbr{ \frac{1}{2} - \frac{ \gamma^t_{\*B} L }{2} } \gamma^t_{\*B} \| \nabla_{\*B} \ell^t_{\ol{S_t}} \|_F^2
 % + \frac{ \gamma^t_{\*B} }{ 2 } \| (\nabla_{\*B} \ell^t)_{\ol{S}_{t}} \|^2_F
 \\
 & \qquad
 - \frac{\gamma^t_{\*B}}{2} \| \nabla_{\*B} \ell^t_{\ol{S}_{t+1}} \|_F^2
 \\ % ---
 & \leq
 - \rbr{ \frac{1}{2} - \frac{ \gamma^t_{\*B} L }{2} } \gamma^t_{\*B}
 \rbr{
 \| \nabla_{\*B} \ell^t_{S_t \cap S_{t+1}} \|_F^2
 + \| \nabla_{\*B} \ell^t_{\ol{S_t}} \|_F^2
 + \| \nabla_{\*B} \ell^t_{\ol{S}_{t+1}} \|_F^2
 }
 \\ % ---
 & =
 - \rbr{ 1 - \gamma^t_{\*B} L } \frac{ \gamma^t_{\*B} }{ 2 }
 \| \nabla_{\*B} \ell^t_S \|_F^2. 
\end{align*}
Further, using
$L \leq \rho / \gamma^t_{\*B}$, 
we obtain 
% ($1 / \gamma^t_{\*B}$だと消えてしまう．もう一段階余裕を見てる)
\begin{align}
 \ell(\*B^{t}, \*b^t, \*\theta^t) - \ell(\*B^{t+1}, \*b^t, \*\theta^t) 
 & \geq 
 \rbr{ 1 - \gamma^t_{\*B} L } \frac{ \gamma^t_{\*B} }{ 2 } \| \nabla_{\*B} \ell^t_S \|_F^2
 \notag \\ % ---
 & \geq 
 \rbr{ 1 - \rho } \frac{ \gamma^t_{\*B} }{ 2 } \| \nabla_{\*B} \ell^t_S \|_F^2
 \notag \\ % ---
 & \geq 
 c_{\*B} \rbr{ 1 - \rho } \frac{ \gamma^t_{\*B} }{ 2 } \| \nabla_{\*B} \ell^t_S \|_F^2. 
 \label{eq:sufficient-decrease-B}
\end{align}
As a result, we see that \eq{eq:BT-condition-B} can be satisfied by a sufficiently small $\gamma_{\*B}^t$.
% となり，十分小さいステップ幅$\gamma^t_{\*B}$をとればBacktrackの条件が成立することがわかる．
Further, since \eq{eq:BT-condition-B} holds when $L \leq \rho / \gamma_{\*B}^t$, which means $\gamma_{\*B}^t \leq \rho / L$. 
We also see 
$\gamma_{\*B}^t \geq \rho^2 / L$ 
(the backtrack stops before $\gamma_{\*B}^t$ becomes smaller than $\rho^2 / L$).

% $L \leq 1/\gamma^t_{\*b}$, $\gamma^t_{\*b}$
For $\*b$, the condition \eq{eq:BT-condition-b-inner} can be directly obtained from $L$-smoothness and the assumption 
$L \leq 1/\gamma_{\*b}^{t,\tau}$ 
as
\begin{align*}
 \ell(\*B^t, \*b^{t,\tau+1}, \*\theta^t)
 & \leq 
 \ell(\*B^t, \*b^{t,\tau}, \*\theta^t) + 
 \inner{ \nabla_{\*b} \ell^{t,\tau} }{\*b^{t,\tau+1} - \*b^{t,\tau}} 
 \notag \\
 & \qquad
 + \frac{1}{2 \gamma^{t,\tau}_{\*b}} \| \*b^{t,\tau+1} - \*b^{t,\tau} \|^2_2
 \\ % ---
 & =
 \ell(\*B^t, \*b^{t,\tau}, \*\theta^t) -
 \gamma^{t,\tau}_{\*b}
 \nbr{ \nabla_{\*b} \ell^{t,\tau} }_2^2
 + \frac{ \gamma^{t,\tau}_{\*b} }{2} \nbr{ \nabla_{\*b} \ell^{t,\tau} }_2^2, 
\end{align*}
from which we see 
\begin{align*}
 \ell(\*B^t, \*b^{t,\tau+1}, \*\theta^t) - \ell(\*B^t, \*b^{t,\tau}, \*\theta^t) 
 \geq
 \frac{ \gamma^{t,\tau}_{\*b} }{2} \nbr{ \nabla_{\*b} \ell^{t,\tau} }_2^2
 \geq
 c_{\*b} \frac{ \gamma^{t,\tau}_{\*b} }{2} \nbr{ \nabla_{\*b} \ell^{t,\tau} }_2^2
\end{align*}
For $\*\theta$, \eq{eq:BT-condition-theta-inner} can be obtained by the same procedure.
We also see 
$\gamma_{\*b}^{t,\tau} \geq \rho / L$
and
$\gamma_{\*\theta}^{t,\tau} \geq \rho / L$
because of the same reason as 
$\gamma_{\*B}^t \geq \rho^2 / L$.
\end{proof}

% ==================================================
\section{Definition of Subdifferential}
\label{app:subdifferential}

Let
$f: \RR^d \to \RR \cup \{ \infty \}$
be a proper and lower semicontinous function.
Fr{\'e}chet subdifferential of $f$ is 
\begin{align*}
 \hat{\partial} f(\*x) =
 \cbr{
 \*g \mid \lim_{\*y \to \*x} \inf_{\*y \neq \*x}
 \frac{ f(\*y) - f(\*x) - \inner{\*g}{\*y - \*x} }{ \| \*y - \*x \|_2 }
 \geq 0
 }. 
\end{align*}
If $\*x \notin \mr{dom} f$, then $\hat{\partial} f(\*x) = \emptyset$.
If $f$ is a $C^1$ function, 
$\hat{\partial} f(\*x)$
coincides with the usual gradient.
% This can be seen as a generalization of the well-known subgradient of a convex function: 
% $\{ \*g \mid f(\*y) \geq f(\*x) + \*g^\top (\*y - \*x) \}$.
%
However, it is known that, under this definition, $\hat{\partial} f(\*x)$ becomes unnaturally empty in some cases (e.g., $f(x) = - |x|$ at $x = 0$), and also $\hat{\partial} f(\*x)$ is not a closed mapping. 
The limiting subdifferential of $f$ is defined through the limit values of Fr{\'e}chet subdifferential: 
\begin{align*}
 \partial f(\*x) = 
 \cbr{
 \*g \mid \exists \*x^k \to \*x, f(\*x^k) \to f(\*x), \*g^k \in \hat{\partial} f(\*x^k) \to \*g
 }. 
\end{align*}
The limiting subdifferential satisfies Fermat's rule, i.e.,
$\partial f(\*x) \ni \*0$
if $\*x$ is a local minimum, and is widely used for analysis of a nonsmooth nonconvex optimization.

% ==================================================
\section{Proof of Theorem~\ref{theo:theo-stationary-point}}
\label{app:proof-stationary-point}

% --------------------------------------------------
\subsection{Auxiliary Lemmas}

% Theorem~\ref{theo:theo-stationary-point}の証明のために以下のlemma~\ref{lem:obj-low}, \ref{lem:lb-outer-update}, \ref{lem:grad-upper}を先に証明する．
To prove Theorem~\ref{theo:theo-stationary-point}, we first prove the auxiliary lemmas~\ref{lem:obj-low}, \ref{lem:lb-outer-update}, \ref{lem:grad-upper}, and \ref{lem:accumulation}

% --------------------------------------------------
% Lemma: Lower bound of obj and upper bound of param update
% --------------------------------------------------
\begin{lem}
 There exists a constant $C_1$ that satisfies
 \begin{align}
  % \red{\ell(\*B^{t+1}, \*b^{t+1}, \*\theta^{t+1}) - \ell(\*B^{t}, \*b^{t}, \*\theta^t) }
  \ell(\*B^{t}, \*b^{t}, \*\theta^t) - \ell(\*B^{t+1}, \*b^{t+1}, \*\theta^{t+1}) 
  \geq 
  C_1 \nbr{ \*z^{t+1} - \*z^t }_2^2.  
  \label{eq:obj-low}
 \end{align}
 \label{lem:obj-low}
\end{lem}

\begin{proof}
The left-hand-side of \eq{eq:obj-low} can be decomposed as
\begin{align}
 & \ell(\*B^{t}, \*b^{t}, \*\theta^t) - \ell(\*B^{t+1}, \*b^{t+1}, \*\theta^{t+1}) 
 \notag \\ 
 & =  
 \rbr{
 \ell(\*B^t, \*b^t, \*\theta^t) - \ell(\*B^{t+1}, \*b^{t}, \*\theta^t) 
 }
 \notag \\ 
 & \qquad + \rbr{
 \ell(\*B^{t+1}, \*b^{t}, \*\theta^{t}) - \ell(\*B^{t+1}, \*b^{t+1}, \*\theta^t) 
 }
 \notag \\
 & \qquad + \rbr{
 \ell(\*B^{t+1}, \*b^{t+1}, \*\theta^{t}) - \ell(\*B^{t+1}, \*b^{t+1}, \*\theta^{t+1}) 
 }. 
 \label{eq:obj-low-decomposed}
\end{align}
We derive lower bounds of each one of the three terms.

Consider decomposing the update of $\*B$ into $S_t {\cap} S_{t+1}$, $\ol{S}_t$, and $\ol{S}_{t+1}$: 
\begin{align*}
 \| \*B^{t+1} - \*B^{t} \|_F^2
 &= 
 \| H_s(\*B^{t}_S - \gamma_{\*B}^t \nabla_{\*B} L^t_S) - \*B_S^t \|_F^2
 \\ % ---
 &=
 \| \*B^{t}_{S_t {\cap} S_{t+1}} - \gamma_{\*B}^t \nabla_{\*B} L^t_{S_t {\cap} S_{t+1}} - \*B^t_{S_t {\cap} S_{t+1}} \|_F^2
 \\
 & \qquad 
 + 
\| \gamma_{\*B}^t \nabla_{\*B} L^t_{\ol{S}_t} \|_F^2
 + 
 \| \*B^t_{\ol{S}_{t+1}} \|^2_F
 \\ % ---
 &=
 \| \gamma_{\*B}^t \nabla_{\*B} L^t_{S_t {\cap} S_{t+1}} \|_F^2
 + 
\| \gamma_{\*B}^t \nabla_{\*B} L^t_{\ol{S}_t} \|_F^2
 + 
 \| \*B^t_{\ol{S}_{t+1}} \|^2_F
\end{align*}
An upper bound of the last term is 
\begin{align*}
 \| \*B^t_{\ol{S}_{t+1}} \|_F
 & \leq 
 \| \*B^t_{\ol{S}_{t+1}} - \gamma_{\*B}^t \nabla_{\*B} \ell^t_{\ol{S}_{t+1}} \|_F
 +
 \| \gamma_{\*B}^t \nabla_{\*B} \ell^t_{\ol{S}_{t+1}} \|_F
 \\ % ---
 &\leq
 \gamma^t_{\*B} \| \nabla_{\*B} \ell^t_{\ol{S}_{t}} \|_F
 +
 \| \gamma_{\*B}^t \nabla_{\*B} \ell^t_{\ol{S}_{t+1}} \|_F. 
\end{align*}
The last inequality is from \eq{eq:ub-old-vars-B}.
% 最後の不等式は\eq{eq:ub-old-vars-B}より．
%
By the above inequality, we can further create an upper bound of 
$\| \*B^{t+1} - \*B^{t} \|_F^2$ 
as
% そのため
\begin{align}
 \| \*B^{t+1} - \*B^{t} \|_F^2 
 & \leq 
 \| \gamma_{\*B}^t \nabla_{\*B} L^t_{S_t {\cap} S_{t+1}} \|_F^2
 + 
 \| \gamma_{\*B}^t \nabla_{\*B} L^t_{\ol{S}_t} \|_F^2
 \notag \\
 & \qquad +
 \rbr{
 \gamma^t_{\*B} \| \nabla_{\*B} L^t_{\ol{S}_{t}} \|_F
 +
 \| \gamma_{\*B}^t \nabla_{\*B} L^t_{\ol{S}_{t+1}} \|_F }^2 
 \notag \\ % ---
 & \leq 
 \| \gamma_{\*B}^t \nabla_{\*B} L^t_{S_t {\cap} S_{t+1}} \|_F^2
 + 
 \| \gamma_{\*B}^t \nabla_{\*B} L^t_{\ol{S}_t} \|_F^2
 \notag \\
 & \qquad +
 2 (\gamma^t_{\*B})^2 \| \nabla_{\*B} L^t_{\ol{S}_{t}} \|_F^2
 + 
 2 (\gamma_{\*B}^t)^2 \| \nabla_{\*B} L^t_{\ol{S}_{t+1}} \|_F^2 
 % \rbr{
 % \gamma^t_{\*B} \| (\nabla_{\*B} L^t)_{\ol{S}_{t}} \|_F
 % +
 % \gamma_{\*B}^t \nabla_{\*B} L^t_{\ol{S}_{t+1}} \| 
 % }^2 
 \notag \\ % ---
 & 
 \leq 
 (\gamma^t_{\*B})^2
 \| \nabla_{\*B} L^t_{S} \|_F^2
 + 
 4 (\gamma^t_{\*B})^2
 \| \nabla_{\*B} L^t_{{S}} \|_F^2
 \notag \\ % ---
 & =
 5 (\gamma^t_{\*B})^2
 \| \nabla_{\*B} L^t_S \|_F^2
 \notag \\ % ---
 & \leq 
 5 (\gamma^t_{\*B})^2
 \frac{ 1 }{ c_{\*B} (1 - \rho) } \frac{ 2 }{ \gamma^t_{\*B} }
 \rbr{ \ell(\*B^t, \*b^t, \*\theta^t) - \ell(\*B^{t+1}, \*b^{t}, \*\theta^t) }
 % \to \*0
 % \label{eq:ub-update-B}
 \label{eq:obj-lower-B}
\end{align}
The second inequality is from the inequality of arithmetic and geometric means and the last inequality is from the backtrack condition \eq{eq:BT-condition-B}.

The improvement of the objective function in the inner loop of $\*b$ and $\*\theta$ can be bounded by
\begin{align}
 &
 \ell(\*B^{t+1}, \*b^{t}, \*\theta^t) - \ell(\*B^{t+1}, \*b^{t+1}, \*\theta^t) 
 \notag \\
 &= 
 \ell(\*B^{t+1}, \*b^{t,0}, \*\theta^t) - \ell(\*B^{t+1}, \*b^{t,\tau_{\max}}, \*\theta^t)
 \notag \\ % ---
 & =
 \sum_{\tau = 0}^{\tau_{\max} - 1}
 \ell(\*B^{t+1}, \*b^{t,\tau}, \*\theta^t) - \ell(\*B^{t+1}, \*b^{t,\tau+1}, \*\theta^t)
 \notag \\ % ---
 & 
 \geq 
 \sum_{\tau = 0}^{\tau_{\max}}
 c_{\*b} \frac{ \gamma^{t,\tau}_{\*b} }{2} \nbr{ \nabla_{\*b} \ell^{t,\tau} }_2^2, 
 \label{eq:sufficient-decrease-b}
\end{align}	      
and
\begin{align}
 &
 \ell(\*B^{t+1}, \*b^{t+1}, \*\theta^t) - \ell(\*B^{t+1}, \*b^{t+1}, \*\theta^{t+1}) 
 \notag \\
 &= 
 \ell(\*B^{t+1}, \*b^{t+1}, \*\theta^{t,0}) - \ell(\*B^{t+1}, \*b^{t+1}, \*\theta^{t,\tau_{\max}})
 \notag \\ % ---
 & =
 \sum_{\tau = 0}^{\tau_{\max} - 1}
 \ell(\*B^{t+1}, \*b^{t+1}, \*\theta^{t,\tau}) - \ell(\*B^{t+1}, \*b^{t+1}, \*\theta^{t,\tau+1})
 \notag \\ % ---
 & 
 \geq 
 \sum_{\tau = 0}^{\tau_{\max}}
 c_{\*b} \frac{ \gamma^{t,\tau}_{\*\theta} }{2} \nbr{ \nabla_{\*\theta} \ell^{t,\tau} }_2^2. 
 \label{eq:sufficient-decrease-theta}
\end{align}

Let $\gamma_{\max} = \max \Gamma$. 
From
\eq{eq:sufficient-decrease-b}
and
$\*b^{t,\tau+1} = \*b^{t,\tau} - \gamma^{t,\tau}_{\*b} \nabla_{\*b} L^{t,\tau}$, 
% \begin{align*}
%  L(\*B^{t+1}, \*b^{t,\tau}, \*\theta^t) - L(\*B^{t+1}, \*b^{t,\tau+1}, \*\theta^t) &\geq 
%  c_{\*b} \frac{ \gamma^{t,\tau}_{\*b} }{2} \nbr{ \nabla_{\*b} L^{t,\tau} }_2^2
%  \\ % ---
%  &=
%  c_{\*b} \frac{ 1 }{2 \gamma^{t,\tau}_{\*b}} 
%  \nbr{ \*b^{t,\tau+1} - \*b^{t,\tau} }_2^2
% \end{align*}
\begin{align}
 &
 \ell(\*B^{t+1}, \*b^{t}, \*\theta^t) - \ell(\*B^{t+1}, \*b^{t+1}, \*\theta^t) 
 \notag \\
 &
 \geq 
 \sum_{\tau = 0}^{\tau_{\max} - 1}
 c_{\*b} \frac{ \gamma^{t,\tau}_{\*b} }{2} \nbr{ \nabla_{\*b} \ell^{t,\tau} }_2^2
 \notag \\ % ---
 &=
 \sum_{\tau = 0}^{\tau_{\max} - 1}
 c_{\*b} \frac{ 1 }{2 \gamma^{t,\tau}_{\*b}} 
 \nbr{ \*b^{t,\tau+1} - \*b^{t,\tau} }_2^2
 \notag \\ % ---
 &\geq
 c_{\*b} \frac{ 1 }{2 \gamma_{\max}} 
 \sum_{\tau = 0}^{\tau_{\max} - 1}
 \nbr{ \*b^{t,\tau+1} - \*b^{t,\tau} }_2^2
 \notag \\ % ---
 &\geq
 c_{\*b} \frac{ 1 }{2 \gamma_{\max}} 
 \frac{1}{\tau_{\max}} 
 \rbr{ \sum_{\tau = 0}^{\tau_{\max} - 1} \nbr{ \*b^{t,\tau+1} - \*b^{t,\tau} }_2 }^2 
 \notag \\
 % & \qquad \text{ \# コーシーシュワルツ ($(\sum a_j)^2 \leq n \sum a_j^2$, $(\*a^\top \*1 \leq \| \*a \|^2 \| \*1 \|^2 )$)}
 % \\ % ---
 &\geq
 c_{\*b} \frac{ 1 }{2 \gamma_{\max}} 
 \frac{1}{\tau_{\max}} 
 \nbr{ \sum_{\tau = 0}^{\tau_{\max} - 1} \rbr{ \*b^{t,\tau+1} - \*b^{t,\tau} } }_2^2
 \notag \\ % ---
 &=
 c_{\*b} \frac{ 1 }{2 \gamma_{\max}} 
 \frac{1}{\tau_{\max}} 
 \nbr{ \*b^{t,\tau_{\max}} - \*b^{t,0} }_2^2
 % \\
 % &\geq
 % c_{\*b} \frac{ 1 }{2 \gamma_{\max}}  
 % \nbr{ \*b^{t,\tau+1} - \*b^{t,\tau} }_2^2
 \label{eq:obj-lower-b}
\end{align}
% From the third to the fourth line, we used a relation from the Cauchy-Schwarz inequality 
From the fourth to the fifth line, we used a relation from the Cauchy-Schwarz inequality 
(for $a_i \in \RR$, $\sum_{i=1}^n a_i^2 = (\sum_{i=1}^n a_i^2) (\sum_{i=1}^n 1^2) / n \geq (\sum_{i=1}^n (a_i \times 1) )^2 / n$). 
% $n \| \*a \|_2^2 = \| \*a \|_2^2 \| \*1 \|_2^2 \geq ( \inner{\*a}{\*1} )^2$ 
% for $a_i \in \RR$.
%
% 同じく
The same inequality for $\*\theta$ is derived by using
\eq{eq:sufficient-decrease-theta}
and
% と
$\*\theta^{t,\tau+1} = \*\theta^{t,\tau} - \gamma^{t,\tau}_{\*b} \nabla_{\*\theta} \ell^{t,\tau}$ as
\begin{align}
 \ell(\*B^{t+1}, \*b^{t+1}, \*\theta^t) - \ell(\*B^{t+1}, \*b^{t+1}, \*\theta^{t+1}) 
 \geq 
 c_{\*\theta} \frac{ 1 }{2 \gamma_{\max}} 
 \frac{1}{\tau_{\max}} 
 \nbr{ \*\theta^{t,\tau_{\max}} - \*\theta^{t,0} }_2^2
 \label{eq:obj-lower-theta}
\end{align}

By substituting \eq{eq:obj-lower-B}, \eq{eq:obj-lower-b}, and \eq{eq:obj-lower-theta} into \eq{eq:obj-low-decomposed}, we obtain
\begin{align*}
 &
 \ell(\*B^{t+1}, \*b^{t+1}, \*\theta^{t+1}) - \ell(\*B^{t}, \*b^{t}, \*\theta^t) 
 \notag \\ % ---
 & \geq 
 c_{\*B} \rbr{ 1 - \rho } \frac{1}{10 \gamma^t_{\*B}}
 \| \*B^{t+1} - \*B^{t} \|_F^2
 +
 c_{\*b} \frac{ 1 }{2 \gamma_{\max}} \frac{1}{\tau_{\max}} 
 \nbr{ \*b^{t+1} - \*b^{t} }_2^2
 \\
 & \qquad +
 c_{\*\theta} \frac{ 1 }{2 \gamma_{\max}} \frac{1}{\tau_{\max}} 
 \nbr{ \*\theta^{t+1} - \*\theta^{t} }_2^2
 \notag \\ % ---
 & \geq 
 C_1
 \nbr{ \*z^{t+1} - \*z^t }_2^2,
 % \label{eq:sufficient-decrease-Z}
\end{align*}
where 
$C_1 = \min\{ 
c_{\*B} \rbr{ 1 - \rho } \frac{1}{10 \gamma^t_{\*B}}, 
c_{\*b} \frac{ 1 }{2 \gamma_{\max}} \frac{1}{\tau_{\max}}, 
c_{\*\theta} \frac{ 1 }{2 \gamma_{\max}} \frac{1}{\tau_{\max}} \}$ 

\end{proof}

% --------------------------------------------------
% Lemma: Lower bound of outer update of b and theta
% --------------------------------------------------
\begin{lem}
% $\| \*b^{t, \tau_{\max}} - \*b^{t, \tau_{\max}-1} \|_2^2 \leq C_{\*b} \| \*b^{t+1} - \*b^{t} \|_2^2$
% for some constant $C_{\*b}$.
 For some constant {$C_{\*b}$} and {$C_{\*\theta}$},
 \begin{align}
  \nbr{ \nabla_{\*b} \ell^t }_2
  & \leq 
  C_{\*b}
  \| \*b^t - \*b^{t+1} \|_2, 
  \label{eq:lb-outer-b-update} \\ % --- 
  \nbr{ \nabla_{\*\theta} \ell^t }_2
  & \leq 
  C_{\*\theta}
  \| \*\theta^t - \*\theta^{t+1} \|_2.
  \label{eq:lb-outer-theta-update}
 \end{align} 
 \label{lem:lb-outer-update}
\end{lem}
\begin{proof}
 From \eq{eq:BT-condition-b-inner}, we have
 \begin{align*}  
  \ell(\*B^{t+1}, \*b^{t}, \*\theta^t) - \ell(\*B^{t+1}, \*b^{t+1}, \*\theta^t)
  & \geq 
  \ell(\*B^{t+1}, \*b^{t,0}, \*\theta^t) - \ell(\*B^{t+1}, \*b^{t,1}, \*\theta^t)
  \\
  & \geq 
  c_{\*b}
  \frac{\gamma^{t,0}_{\*b}}{2} \| \nabla_{\*b} \ell^{t,0} \|_2^2,
 \end{align*}
 and from the multi-dimensional variant of the mean value theorem, there exists $\*\xi \in (\*b^t, \*b^{t+1})$ that satisfies
 \begin{align*}
  \ell(\*B^{t+1}, \*b^{t}, \*\theta^t) - \ell(\*B^{t+1}, \*b^{t+1}, \*\theta^t)
  = 
  \inner{ \pd{\ell(\*B^{t+1}, \*\xi, \*\theta^t) }{ \*b } }{ \*b^{t} - \*b^{t+1} }.
  % \| \*b^{t+1} - \*b^{t} \|_2^2
 \end{align*} 
 Therefore, we see
 \begin{align}
  c_{\*b}
  \frac{\gamma^{t,0}_{\*b}}{2} \| \nabla_{\*b} \ell^{t,0} \|_2^2
  & \leq 
  \inner{ \pd{\ell(\*B^{t+1}, \*\xi, \*\theta^t) }{ \*b } }{ \*b^{t} - \*b^{t+1} }
  \notag \\ % ---
  & \leq 
  \nbr{ \pd{\ell(\*B^{t+1}, \*\xi, \*\theta^t)}{ \*b } }_2
  d_b,
  % \nbr{ \*b^{t} - \*b^{t+1} }_2.
  \label{eq:mean-value-theo-b}
 \end{align}
 where 
 $d_b \coloneqq \nbr{ \*b^{t} - \*b^{t+1} }_2$.
 On the other hand, from $L$-smoothness, we see
 \begin{align*}
  \nbr{ \pd{\ell(\*B^{t+1}, \*\xi, \*\theta^t)}{ \*b } - \pd{\ell(\*B^{t+1}, \*b^{t}, \*\theta^t)}{ \*b } }_2
  \leq 
  L \nbr{ \*\xi - \*b^{t} }_2
  \leq 
  L d_b, % \nbr{ \*b^t - \*b^{t+1} }_2,
 \end{align*}
 from which we further obtain
 \begin{align}
  & 
  \nbr{ \pd{\ell(\*B^{t+1}, \*\xi, \*\theta^t)}{ \*b } }_2
  \notag \\
  &=
  \nbr{ \pd{\ell(\*B^{t+1}, \*\xi, \*\theta^t)}{ \*b } - \pd{\ell(\*B^{t+1}, \*b^{t}, \*\theta^t)}{ \*b } + \pd{\ell(\*B^{t+1}, \*b^{t}, \*\theta^t)}{ \*b }}_2
  \notag \\ % ---
  & \leq 
  \nbr{ \pd{\ell(\*B^{t+1}, \*\xi, \*\theta^t)}{ \*b } - \pd{\ell(\*B^{t+1}, \*b^{t}, \*\theta^t)}{ \*b } }_2
  + 
  \nbr{ \pd{\ell(\*B^{t+1}, \*b^{t}, \*\theta^t)}{ \*b } }_2
  \notag \\ % ---
  & \leq 
  L d_b % \nbr{ \*b^t - \*b^{t+1} }_2
  + 
  \nbr{ \pd{\ell(\*B^{t+1}, \*b^{t}, \*\theta^t)}{ \*b } }_2. 
  \label{eq:ub-mean-value-theo-b}
 \end{align}
 By combining \eq{eq:mean-value-theo-b} and \eq{eq:ub-mean-value-theo-b}, 
 \begin{align*}
  c_{\*b}
  \frac{\gamma^{t,0}_{\*b}}{2} \| \nabla_{\*b} \ell^{t,0} \|_2^2
  & \leq
  \rbr{
  L d_b
  + 
  \nbr{ \pd{\ell(\*B^{t+1}, \*b^{t}, \*\theta^t)}{ \*b } }_2
  } d_b
  \\ % ---
  &=
  \rbr{
  L d_b
  + 
  \nbr{ 
  \nabla_{\*b} \ell^{t,0}
  % \pd{\ell(\*B^{t+1}, \*b^{t,0}, \*\theta^t)}{ \*b } 
  }_2
  } d_b
 \end{align*}
 from which we obtain
 \begin{align*}
  L d_b^2 +
  \nbr{ 
  % \pd{\ell(\*B^{t+1}, \*b^{t+1}, \*\theta^t)}{ \*b } 
  \nabla_{\*b} \ell^{t,0}
  }_2 d_b
  -
  c_{\*b} \frac{\gamma^{t,\tau_{\max}-1}_{\*b}}{2} \| \nabla_{\*b} \ell^{t,0} \|_2^2
  \geq 0. 
 \end{align*}
 By regarding the left-hand-side as a quadratic function of $d_b$, 
 \begin{align*}
  d_b & \geq 
  \frac{ - \nbr{ \nabla_{\*b} \ell^{t,0} }_2  
         + \sqrt{ \nbr{ \nabla_{\*b} \ell^{t,0}}_2^2
                 + 4 L c_{\*b} \frac{\gamma^{t,0}_{\*b}}{2} \| \nabla_{\*b} \ell^{t,0} \|_2^2 } 
  }
  { 2 L }
  \\ % ---
  & = 
  \rbr{ \sqrt{1 + 2 L c_{\*b} \gamma^{t,0}_{\*b} } - 1 }
  \frac{ \nbr{ \nabla_{\*b} \ell^{t,0} }_2 }{ 2 L } 
 \end{align*}
 Therefore, 
 \begin{align*}
  \nbr{ \nabla_{\*b} \ell^{t,0} }_2
  & \leq 
  \frac{
  2 L
  }{
  \sqrt{1 + 2 L c_{\*b} \gamma^{t,0}_{\*b} } - 1 
  }
  \| \*b^t - \*b^{t+1} \|_2
  \\ % ---
  & \leq 
  \frac{
  2 L
  }{
  \sqrt{1 + 2 \rho c_{\*b}  } - 1 
  }
  \| \*b^t - \*b^{t+1} \|_2.
 \end{align*}
 The last inequality is from $\gamma_{\*b}^{t,0} \geq \rho / L$. 
 The same derivation can be applied to $\*\theta$. 
\end{proof}

% --------------------------------------------------
% Lemma: Parameter update can be a lower bound of gradient
% --------------------------------------------------
\begin{lem}
 There exists 
 % $\nabla_{\*z} \tilde{\ell}(\*B^{t+1}, \*b^{t+1}, \*\theta^{t+1}) \in \partial \tilde{\ell}(\*z)$}
 $\nabla_{\*z} \tilde{\ell}(\*z^{t+1}) \in \partial \tilde{\ell}(\*z^{t+1})$
 such that 
 \begin{align}
  % \| \nabla_{\*z} \tilde{\ell}(\*B^{t+1}, \*b^{t+1}, \*\theta^{t+1}) \|_2^2
  \| \nabla_{\*z} \tilde{\ell}(\*z^{t+1}) \|_2^2
  \leq 
  C_2 \| \*z^{t+1} - \*z^t \|_2^2,
  \label{eq:grad-upper}
 \end{align}
 where $C_2$ is a constant.
 \label{lem:grad-upper}
\end{lem}

\begin{proof}
The hard thresholding can be written as an optimization problem:
\begin{align}
 \*B^{t+1} 
 &= H_s(\*B^t - \gamma_{\*B} \nabla_{\*B} \ell^t ) 
 \notag \\
 &= 
 \argmin_{\*B}
 \frac{1}{2 \gamma_{\*B}^t}
 \| \*B - (\*B^t - \gamma_{\*B}^t \nabla_{\*B} \ell^t) \|_F^2
 + \delta_s(\*B).
 \label{eq:IHT-as-prox}
\end{align}
% where
% \begin{align*}
%  \delta_s(\*B)
%  =
%  \begin{cases}
%   0, & \text{ if } \| \*B \|_{2,0} \leq s,
%   \\
%   \infty, & \text{ otherwise. }
%  \end{cases}
% \end{align*}
%
Taking the subdifferential of \eq{eq:IHT-as-prox}
\begin{align*}	
 \*0 \in &
 \frac{1}{\gamma_{\*B}^t}
 ( \*B^{t+1} - (\*B^t - \gamma_{\*B}^t \nabla_{\*B} \ell^t) )
 + \partial \delta_s(\*B^{t+1})
 \\ % ---
 - \frac{1}{\gamma_{\*B}^t}
 ( \*B^{t+1} - (\*B^t - \gamma_{\*B}^t \nabla_{\*B} \ell^t) )
 &\in \partial \delta_s(\*B^{t+1}),
\end{align*}
where 
$\partial \delta_s(\*B)$
is the subdifferential of 
$\delta_s(\*B)$.
By using 
$\partial \delta_s(\*B^{t+1})$,
the subdifferential 
$\partial \tilde{\ell}(\*z^{t+1})$
can be written as
\begin{align*}
 \partial \tilde{\ell}(\*z^{t+1}) 
 &= \nabla_{\*B} \ell^{t+1}
 % \ell(\*B^{t+1}, \*b^{t+1}, \*\theta^{t+1}) 
 + \partial \delta_s(\*B^{t+1})
 \\ % --
 & \ni
 \nabla_{\*B} \ell^{t+1}
 % \ell(\*B^{t+1}, \*b^{t+1}, \*\theta^{t+1}) 
 - \frac{1}{\gamma_{\*B}^t}
 ( \*B^{t+1} - (\*B^t - \gamma_{\*B}^t \nabla_{\*B} \ell^t) )
 \\
 & =
 \nabla_{\*B} \ell^{t+1}
 % \ell(\*B^{t+1}, \*b^{t+1}, \*\theta^{t+1}) 
 - \nabla_{\*B} \ell^t - \frac{1}{\gamma_{\*B}^t} ( \*B^{t+1} - \*B^t )
 \eqqcolon \nabla_{\*B} \tilde{\ell}^{t+1}
\end{align*}
From the triangle inequality and $L$-smoothenss, 
\begin{align}
 \| \nabla_{\*B} \tilde{\ell}^{t+1} \|_F^2
 & \leq
 \rbr{
 \| \nabla_{\*B} \ell^{t+1} - \nabla_{\*B} \ell^t \|_F 
 + \frac{1}{\gamma_{\*B}^t} \| \*B^{t+1} - \*B^t \|_F }^2
 \notag \\ % ---
 & \leq
 \rbr{
 L \| \*z^{t+1} - \*z \|_F 
 + \frac{1}{\gamma_{\*B}^t} \| \*B^{t+1} - \*B^t \|_F }^2
 \notag \\ % ---
 & \leq 2 L^2
 \rbr{ \| \*B^{t+1} - \*B^t \|^2_F 
 + \| \*b^{t+1} - \*b \|_2^2 
 + \| \*\theta^{t+1} - \*\theta^t \|_2^2 }
 \notag \\
 &
 \qquad + \frac{2}{(\gamma_{\*B}^t)^2} \| \*B^{t+1} - \*B^t \|_F^2
 \notag \\ % ---
 & \leq 
 2 \rbr{ L^2 + \frac{1}{ (\gamma_{\*B}^t)^2 }} 
 \| \*B^{t+1} - \*B^t \|_F^2 
 \notag \\
 & \qquad + 2 L^2 \| \*b^{t+1} - \*b \|_2^2 
 + 2 L^2 \| \*\theta^{t+1} - \*\theta^t \|_2^2 
 \label{eq:grad-upper-B}
\end{align}

An upper bound of the derivative of $\*b$ is
\begin{align}
 &
 \| \nabla_{\*b} \ell(\*B^{t+1}, \*b^{t+1}, \*\theta^{t+1}) \|_2^2
 \notag \\ % ---
 & \leq 
 \bigl(
 \| \nabla_{\*b} \ell(\*B^{t+1}, \*b^{t+1}, \*\theta^{t+1}) - \nabla_{\*b} \ell(\*B^{t+1}, \*b^t, \*\theta^{t}) \|_2
 \notag \\
 & \qquad + 
 \| \nabla_{\*b} \ell(\*B^{t+1}, \*b^t, \*\theta^t) \|_2
 \bigr)^2
 \notag \\ % ---
 & \leq 
 2 {L}^2 \rbr{ \| \*b^{t+1} - \*b^t \|_2^2 + \| \*\theta^{t+1} - \*\theta^t \|_2^2 }
 + 2 
 \| \nabla_{\*b} \ell^t \|_2^2
 \notag \\
 & \leq 
 % 2 {L}^2 \rbr{ \| \*b^{t+1} - \*b^t \|_2^2 + \| \*\theta^{t+1} - \*\theta^t \|_2^2 } + 2 C^2_{\*b} \| \*b^t - \*b^{t+1} \|_2^2
 2 \rbr{ L^2 + C_{\*b}^2 } \| \*b^{t+1} - \*b^t \|_2^2 
 + 2 {L}^2 \| \*\theta^{t+1} - \*\theta^t \|_2^2 
 \label{eq:grad-upper-bias}
\end{align}
The last inequality is from \eq{eq:lb-outer-b-update}.

Similarly, for $\*\theta$,
\begin{align}
 &
 \| \nabla_{\*\theta} \ell(\*B^{t+1}, \*b^{t+1}, \*\theta^{t+1}) \|_2^2
 \notag \\ % ---
 & \leq 
 \bigl\{
 \| \nabla_{\*\theta} \ell(\*B^{t+1}, \*b^{t+1}, \*\theta^{t+1}) - \nabla_{\*\theta} \ell(\*B^{t+1}, \*b^{t+1}, \*\theta^t) \|_2
 \notag \\
 & \qquad + 
 \| \nabla_{\*\theta} \ell(\*B^{t+1}, \*b^{t+1}, \*\theta^t) \|_2 \bigr\}^2
 \notag \\ % ---
 & \leq 
 2 L^2 \| \*\theta^{t+1} - \*\theta^t \|_2^2
 + 
 2 \| \nabla_{\*\theta} \ell^t \|_2^2
 % \frac{2}{(\gamma_{\*\theta}^{t,\tau_{\max}-1})^2} \| \*\theta^{t,\tau_{\max}} - \*\theta^{t,\tau_{\max}-1} \|_2^2
 \notag \\ % ---
 & \leq 
 2 \rbr{L^2 + C_{\*\theta}^2} 
 \| \*\theta^{t+1} - \*\theta^t \|_2^2
 % \\ % ---
 % & \leq 
 % 2 \rbr{L^2 - \frac{1}{(\gamma_{\*\theta}^{t,\tau_{\max}-1})^2}} 
 % C_{\*\theta} \| \*\theta^{t+1} - \*\theta^{t-1} \|_2^2.
 \label{eq:grad-upper-theta}
\end{align}
The last inequality is from \eq{eq:lb-outer-theta-update}.

By combining \eq{eq:grad-upper-B}, \eq{eq:grad-upper-bias}, and \eq{eq:grad-upper-theta}, we see
\begin{align*}
 &
 % \| \nabla_{\*z} \tilde{\ell}(\*B^{t+1}, \*b^{t+1}, \*\theta^{t+1}) \|_2^2
 \| \nabla_{\*z} \tilde{\ell}(\*z^{t+1}) \|_2^2
 \\
 & = 
 \| \nabla_{\*B} \tilde{\ell}^{t+1} \|_F^2
 + \| \nabla_{\*b} \ell(\*B^{t+1}, \*b^{t+1}, \*\theta^{t+1}) \|_2^2
 + \| \nabla_{\*\theta} \ell(\*B^{t+1}, \*b^{t+1}, \*\theta^{t+1}) \|_2^2 
 \\ % ---
 & \leq
 2 \rbr{ L^2 + \frac{1}{ (\gamma_{\*B}^t)^2 }} 
 \| \*B^{t+1} - \*B^t \|_F^2 
 + 2 {L}^2 \| \*b^{t+1} - \*b^t \|_2^2 
 + 2 {L}^2 \| \*\theta^{t+1} - \*\theta^t \|_2^2 
 \\
 & \qquad
 % + 2 \rbr{{L}^2 - \frac{1}{(\gamma_{\*b}^{t,\tau_{\max}-1})^2}} C_{\*b} 
 + 2 \rbr{L^2 + C_{\*b}^2 } 
 \| \*b^{t+1} - \*b^{t} \|_2^2
 + 2 L^2 \| \*\theta^{t+1} - \*\theta^t \|_2^2
 \\
 & \qquad
 % + 2 \rbr{{L}^2 - \frac{1}{(\gamma_{\*\theta}^{t,\tau_{\max}-1})^2}} C_{\*\theta} 
 + 2 \rbr{L^2 + C_{\*\theta}^2 }
 \| \*\theta^{t+1} - \*\theta^t \|_2^2
 \\ % ---
 & =
 2 \rbr{ L^2 + \frac{1}{ (\gamma_{\*B}^t)^2 }} 
 \| \*B^{t+1} - \*B^t \|_F^2 
 \\
 & \qquad +
 % \cbr{ 2 L^2 + 2 \rbr{{L}^2 - \frac{1}{(\gamma_{\*b}^{t,\tau_{\max}-1})^2} C_{\*b} } }
 2 \rbr{ 2 L^2 + C_{\*b}^2 }
 \| \*b^{t+1} - \*b^t \|_2^2
 \\
 & \qquad +
 % \cbr{ 4 L^2 + 2 \rbr{{L}^2 - \frac{1}{(\gamma_{\*\theta}^{t,\tau_{\max}-1})^2}}  C_{\*\theta} }
 2 \rbr{ 3 L^2 + C_{\*\theta}^2 }
 \| \*\theta^{t+1} - \*\theta^t \|_2^2
 \\ % ---
 & \leq
 C_{2} \| \*z^{t+1} - \*z^t \|_2^2,
\end{align*}
where 
$C_2 = \max\cbr{
2 \rbr{ L^2 + \frac{1}{ (\gamma_{\*B}^t)^2 }}, 
2 \rbr{ 2 L^2 + C_{\*b}^2 }, 
2 \rbr{ 3 L^2 + C_{\*\theta}^2 }
}$. 
% $C_2 = \max\cbr{
% 2 \rbr{ L^2 + \frac{1}{ (\gamma_{\*B}^t)^2 }}, 
% { 2 L^2 + 2 \rbr{{L}^2 - \frac{1}{(\gamma_{\*b}^{t,\tau_{\max}-1})^2} C_{\*b} } }, 
% { 4 L^2 + 2 \rbr{{L}^2 - \frac{1}{(\gamma_{\*\theta}^{t,\tau_{\max}-1})^2}}  C_{\*\theta} }
% }$. 

\end{proof}

% --------------------------------------------------
% Lemma: accumulation point (z^tは発散せず勾配0の点へ行く部分列を含む)
% --------------------------------------------------
\begin{lem}
 % $\*z^t \to \*z^*$
 $\*z^t$ has 
 at least one accumulation point $z^*$
 % a convergent subsequence to $z^*$
 such that 
 $\| \nabla_{\*z} \tilde{\ell}(\*z^*) \|_2^2 = 0$. 
 \label{lem:accumulation}
\end{lem}
\begin{proof}
 Summing \eq{eq:obj-low} from $t = 1$ to $T - 1$, we obtain
 \begin{align*}
  % \ell(\*B^{t}, \*b^{t}, \*\theta^t) - \ell(\*B^{t+1}, \*b^{t+1}, \*\theta^{t+1}) 
  % \geq 
  % C_1 \nbr{ \*z^{t+1} - \*z^t }_2^2
  \ell(\*B^{0}, \*b^{0}, \*\theta^0) - \ell(\*B^{T}, \*b^{T}, \*\theta^{T}) 
  \geq 
  C_1 \sum_{t=1}^T \nbr{ \*z^t - \*z^{t-1} }_2^2
 \end{align*}
 Considering $T \to \infty$ and combining with \eq{eq:grad-upper}
 \begin{align*}
  \infty & >
  \lim_{T \to \infty}
  \ell(\*B^{0}, \*b^{0}, \*\theta^0) - \ell(\*B^{T}, \*b^{T}, \*\theta^{T}) 
  \\ % ---
  & \geq
  C_1 \lim_{T \to \infty} \sum_{t=1}^T \nbr{ \*z^t - \*z^{t-1} }_2^2
  \\ % ---
  & \geq
  \frac{C_1}{C_2}
  \lim_{T \to \infty} \sum_{t=1}^T
  \| \nabla_{\*z} \tilde{\ell}(\*z^{t}) \|_2^2
 \end{align*}
 From the monotone convergence theorem, 
 % $\nbr{ \*z^t - \*z^{t-1} }_2^2 \to 0$
 % and
 $\| \nabla_{\*z} \tilde{\ell}(\*z^t) \|_2^2 \to 0$. 
 Since the sequence $z^t$ is assumed be bounded, we obtain the claim of the lemma.
 %
 % In other words,
 % $\*z^t \to \*z^*$
 % such that 
 % $\| \nabla_{\*z} \tilde{\ell}(\*z^*) \|_2^2 = 0$. 
 %
 %
 % $\| \nabla_{\*z} \tilde{\ell}(\*z^{t+1}) \|_2^2
 %  \leq 
 %  C_2 \| \*z^{t+1} - \*z^t \|_2^2,  $
\end{proof}

% --------------------------------------------------
\subsubsection*{Proof of Theorem~\ref{theo:theo-stationary-point}}

In the proof, we use the well-known Kurdyka-{\L}ojasiewicz property.
%
% --------------------------------------------------
% Definition: KL Property
% --------------------------------------------------
\begin{defini} 
 (Kurdyka-{\L}ojasiewicz property)
 A function $\varphi(\*u)$ satisfies the {\it Kurdyka-{\L}ojasiewicz} (KL) property at a point 
 $\bar{\*u} \in \mr{dom}(\partial \varphi)$
 if there exist 
 $\eta > 0$,
 a neighborhood 
 $\cB_{\rho}(\bar{\*u}) \coloneqq \{ \*u \mid \| \*u - \bar{\*u} \|_2 < \rho \}$,
 and a concave function 
 $\phi(s) = c s^{1 - \theta}$
 for some $c > 0$ and $\theta \in [0, 1)$ such that for any 
 $\*x \in \cB_{\rho}(\bar{\*u}) \cap \mr{dom}(\partial \varphi)$
 and
 $\varphi(\bar{\*u}) < \varphi(\*u) < \varphi(\bar{\*u}) + \eta$, 
 it holds 
 \begin{align*}
  \phi^\prm( | \varphi(\*u) - \varphi(\bar{\*u}) | ) \mr{dist}(\*0, \partial \varphi(\*x)) \geq 1
 \end{align*}
 where 
 $\mr{dom}(\partial \varphi) = \{ \*u \mid \partial \varphi(\*u) \neq \emptyset \}$
 and 
 $\mr{dist}(\*0,\partial \varphi(\*u)) = \min \{ \| \*v \|_2 \mid \*v \in \partial \varphi(\*u) \}$.
\end{defini}

The proof of Theorem~\ref{theo:theo-stationary-point} is as follows.
\begin{proof}
It is known that a lower semicontinuous `definable' function satisfies the KL property \citep{bolte2007clarke} (see Theorem~14 and Remark~8). 
Our $\ell$ is the cross-entropy loss function for a neural network with the sigmoid activation function.
These components are known as definable functions \citep{ding2025stochastic,kranz2026sad}, and the compositions of definable functions are also definable \citep{van1996geometric,attouch2010proximal}.
L0 penalty $\delta_s$ is a semialgebratic function \citep{bao2014ell0}. 
A semialgebratic function is definable \citep{dries1998tame}, and the sum of definable functions is also definable \citep{attouch2010proximal}.
As a result, $\tilde{\ell}$ is a lower semicontinuous definable function.

% 一旦，
% $\*z^t \in \cB_\rho(\*z^*)$ for $\forall t$
% を仮定する（このような系列の存在は後で証明する）．
Here, we temporarily assume  
$\*z^t \in \cB_\rho(\*z^*)$ 
for $\forall t$ (we will prove the existence of such a sequence later in the proof). 
% We use the well-known {\it Kurdyka-Lojasiewicz} (KL) property 
From the KL property, 
\begin{align*}
 \phi^\prm(\tilde{\ell}(\*z) - \ell^*)
 \cdot
 \mr{dist}(0, \partial \tilde{\ell}(\*z))
 & \geq 1
 \\ % ---
 \phi^\prm(\tilde{\ell}(\*z) - \ell^*)
 & \geq
 \frac{1}{ \mr{dist}(0, \partial \tilde{\ell}(\*z)) }
 \geq \frac{1}{ \|  { \nabla_{\*z} \tilde{\ell}(\*z) } \|_2 },
\end{align*}
where 
$\ell^* = \tilde{\ell}(\*z^*)$,
{$\forall \nabla_{\*z} \tilde{\ell}(\*z) \in \partial \tilde{\ell}(\*z)$},
and 
$\phi(s) = c s^{1 - \theta}$
is a concave function for some $c > 0$ and $\theta \in [0,1)$.
Since 
$\phi$
is concave,
$\phi(u) - \phi(v) \geq \phi^\prm(u)(u-v)$.
Substituting 
$u = \tilde{\ell}(\*z^t) - \ell^*$ 
and
$v = \tilde{\ell}(\*z^{t+1}) - \ell^*$
into this inequality, we obtain
\begin{align*}
 \phi(\tilde{\ell}(\*z^t) - \ell^*) - \phi(\tilde{\ell}(\*z^{t+1}) - \ell^*) 
 & \geq 
 \phi^\prm(\tilde{\ell}(\*z^t) - \ell^*) \cdot (\tilde{\ell}(\*z^t) - \tilde{\ell}(\*z^{t+1}))
 \\
 & \geq  
 \frac{
 \tilde{\ell}(\*z^t) - \tilde{\ell}(\*z^{t+1})
 }{
 \| {\nabla_{\*z} \tilde{\ell}(\*z^t)} \|_2
 % \mr{dist}(0, \partial \tilde{\ell}(\*z^t))
 }. 
\end{align*}
% $\tilde{L}$は実質$L$と書いてもよいが...
By using \eq{eq:obj-low} of lemma~\ref{lem:obj-low} and \eq{eq:grad-upper} of lemma~\ref{lem:grad-upper}, there exists 
${\nabla_{\*z} \tilde{\ell}(\*z^t)} \in \partial \tilde{\ell}(\*z^t)$
that satisfies
%
%$\tilde{L}$の十分減少\eq{eq:sufficient-decrease-Z}と
%$\phi^\prm(\tilde{L}(\*Z) - L^*) \geq 1 / (\mr{dist}(0, \partial \tilde{L}(\*Z)))$
%	      を使って
\begin{align*}
 \phi(\tilde{\ell}(\*z^t) - \ell^*) - \phi(\tilde{\ell}(\*z^{t+1}) - \ell^*) 
 \geq 
 \frac{ C_1 \nbr{ \*z^{t+1} - \*z^t }_2^2 }
 { \| { \nabla_{\*z} \tilde{\ell}(\*z^t)  } \|_2 }
 % { \| \nabla_{\*z} \tilde{\ell}(\*B^t, \*b^t, \*\theta^t) \|_2 }
 \geq 
 \frac{ C_1 \nbr{ \*z^{t+1} - \*z^t }_2^2 }
 { \sqrt{ C_2 } \| \*z^t - \*z^{t-1} \|_2 }
\end{align*}
By arranging this inequality, 
\begin{align}
 \nbr{ \*z^{t+1} - \*z^t }_2^2 
 & \leq 
 \frac{ \sqrt{C_2} }{ C_1 }
 \| \*z^t - \*z^{t-1} \|_2
 \rbr{
 \phi(\tilde{\ell}(\*z^t) - \ell^*) - \phi(\tilde{\ell}(\*z^{t+1}) - \ell^*) }
 \notag \\ % ---
% \end{align*}
%       相加相乗$2 \sqrt{A} \sqrt{B} \leq A + B$
%       \begin{align}
	% \nbr{ \*Z^{t+1} - \*Z^t }_2 \leq
 2 \nbr{ \*z^{t+1} - \*z^t }_2 
 & \leq
 \| \*z^t - \*z^{t-1} \|_2 
 + 
 \frac{ \sqrt{C_2} }{ C_1 }
 \rbr{
 \phi(\tilde{\ell}(\*z^t) - \ell^*) - \phi(\tilde{\ell}(\*z^{t+1}) - \ell^*) }.
 \label{eq:KL-ineq-sub1}
\end{align}
To derive the last inequality, the inequality of arithmetic and geometric means is used.
% \red{相加相乗$2 \sqrt{A} \sqrt{B} \leq A + B$.}
%
% $t = 1$から$t = T$まで足すと
The sum from $t = 1$ to $t = T$ becomes
\begin{align}
 2 \sum_{t = 1}^T \nbr{ \*z^{t+1} - \*z^t }_2 
 & \leq
 \sum_{t = 1}^T \| \*z^t - \*z^{t-1} \|_2 
 \notag \\
 & \qquad 
 + \sum_{t = 1}^T \frac{ \sqrt{ C_2 } }{ C_1 }
 \rbr{
 \phi(\tilde{\ell}(\*z^t) - \ell^*) - \phi(\tilde{\ell}(\*z^{t+1}) - \ell^*) }
 \notag \\ % ---
 \sum_{t = 1}^T \nbr{ \*z^{t+1} - \*z^t }_2 
 & \leq
 \| \*z^1 - \*z^{0} \|_2 
 \notag \\
 & \qquad + \frac{ \sqrt{ C_2 } }{ C_1 }
 \rbr{
 \phi(\tilde{\ell}(\*z^1) - \ell^*) - \phi(\tilde{\ell}(\*z^{T+1}) - \ell^*) }
 \notag \\ % ---
 &
 \leq
 \| \*z^1 - \*z^{0} \|_2 
 + \frac{ \sqrt{ C_2 } }{ C_1 } \phi(\tilde{\ell}(\*z^1) - \ell^*) 
 \label{eq:ub-z-sequence}
\end{align}
We see that the last upper bound does not depend on $T$, by which we obtain
\begin{align*}
 \lim_{T \to \infty}
 \sum_{t = 1}^T \nbr{ \*z^{t+1} - \*z^t }_2 < \infty. 
\end{align*}
Therefore, we obtain
$\lim_{t \to \infty} \nbr{ \*z^{t+1} - \*z^t }_2 = 0$, 
which is a necessary condition for the convergence of an infinite series. 
%
% $\*z^t$はコーシー列となる．
This means that $\*z^t$ is a Cauchy sequence.
From \eq{eq:grad-upper} of lemma~\ref{lem:grad-upper}, 
\begin{align*}
 \infty > 
 \lim_{T \to \infty}
 \sum_{t = 1}^T \nbr{ \*z^{t+1} - \*z^t }_2
 \geq 
 \lim_{T \to \infty}
 \sum_{t = 1}^T
 \frac{1}{C_2}
 \| { \nabla_{\*z} \tilde{\ell}(\*z^{t+1}) } \|_2^2. 
 % \| \nabla_{\*z} \tilde{L}(\*B^{t+1}, \*b^{t+1}, \*\theta^{t+1}) \|_2
\end{align*}
Here again, to assure the convergence of an infinite series to a finite value, we have
\begin{align*}
 \lim_{t \to \infty}
 \| \nabla_{\*z} \tilde{\ell}(\*z^t) \|_2^2 \to 0. 
 % \| \nabla_{\*z} \tilde{L}(\*B^{t}, \*b^{t}, \*\theta^{t}) \|_2 \to \*0
\end{align*}

Next, we prove our temporarily assumption $\*z^t \in \cB_{\rho}(\*z^*)$ by using similar approach to Theorem~8 of \citep{attouch2010proximal}.
% 以下を満たす$\*z^t$を$t = 0$として置き直す．
% \memo{(\citep{attouch2010proximal}のTheorem~8参照)}
We re-define the index $t$ such that it satisfies the following inequality as $t = 0$: 
\begin{align}
 % \| \*z^* - \*z_0 \|_2 + 2 \sqrt{ \frac{ \tilde{\ell}(\*z^0) - \tilde{\ell}(\*z^1) }{ C_1 } } + \frac{\sqrt{C_2}}{C_1} \phi(\tilde{\ell}(\*z^1) - \ell^*)
 % \| \*z^* - \*z^0 \|_2 + 2 \sqrt{ \frac{ \tilde{\ell}(\*z^0) - \ell^* }{ C_1 } } + \frac{\sqrt{C_2}}{C_1} \phi(\tilde{\ell}(\*z^0) - \ell^*)
 \| \*z^* - \*z^t \|_2 + 2 \sqrt{ \frac{ \tilde{\ell}(\*z^t) - \ell^* }{ C_1 } } + \frac{\sqrt{C_2}}{C_1} \phi(\tilde{\ell}(\*z^t) - \ell^*)
 \leq \rho.
 \label{eq:z0-condition}
\end{align}
% $\*z^t$は$\*z^*$へ収束する部分列を含むのでこれが成立する$t$が存在するはず．
Because of lemma~\ref{lem:accumulation}, there exist $t$ that satisfies the above inequality. 
After replacing the above $t$ with $0$,
% この$\*z^0$は当然，
% $\*z^0 \in \cB_\rho(\*z^*)$．
obviously, 
$\*z^0 \in \cB_\rho(\*z^*)$.
%
% このとき，三角不等式とlemma~\ref{lem:obj-low}の\eq{eq:obj-low}より
Then, from the triangle inequality and \eq{eq:obj-low} of lemma~\ref{lem:obj-low}, we see 
\begin{align}
 \| \*z^* - \*z^1 \|_2 
 & \leq 
 \| \*z^* - \*z^0 \|_2 + \| \*z^0 - \*z^1 \|_2 
 \notag \\ % ---
 & \leq 
 \| \*z^* - \*z^0 \|_2 + \sqrt{ \frac{ \ell(\*z^0) - \ell(\*z^1) }{ C_1 } }
 \notag \\ % ---
 & \leq 
 \| \*z^* - \*z^0 \|_2 + \sqrt{ \frac{ \tilde{\ell}(\*z^0) - \ell^* }{ C_1 } }
 \leq \rho.
 \label{eq:ub-z1}
\end{align}
% よって，
% $\*z^1 \in \cB_\rho(\*z^*)$
Therefore, by comparing the last inequality with \eq{eq:z0-condition}, we see
$\*z^1 \in \cB_\rho(\*z^*)$.
%
% 次に，ある$k \geq 1$まで
% $\*z^k \in \cB_\rho(\*z^*)$
% が成立し続けていると仮定する．
Next, we assume 
$\*z^k \in \cB_\rho(\*z^*)$
holds until some 
$k > 1$. 
%
% このとき，三角不等式と\eq{eq:ub-z-sequence}より(今$k$までは$\*z^k \in \cB_\rho(\*z^*)$を仮定してるので適用できる)
From the triangle inequality and \eq{eq:ub-z-sequence}, which can be used because now we assume $\*z^t \in \cB_\rho(\*z^*)$ for $t \leq k$, we obtain
\begin{align*}
 \| \*z^* - \*z^{k+1} \|_2 
 & \leq 
 \| \*z^{k+1} - \*z^1 \|_2 + \| \*z^* - \*z^1 \|_2
 \\ % --
 & \leq 
 \sum_{t = 1}^k \| \*z^{t+1} - \*z^{t} \|_2
 + \| \*z^* - \*z^1 \|_2
 \\ % --
 & \leq 
 \| \*z^1 - \*z^0 \|_2
 + \frac{\sqrt{C_2}}{C_1} \phi(\tilde{\ell}(\*z^1) - \ell^*)
 + \| \*z^* - \*z^1 \|_2
 \\ % --
 & \leq 
 \sqrt{ \frac{ \ell(\*z^0) - \ell(\*z^1) }{ C_1 } }
 + \frac{\sqrt{C_2}}{C_1} \phi(\tilde{\ell}(\*z^1) - \ell^*)
 \\
 & \qquad + \| \*z^* - \*z^0 \|_2 + \sqrt{ \frac{ \tilde{\ell}(\*z^0) - \ell^* }{ C_1 } }
 \\ % --
 & \leq 
 \| \*z^* - \*z^0 \|_2 + 2 \sqrt{ \frac{ \tilde{\ell}(\*z^0) - \ell^* }{ C_1 } } + \frac{\sqrt{C_2}}{C_1} \phi(\tilde{\ell}(\*z^0) - \ell^*).
\end{align*}
% 最後から二つ目の不等式は\eq{eq:obj-low}と\eq{eq:ub-z1}からである．
The second last inequality is from \eq{eq:obj-low} of lemma~\ref{lem:obj-low} and \eq{eq:ub-z1}. 
%
% よって，\eq{eq:z0-condition}から
% $\*z^{k+1} \in \cB_\rho(\*z^*)$
% となる．
Therefore, from \eq{eq:z0-condition}, we see
$\*z^{k+1} \in \cB_\rho(\*z^*)$.
%
% 以上より，帰納法により$\*z^0$以降の任意の$\*z^t \in \cB_\rho(\*z^*)$となる．
By induction, we obtain 
$\*z^t \in \cB_\rho(\*z^*)$
for any $t \geq 0$.

\end{proof}

% --------------------------------------------------
\section{Proof of theorem~\ref{thm:convergence-rate}}
\label{app:proof-rate}

% \red{\citep{attouch2009convergence}}

% $\Delta_k \coloneqq \sum_{t=k}^\infty \| \*z^{t+1} - \*z^t \|_2$ とする．
Let
$\Delta_k \coloneqq \sum_{t=k}^\infty \| \*z^{t+1} - \*z^t \|_2$.
%
% Note that 三角不等式より$\Delta_k \geq \| \*z^{k} - \*z^* \|_2$，かつ$\Delta_{k-1} - \Delta_{k} = \| \*z^{k} - \*z^{k-1} \|_2$．
Note that, from the triangle inequality, 
$\Delta_k \geq \| \*z^{k} - \*z^* \|_2$
and
$\Delta_{k-1} - \Delta_{k} = \| \*z^{k} - \*z^{k-1} \|_2$. 
%
% 式\eq{eq:ub-z-sequence}と同様の導出により，
Based on the same derivation as \eq{eq:ub-z-sequence}, we have
\begin{align*}
 % \sum_{t = 1}^T \nbr{ \*z^{t+1} - \*z^t }_2 
 % \leq
 % \| \*z^1 - \*z^{0} \|_2 
 % + \frac{ \sqrt{ C_2 } }{ C_1 } \phi(\tilde{\ell}(\*z^1) - \ell^*) 
 \sum_{t = k}^T \nbr{ \*z^{t+1} - \*z^t }_2 
 \leq
 \| \*z^k - \*z^{k-1} \|_2 
 + \frac{ \sqrt{ C_2 } }{ C_1 } \phi(\tilde{\ell}(\*z^k) - \ell^*). 
\end{align*}
%
% 右辺は$T$に依存してないので，
Since the right-hand-side of the above inequality does not depend on $T$, 
\begin{align}
 \Delta_k =
 \lim_{T \to \infty} \sum_{t = k}^T \nbr{ \*z^{t+1} - \*z^t }_2 
 & \leq
 \| \*z^k - \*z^{k-1} \|_2 
 + \frac{ \sqrt{ C_2 } }{ C_1 } \phi(\tilde{\ell}(\*z^k) - \ell^*) 
 \notag \\ % ---
 & =
 \Delta_{k-1} - \Delta_{k} 
 + \frac{ c \sqrt{ C_2 } }{ C_1 } (\tilde{\ell}(\*z^k) - \ell^*)^{1 - \theta}
 \label{eq:ub-delta-k}
\end{align}
%
% 以降，一般性を失うことなく，$\ell^* = 0$とする（cf,\citep{attouch2009convergence}）．
Hereafter, we set $\ell^* = 0$ without loss of generality \citep{attouch2009convergence}.
%
% この時，KL不等式より
Then, from the KL inequality, 
\begin{align}
 \phi^\prm(\tilde{\ell}(\*z)) 
 &\geq 
 % \frac{1}{ \| \partial \tilde{\ell}(\*z) \|_2}
 \frac{1}{ \| \nabla_{\*z} \tilde{\ell}(\*z) \|_2}
 % \memo{\qquad \text{劣微分の扱い}}
 \notag \\ % ---
 c ( 1 - \theta )
 (\tilde{\ell}(\*z))^{- \theta}
 & \geq 
 % \frac{1}{ \| \partial \tilde{\ell}(\*z) \|_2}
 \frac{1}{ \| \nabla_{\*z} \tilde{\ell}(\*z) \|_2}
 \notag \\ % ---
 (\tilde{\ell}(\*z))^{\theta}
 & \leq
 % c ( 1 - \theta ) \| \partial \tilde{\ell}(\*z) \|_2
 c ( 1 - \theta ) \| \nabla_{\*z} \tilde{\ell}(\*z) \|_2
 \label{eq:ub-obj-from-KL}
\end{align}
%
% \eq{eq:ub-delta-k}に代入し，\eq{eq:grad-upper}を使うと
Substituting this into \eq{eq:ub-delta-k} and combining with \eq{eq:grad-upper} of lemma~\ref{lem:grad-upper}, we see
\begin{align}
 \Delta_k &\leq 
 \Delta_{k-1} - \Delta_k
 % + \frac{ c \sqrt{ C_2 } }{ C_1 } ( c ( 1 - \theta ) \| \partial \tilde{\ell}(\*z) \|_2 )^{\frac{1 - \theta}{\theta}}.
 + \frac{ c \sqrt{ C_2 } }{ C_1 } ( c ( 1 - \theta ) \| \nabla_{\*z} \tilde{\ell}(\*z^k) \|_2 )^{\frac{1 - \theta}{\theta}}
 \notag \\ %--
 &\leq 
 \Delta_{k-1} - \Delta_k
 + \frac{ c  \sqrt{C_2}  }{ C_1 } ( c C_2 ( 1 - \theta ) )^{\frac{1 - \theta}{\theta}}
 \| \*z^k - \*z^{k-1} \|_2^{\frac{1 - \theta}{\theta}} 
 \notag \\ %--
 &=
 \Delta_{k-1} - \Delta_k
 + \frac{ c  \sqrt{C_2}  }{ C_1 } ( c C_2 ( 1 - \theta ) )^{\frac{1 - \theta}{\theta}}
 (\Delta_{k-1} - \Delta_k)^{\frac{1 - \theta}{\theta}}.
 \label{eq:ub-delta-k-by-delta}
\end{align}
%

% $\theta \in (1/2, 1)$ の場合 $\frac{1 - \theta}{\theta} < 1$.
When 
$\theta \in (1/2, 1)$,
we have $\frac{1 - \theta}{\theta} < 1$.
%
% $k \to \infty$
% で
% $\Delta_k \to 0$
% なので，
% $\Delta_{k-1} - \Delta_k \to 0$．
Further, when  
$k \to \infty$, 
we have
$\Delta_{k-1} - \Delta_k \to +0$. 
%
% ある$k \geq N_1$で
% $\Delta_{k-1} - \Delta_k \leq (\Delta_{k-1} - \Delta_k)^{\frac{1 - \theta}{\theta}}$．
Then, there exists $N_1$ such that 
$\Delta_{k-1} - \Delta_k \leq (\Delta_{k-1} - \Delta_k)^{\frac{1 - \theta}{\theta}}$ 
holds for
$\forall k \geq N_1$.
%
% よって，適当な定数$Z_1$により
Therefore, there exists a constant $Z_1$ that satisfies 
\begin{align*}
 \Delta_k^{\frac{\theta}{1-\theta}}
 \leq 
 Z_1
 (\Delta_{k-1} - \Delta_k)
\end{align*}
% という$k \geq N_1$で成立する関係を作ることができる．
for $k \geq N_1$.
%
% $h(s) = s^{- \frac{\theta}{1-\theta}}$と$R \in (1, \infty)$を定義する．
Define 
$h(s) = s^{- \frac{\theta}{1-\theta}}$
and
$R \in (1, \infty)$. 
For
$k \geq N_1$,
we first consider  the case that 
$h(\Delta_k) \leq R h(\Delta_{k-1})$. 
% とする (逆の場合は後で). 
%
% 上の不等式から
From the above inequality,
\begin{align*}
 1
 & \leq 
 Z_1
 (\Delta_{k-1} - \Delta_k)
 \Delta_k^{-\frac{\theta}{1-\theta}}
 \\ % ---
 & =
 Z_1
 (\Delta_{k-1} - \Delta_k)
 h(\Delta_k)
 \\ % --
 & \leq
 R Z_1
 (\Delta_{k-1} - \Delta_k)
 h(\Delta_{k-1})
 \\ % --
 & \leq
 R Z_1
 \int_{\Delta_k}^{\Delta_{k-1}}
 h(s)
 \mr{d} s
 \\ % --
 & \leq
 R Z_1 
 \frac{1 - \theta}{1 - 2 \theta}
 ( \Delta_{k-1}^{ \frac{1 - 2 \theta}{1 - \theta} } - \Delta_{k}^{ \frac{1 - 2 \theta}{1 - \theta} }). 
\end{align*}
By defining
$\mu = \frac{2\theta - 1}{(1-\theta) R Z_1} > 0$
and
$\nu = \frac{1 - 2\theta}{1 - \theta} < 0$,
we obtain
\begin{align*}
 0 < \mu \leq \Delta_k^\nu - \Delta_{k-1}^\nu.
\end{align*}
%
% 次に
% $h(\Delta_k) > R h(\Delta_{k-1})$
% を考える．
Next, we consider the case 
$h(\Delta_k) > R h(\Delta_{k-1})$.
Let
$q = (\frac{1}{R})^{\frac{1-\theta}{\theta}} \in (0,1)$. 
% と定義する． 
%
From
$h(\Delta_k) = \Delta_k^{-\frac{\theta}{1-\theta}} > R \Delta_{k-1}^{-\frac{\theta}{1-\theta}} = R h(\Delta_{k-1})$,
% から
we have
$\Delta_k^{\frac{\theta}{1-\theta}} \leq \frac{1}{R} \Delta_{k-1}^{\frac{\theta}{1-\theta}}$.
%
%であり，両辺$\frac{1-\theta}{\theta} < 1$乗して
Taking both sides to the power of 
$\frac{1-\theta}{\theta} < 1$,
we obtain 
\begin{align*}
 \Delta_k \leq q \Delta_{k-1}.
\end{align*}
% $\nu < 0$なので
Since $\nu < 0$,
\begin{align*}
 \Delta_k^\nu &\geq q^\nu \Delta_{k-1}^\nu
 \\ % ---
 \Delta_k^\nu - \Delta_{k-1}^\nu &\geq (q^\nu - 1) \Delta_{k-1}^\nu.
\end{align*}
Because of
$q^\nu - 1 > 0$
and
$\Delta_k \to +0$ for $k \to \infty$, 
% なので，
we see that there exists $\bar{\mu} > 0$ such that 
$(q^\nu - 1) \Delta_{k-1}^\nu > \bar{\mu}$
for
$k \geq N_1$.
% %となる
% $\bar{\mu} > 0$
% が
% $k \geq N_1$
% で存在することになる．
%
% よって
Therefore,
\begin{align*}
 \Delta_k^\nu - \Delta_{k-1}^\nu &\geq \bar{\mu}.
\end{align*}

Define
$\hat{\mu} = \min\{ \mu, \bar{\mu} \} > 0$.
For $k \geq N_1$,
\begin{align*}
 \Delta_k^\nu - \Delta_{k-1}^\nu &\geq \hat{\mu} > 0.
\end{align*}
%
% この不等式を
% $k - 1 = N_1$
% から
% $k = N > N_1$な$N$まで足すと
By summing this inequality from 
$k = N_1 + 1$
to
$k = N > N_1 + 1$, 
\begin{align*}
 \Delta_N^\nu - \Delta_{N_1}^\nu &\geq \hat{\mu} (N - N_1). 
\end{align*}
% この式を変形して任意の$N \geq N_1 + 1$に対して \memo{($\nu < 0$に注意)}
Since $\nu < 0$, there exists a positive constant $a$ such that 
\begin{align*}
 \Delta_N  &\leq  ( \Delta_{N_1}^\nu + \hat{\mu} (N - N_1) )^{1/\nu} \leq 
 ( \hat{\mu} (N - N_1) )^{1/\nu} \leq 
 a N^{-\frac{1-\theta}{2\theta-1}}. 
\end{align*}
% となる正の定数$a$が存在する．
%
%よって，$k \geq N_1$において
As a result, we obtain
\begin{align*}
 \| \*z^k - \*z^* \|_2 \leq \Delta_k \leq  a k^{-\frac{1-\theta}{2\theta-1}}
\end{align*}
for $k \geq N_1$.

%
% ($N \geq N_1 + 1$なので)
% \memo{
% \begin{itemize}
%  \item 定数の存在
% %
% \begin{align*}
%  ( \hat{\mu} (N - N_1) )^{1/\nu} \leq 
%  a N^{1/\nu}
% \end{align*}
% なる$a$の存在を考える．$b = -1/nu > 0$として
% \begin{align*}
%  ( \hat{\mu} (N - N_1) )^{-b} &\leq 
%  a N^{-b}
%  \\
%  \hat{\mu}^{-b} \rbr{ \frac{ N }{ N - N_1 } }^b
%  &\leq 
%  a 
% \end{align*}
% よって，$N \geq N_1 + 1$に対して
% $\frac{ N }{ N - N_1 }$
% の上限を考えればよい．
% %
% これは$N$に対して単調減少なので$N \geq N_1 + 1$において，
% $\frac{ N }{ N - N_1 } \leq N_1 + 1$.
% %
% よって，
% $a \geq \hat{\mu}^b (N_1 + 1)^b$
% なる$a$で不等式を成立させられる．
% \end{itemize}}
% }

% 次に$\theta \in (0,1/2]$の場合，$\frac{1-\theta}{\theta} \geq 1$．
Next, in the case of $\theta \in (0,1/2]$, we have $\frac{1-\theta}{\theta} \geq 1$.
%
% $k$が大きい時
% $\Delta_{k-1} - \Delta_k \to 0$
% なので，\eq{eq:ub-delta-k-by-delta}より適当な定数$Z_2$により
Since 
$\Delta_{k-1} - \Delta_k \to 0$, 
\eq{eq:ub-delta-k-by-delta} indicates that there exits a constant $Z_2$ for sufficiently large $k$: 
\begin{align*}
 \Delta_k \leq Z_2 (\Delta_{k-1} - \Delta_k)
 \\ %---
 \Delta_k \leq \frac{Z_2}{ 1 + Z_2 } \Delta_{k-1}
\end{align*}
% よって，$Q = \frac{Z_2}{ 1 + Z_2 } \in (0,1)$として適当な正の定数$Z_3$に対して
As a result, for a positive constant $Z_3$, it holds 
\begin{align*}
 \| \*z^k - \*z^* \|_2  
 \leq
 \Delta_k 
 \leq 
 Z_3 Q^k,  
\end{align*}
where 
$Q = \frac{Z_2}{ 1 + Z_2 } \in (0,1)$.

% $\theta = 0$の場合，\eq{eq:ub-obj-from-KL}より，$\*z \neq \*z^*$において
If $\theta = 0$, \eq{eq:ub-obj-from-KL} indicates 
\begin{align*}
 \| \nabla_{\*z} \tilde{\ell}(\*z) \|_2 \geq 1/c
\end{align*}
for 
$\*z \neq \*z^*$.
Then, from \eq{eq:obj-low} in lemma~\ref{lem:obj-low} and \eq{eq:grad-upper} in lemma~\ref{lem:grad-upper},
%lemma~\ref{lem:obj-low}の\eq{eq:obj-low}と
%lemma~\ref{lem:grad-upper}の\eq{eq:grad-upper}より
\begin{align*}
 \ell(\*B^{t+1}, \*b^{t+1}, \*\theta^{t+1}) 
 & \leq 
 \ell(\*B^{t}, \*b^{t}, \*\theta^t) 
 -
 C_1 \nbr{ \*z^{t+1} - \*z^t }_2^2
 \\ % ---
 & \leq 
 \ell(\*B^{t}, \*b^{t}, \*\theta^t) 
 - \frac{1}{C_2}
 \| \nabla_{\*z} \tilde{\ell}(\*z^{t+1}) \|_2^2
 \\ % ---
 & \leq 
 \ell(\*B^{t}, \*b^{t}, \*\theta^t) 
 - \frac{1}{c C_2}
\end{align*}
% This implies that 有限回の更新で$\*z^*$に至る
This implies that $\*z^*$ should be achieved by finite iterations.

As a results, the worst case of $\theta \in (1/2,1]$, $\theta \in (0,1/2]$, and $\theta = 0$ can be represented as shown in Theorem~\ref{thm:convergence-rate}.

% --------------------------------------------------
\section{Discussion on $L_{1,2}$ based EIN}
\label{app:L12}

% 今回は$L_{0,2}$制約を使ったが，$L_{1,2}$使っても同じような定式化はできる
While we employ the $L_{0,2}$ constraint, the similar sparse selection of subgraphs can also be formulated by an $L_{1,2}$ based approach.
%
% $L_{0,2}$に実践的なadvantageがいくつかあり採用した
% Because of a few practical advantages, we 
%
% まず，$L_{1,2}$の定式化を簡単に述べ，その後，違いを議論する
We here briefly describe an $L_{1,2}$ based formulation of EIN, denoted as EIN-$L_{1,2}$, and discuss a few practical advantages of our $L_{0,2}$ approach compared with $L_{1,2}$.

The objective function of EIN-$L_{1,2}$ is written as
\begin{align}
 \min_{\*B, \*b, \*\theta} \ &
 \ell(\*B, \*b, \*\theta)
 + \lambda
 \| \*B \|_{1,2} 
\end{align}
where
$\lambda$
is the regularization coefficient, and 
$\| \*B \|_{1,2} = \sum_{H \in \cH} \| \*\beta_H \|_2$.
The similar alternating update algorithm can also be constructed for this problem.
A natural counterpart of the HT update of $\*B$ \eq{eq:update-B} can be defined by a proximal update 
\begin{align*}
  \*\beta^{t+1}_H & =
  \mr{prox} 
  \left(
  \*\beta_H - \eta \  \nabla_{\*\beta_H} \ell^t  %\*g_H %(\*B)
  \right) \text{ for } H \in \cH, 
\end{align*}
where
\begin{align*}
  \mr{prox}(\*a) = 
  \begin{cases}
   \rbr{ 1 - \dfrac{\eta \lambda}{\| \*a \|_2} } \*a  & \text{if} \: \| \*a \|_2 > \eta \lambda, \\
   \bm{0} &\text{if} \: \| \*a \|_2 \leq \eta \lambda. \\
  \end{cases}
\end{align*}
It can be seen that, for $H \in \{ H \mid \*\beta^{t}_H = \*0 \}$, 
\begin{align*}
  \mathrm{prox} \left(
  \*\beta^t_{H} - \eta \ \nabla_{\*\beta_H} \ell^t
  \right) = \*0	
\end{align*}
if 
$\| \nabla_{\*\beta_H} \ell^t \|_2 \leq \lambda$. 
%
% 勾配のupper boundを使うと同じようにpruningを定義できる
By using the same upper bound \eq{eq:grad-UB}, a similar pruning strategy can be defined.
Assuming $H \sqsubseteq H^\prm$ and $H^\prm \in \{ H \mid \*\beta^{t}_H = \*0 \}$, we have
\begin{align*}
 \mr{UB}(H) \leq \lambda \ \Rightarrow \ 
 \*\beta^t_{H^\prm} = \*0.
\end{align*}
In Algorithm~\ref{alg:regularization-path}, we gradually increase $s$. 
Instead, in EIN-$L_{1,2}$, $\lambda$ is gradually decreased, by which the strength of the sparse regularization can be gradually decreased.
Since the regularization parameter is often controlled in the log space, for example, the $k$-th candidate $\lambda_k$ is created by 
$\log(\lambda_{k}) = \log(\lambda_{k-1}) - \Delta \lambda$, 
where 
$\lambda_{0} = \lambda_{\max}$
is the pre-specified largest value of $\lambda$, and
$\Delta \lambda$ is also a pre-specified value for the decrease of $\lambda$. 
% as a grid in  
% $[\log(\lambda_{\max})$, $\log(0.01\lambda_{\max})]$, 
% where 
% $\lambda_{\max}$
% is the pre-specified largest value of $\lambda$. 
%
% From 
% $\log(\lambda_{\max})$, 
% we iteratively decrease $\Delta \lambda$ five times
% (i.e., $\log(\lambda_{k+1}) = \log(\lambda_{k}) - \Delta \lambda$).

Unlike the $L_{0,2}$ constraint, we cannot explicitly specify the number of non-zero components in $L_{1,2}$. 
This means that we cannot control the number of selected subgraphs. 
In the $L_{1,2}$ formulation, we observed that the number of selected subgraphs sometimes suddenly increases during optimization and when $\lambda$ changes (e.g., more than 10,000 selected subgraphs increase in the middle of optimization even when only a few subgraphs are selected in the initial several steps).
A possible remedy is to control $\lambda$ adaptively to the number of selected subgraphs (i.e., if the number of selected subgraphs largely increases, decrease $\Delta \lambda$ when $k > 0$ or decrease $\lambda_0$ when $k = 0$).
However, this is no more than sensible heuristics, which does not guarantee that reasonable amount of new selected subgraphs can be always generated by this approach. 
On the other hand, the $L_{0,2}$ approach is free from this problem, because the number of selected subgraphs can be explicitly controlled.

We empirically see that the prediction accuracy of EIN-$L_{0,2}$ and EIN-$L_{1,2}$ is not largely different.
Table~\ref{tab:L02-vs-L12} shows a comparison based on the same setting as our main comparison Table~\ref{table-accuracy}. 
Note that for $L_{1,2}$, we used the adaptive $\Delta \lambda$ strategy (details are below).

\paragraph{Scheduling of $\lambda$ in $L_{1,2}$ setting} 
We determined $\lambda_{\max}$ by \\
$\lambda_{\max} = \max_{H \in \cH} \| \nabla_{\*\beta_H} \ell^0 \|_2$, 
in which $\*B^0$ and $\*b^0$ were initialized by a linear model \citep{nakagawa2016safe} and $\*\theta^0$ was randomly initialized.
The initial value of $\Delta \lambda$ is $(\log(\lambda_{\max}) - \log(0.01\lambda_{\max})) / 5$. 
If the number of non-zero $\| \*\beta_H \|$ increase $\geq 10$, then we update
$\Delta \lambda \leftarrow 0.5 \Delta \lambda$.

% --------------------------------------------------
% Table: Accuracy comparison L_{0,2} vs L_{1,2}
% --------------------------------------------------
% \begin{table*}[t]
%  \caption{Accuracy comparison between $L_{0,2}$ and $L_{1,2}$.} % 各モデルの正答率（10回の平均）．L21を除く全モデルの中で最良値を太字強調．
%  \label{tab:L02-vs-L12}
%     \begin{center}
%     \scalebox{0.8}{
%     \begin{tabular}{lccccccccc}
%          & BZR & COX2 & DHFR & ENZYMES & PTC\_MR & ToxCast & SIDER & Cycle & Cycle XOR \\ \hline
%         EIN-$L_{0,2}$ & {86.0$\pm$2.4} & 80.5$\pm$3.0 & 78.4$\pm$2.2 & 59.8$\pm$6.7 & 59.1$\pm$4.5 & 60.0$\pm$1.5 & 68.2$\pm$2.6 & {100.0$\pm$0.0} & {100.0$\pm$0.0} \\
%         EIN-$L_{1,2}$ & 85.9$\pm$3.2 & 80.2$\pm$2.8 & 79.9$\pm$3.4 & 64.5$\pm$6.4 & 59.5$\pm$3.5 & 61.2$\pm$2.8 & 70.0$\pm$3.3 & 100.0$\pm$0.0 & 100.0$\pm$0.0 \\
%         EIN-$L_{0,2}$-GNN & 83.1$\pm$2.4 & 79.7$\pm$2.1 & 77.0$\pm$4.9 & 73.2$\pm$6.7 & 57.1$\pm$5.0 & 59.3$\pm$2.5 & 69.0$\pm$1.1 & {100.0$\pm$0.0} & 98.5$\pm$2.8 \\
%         EIN-$L_{1,2}$-GNN & 87.3$\pm$2.8 & 80.1$\pm$1.8 & 81.8$\pm$2.6 & 70.8$\pm$5.9 & 59.6$\pm$2.7 & 62.4$\pm$2.9 & 69.9$\pm$2.7 & 100.0$\pm$0.0 & 100.0$\pm$0.0 \\ \hline
%         $L_{0,2}$ \# non-zeros & 46$\pm$29 & 63$\pm$41 & 60$\pm$38 & 64$\pm$32 & 60$\pm$36 & 62$\pm$27 & 51$\pm$44 & 1$\pm$0 & 18$\pm$30 \\
%         $L_{1,2}$ \# non-zeros & 71$\pm$21 & 79$\pm$16 & 101$\pm$57 & 74$\pm$9 & 1057$\pm$1795 & 232$\pm$499 & 221$\pm$276 & 2$\pm$0 & 13$\pm$5 \\
%     \end{tabular}
%     }
%     \end{center}
% \end{table*}
\begin{table*}[t]
 \caption{Accuracy comparison between $L_{0,2}$ and $L_{1,2}$.} % 各モデルの正答率（10回の平均）．L21を除く全モデルの中で最良値を太字強調．
 \label{tab:L02-vs-L12}
    \begin{center}
    \scalebox{0.8}{
    \begin{tabular}{lcccccccc}
         & BZR & COX2 & DHFR & ENZYMES & PTC\_MR & ToxCast & Cycle & Cycle XOR \\ \hline
        EIN-$L_{0,2}$ & {86.0$\pm$2.4} & 80.5$\pm$3.0 & 78.4$\pm$2.2 & 59.8$\pm$6.7 & 59.1$\pm$4.5 & 60.0$\pm$1.5 & {100.0$\pm$0.0} & {100.0$\pm$0.0} \\
        EIN-$L_{1,2}$ & 85.9$\pm$3.2 & 80.2$\pm$2.8 & 79.9$\pm$3.4 & 64.5$\pm$6.4 & 59.5$\pm$3.5 & 61.2$\pm$2.8 &  100.0$\pm$0.0 & 100.0$\pm$0.0 \\
        EIN-$L_{0,2}$-GNN & 83.1$\pm$2.4 & 79.7$\pm$2.1 & 77.0$\pm$4.9 & 73.2$\pm$6.7 & 57.1$\pm$5.0 & 59.3$\pm$2.5 & {100.0$\pm$0.0} & 98.5$\pm$2.8 \\
        EIN-$L_{1,2}$-GNN & 87.3$\pm$2.8 & 80.1$\pm$1.8 & 81.8$\pm$2.6 & 70.8$\pm$5.9 & 59.6$\pm$2.7 & 62.4$\pm$2.9 & 100.0$\pm$0.0 & 100.0$\pm$0.0 \\ \hline
        $L_{0,2}$ \# non-zeros & 46$\pm$29 & 63$\pm$41 & 60$\pm$38 & 64$\pm$32 & 60$\pm$36 & 62$\pm$27 & 1$\pm$0 & 18$\pm$30 \\
        $L_{1,2}$ \# non-zeros & 71$\pm$21 & 79$\pm$16 & 101$\pm$57 & 74$\pm$9 & 1057$\pm$1795 & 232$\pm$499 &  2$\pm$0 & 13$\pm$5 \\
    \end{tabular}
    }
    \end{center}
\end{table*}

\end{document}